\pdfoutput=1

\documentclass[11pt]{article}

\usepackage[preprint]{acl}

\usepackage{times}
\usepackage{latexsym}

\usepackage[T1]{fontenc}

\usepackage[utf8]{inputenc}

\usepackage{microtype}

\usepackage{inconsolata}

\usepackage{graphicx}

\usepackage{algorithm}
\usepackage{algorithmic}
\usepackage{booktabs}       
\usepackage{tabularx}

\title{Fighting Randomness with Randomness: Mitigating Optimisation Instability of Fine-Tuning using Delayed Ensemble and Noisy Interpolation}

\author{Branislav Pecher$^{\spadesuit}$$^\dagger$, Jan Cegin$^{\spadesuit}$$^\dagger$, Robert Belanec$^{\spadesuit}$$^\dagger$, \\ {\bf Jakub Simko$^\dagger$}, {\bf Ivan Srba$^\dagger$}, {\bf Maria Bielikova$^\dagger$} \\
  $^{\spadesuit}$ Faculty of Information Technology, Brno University of Technology, Brno, Czechia \\
  $^\dagger$ Kempelen Institute of Intelligent Technologies, Bratislava, Slovakia\\
  \texttt{\{name.surname\}}@kinit.sk}

\begin{document}

\maketitle
\begin{abstract}
While fine-tuning of pre-trained language models generally helps to overcome the lack of labelled training samples, it also displays model performance instability. This instability mainly originates from randomness in initialisation or data shuffling. To address this, researchers either modify the training process or augment the available samples, which typically results in increased computational costs. We propose a new mitigation strategy, called \textbf{Delayed Ensemble with Noisy Interpolation} (DENI), that leverages the strengths of ensembling, noise regularisation and model interpolation, while retaining computational efficiency. We compare DENI with 9 representative mitigation strategies across 3 models, 4 tuning strategies and 7 text classification datasets. We show that: 1) DENI outperforms the best performing mitigation strategy (Ensemble), while using only a fraction of its cost; 2) the mitigation strategies are beneficial for parameter-efficient fine-tuning (PEFT) methods, outperforming full fine-tuning in specific cases; and 3) combining DENI with data augmentation often leads to even more effective instability mitigation.
\end{abstract}

\section{Introduction}

\begin{figure}
    \centering
    \includegraphics[width=0.9\linewidth]{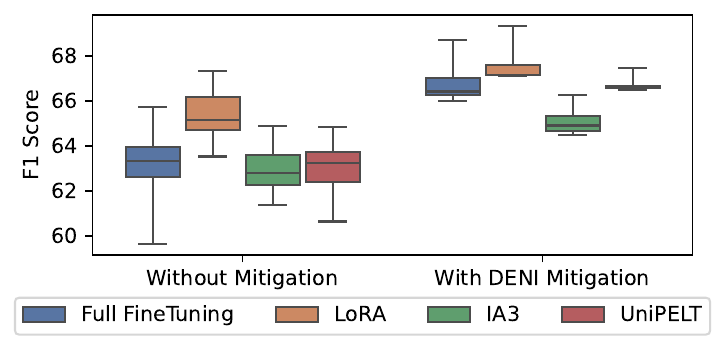}
    \caption{Repeating BERT fine-tuning multiple times without any mitigation leads to significant performance variance. Using DENI for randomness mitigation, the variance is reduced and performance increased.}
    \label{fig:variance_in_results}
\end{figure}

Tuning of pre-trained language models such as BERT or RoBERTa using either full fine-tuning or parameter efficient fine-tuning (PEFT) has achieved significant success across a wide range of natural language processing tasks. They are especially useful when faced with limited labelled data for quickly adapting to the specific task. Despite the success, previous works observed that fine-tuning still remains unstable~\citep{dodge2020fine, mosbach2021on, chen-etal-2022-revisiting}, especially with limited data. Fine-tuning is sensitive to the effects of randomness originating from random initialisation, data shuffling or model randomness (e.g., use of non-deterministic layers such as dropout in the model). As illustrated in Figure~\ref{fig:variance_in_results}, repeating the fine-tuning process multiple times, without mitigating the randomness, leads to large performance variance in the results, both for full fine-tuning and for PEFT methods.

To deal with the fine-tuning instability, researchers propose various strategies to mitigate the effects of randomness~\citep{pecher2024survey}. In most cases, the mitigation strategies focus on modifying the training process~\citep{Lee2020mixout, dodge2020fine}, such as adding noise~\citep{hua-etal-2021-noise, wu-etal-2022-noisytune, hua2023improving}, ensembling multiple models~\citep{hidey-etal-2022-reducing, khurana-etal-2021-emotionally, wang-etal-2023-two, summers2021nondeterminism}, or improving the experimental setup that potentially leads to the instability~\citep{zhang2021revisiting, mosbach2021on}, such as using bias correction or training for longer. As many of these strategies are designed and evaluated on high-resource datasets, almost no focus is dedicated to evaluating the mitigation benefit of data augmentation~\citep{zhang2018mixup, meng2023tuning}. At the same time, the instability of the whole training process is addressed only when considering full fine-tuning. When using PEFT methods, the focus is on the initialisation of soft prompts, even though the factors such as data shuffling still lead to variance in results~\citep{chen-etal-2022-revisiting}. Overall, the best performing mitigation strategies are ensembles and model interpolation methods~\citep{gueta-etal-2023-knowledge, hidey-etal-2022-reducing, wang-etal-2023-two}, which significantly reduce the deviation in results, but also significantly increase the computation costs. The methods that add noise to the model parameters~\citep{hua-etal-2021-noise, wu-etal-2022-noisytune} also perform well, improving generalisability and overall performance, but not necessarily reducing the instability.

Inspired by the success and mutual complementarity between ensemble methods, model interpolation and noise regularisation, we propose a novel mitigation method \textbf{Delayed Ensemble with Noisy Interpolation} (DENI). The \textbf{DENI} method leverages the benefits of ensembling while reducing its computation costs. To achieve this, the ensemble is created at the end of training from a single model by perturbing its parameters using random noise (i.e., fighting randomness with randomness). In addition, the method creates the ensemble by adding noise, trains it for a few steps and then aggregates it into a single model multiple times during training (see Figure~\ref{fig:method-illustration}), which leads to more effective mitigation of the randomness. Using the DENI method leads to lower variance in results and higher performance (as illustrated in Figure~\ref{fig:variance_in_results}).

To evaluate the benefit of the DENI method, we compare it with other representative mitigation strategies that modify the optimisation process. In the comparison, we also include an augmentation strategy that uses large language models to paraphrase samples, as such paraphrasing was observed to improve robustness and stability by~\citet{cegin-etal-2023-chatgpt, cegin2024effects}, especially in limited data settings. Besides full fine-tuning, we also explore the benefit of the proposed mitigation strategy and other baselines for representative parameter-efficient fine-tuning (PEFT) methods, namely LoRA, IA3 and UniPELT. 

Our main contributions and findings are\footnote{To support replicability and extension of our results, we openly publish the source code of our method and experiments at \url{https://github.com/kinit-sk/DENI}}:

\begin{itemize}
    \item We propose DENI - a novel strategy for mitigating the randomness sensitivity of fine-tuning (originating from initialisation, data shuffling and model randomness). The proposed method leverages the benefit of ensembling and model interpolation, while reducing their computation costs using noise.
    \item We compare DENI with 9 representative mitigation strategies, which either modify the training process or augment the data, across 7 text classification datasets. The results show that, in comparison with the best performing baselines, the DENI method improves the overall performance and reduces the deviation in results, while introducing lower computation costs, leading to an efficient mitigation.
    \item We explore the benefit of mitigation strategies across 3 representative PEFT methods. We find they often benefit more from the mitigation strategies, especially data augmentation, showing a larger decrease in deviation and increase in overall performance, even outperforming full fine-tuning in specific cases.
\end{itemize}

\section{Related Work: Mitigating Instability of Fine-Tuning}

The majority of strategies for mitigating the instability of fine-tuning modify the training process or the models themselves, addressing different sources of instability. One set of strategies addresses the suboptimal setup choices, suggesting that training for longer, using an optimiser with bias correction and lower learning rate with warm-up and scheduling can reduce the instability~\citep{mosbach2021on, zhang2021revisiting}. Another set of approaches modifies the initialisation, such as using meta-learning~\citep{dauphin2019metainit}, re-initialising top layers of the pretrained models~\citep{zhang2021revisiting}, or initialising multiple models and stopping the ones that show bad performance early in training~\citep{dodge2020fine}. Other strategies optimise only parts of the network~\citep{xu-etal-2021-raise, zhang2022fine} or run supplementary pre-training on data rich tasks~\citep{phang2018sentence}. Mixout~\citep{Lee2020mixout} randomly replaces parts of the parameters with the original weights. Further strategies add noise to the input or model parameters either before~\citep{wu-etal-2022-noisytune} or during~\citep{hua-etal-2021-noise, hua2023improving, chen-etal-2023-ptp} training. The ensembling strategies provide the best mitigation effectiveness, i.e., highest increase in performance and reduction in results variance, but at the cost of significant increase in computation costs~\citep{hidey-etal-2022-reducing, wang-etal-2023-two, gueta-etal-2023-knowledge}. To reduce the cost, the ensemble can be created from a single model, such using Stochastic Weight Averaging (SWA) that performs an equal average of the weights traversed by the optimiser~\citep{izmailov2018averaging, khurana-etal-2021-emotionally, lu-etal-2022-improving}, or Accelerated Ensembling that uses snapshots from different parts of training~\citep{summers2021nondeterminism, huang2017snapshot}. Another possibility is to share the lower layers and ensemble only the classification heads~\citep{chang-etal-2023-multi, liang-etal-2022-camero}. However, such aggregation also reduces the mitigation effectiveness of the ensemble as it reduces the diversity of the models.

Although many augmentation strategies exist, they are mainly used to improve the overall performance in low-resource settings, without any consideration for the instability~\citep{feng-etal-2021-survey, obadinma2023effectiveness}. Only a few papers consider using data augmentation for addressing the optimisation instability by utilising pretrained language models as data generators~\citep{cegin2024effects, meng_2023_tuning} or using common augmentation methods such as Mixup~\citep{guo2019augmenting, guo2020nonlinear}.

For the parameter-efficient fine-tuning (PEFT) methods, the focus is mostly on initialising prompts and mitigating randomness in prompt-tuning~\citep{chen-etal-2023-ptp, razdaibiedina-etal-2023-residual, pan-etal-2023-self, koksal-etal-2023-meal}, while the focus on mitigating optimisation instability for the remaining PEFT methods is limited~\citep{chen-etal-2022-revisiting}.

\section{Mitigation Method: Delayed Ensemble With Noisy Interpolation}
\label{sec:DENI}

\begin{figure*}
    \centering
    \includegraphics[width=1\linewidth]{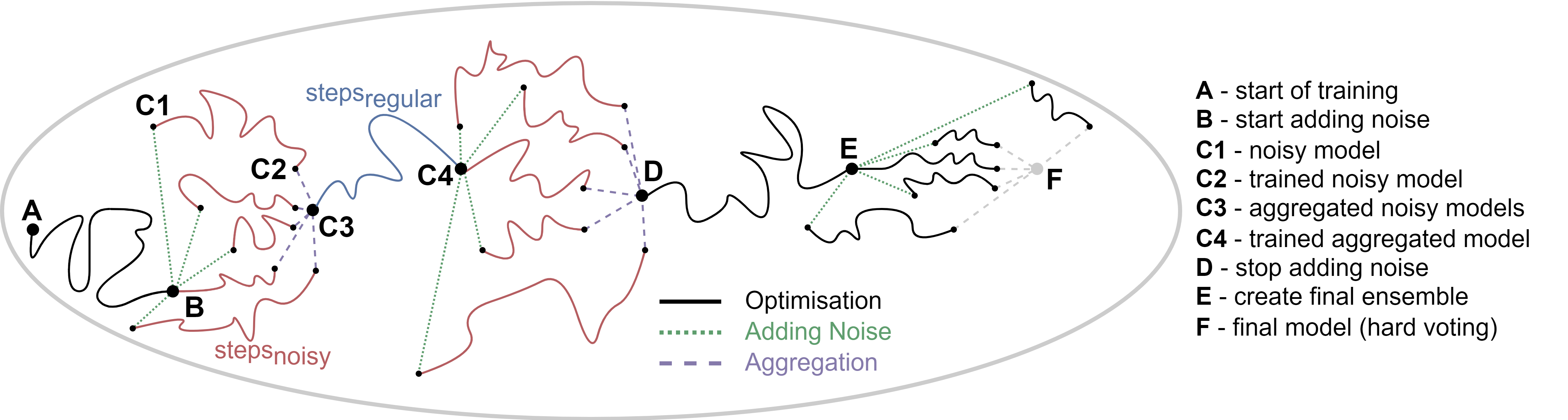}
    \caption{An illustrative example of the Delayed Ensemble with Noisy Interpolation (DENI) method that mitigates model performance instability of model fine-tuning with limited data. DENI alters regular fine-tuning, using noise-adding, model aggregation, and ensembling to steer model(s) towards optimal parameter setup in the parameter space. In comparison to simple ensembling, the method requires only a fraction of computational resources.
    }
    \label{fig:method-illustration}
\end{figure*}

The goal of our proposed method, the \textit{Delayed Ensemble with Noisy Interpolation} (DENI), is to mitigate the randomness in fine-tuning optimisation, reducing the observed variability in results, while keeping or increasing the average model performance and minimising additional computational costs. When designing our method, we drew inspiration from findings and approaches from previous works on ensembling~\citep{summers2021nondeterminism}, model interpolation~\citep{gueta-etal-2023-knowledge} and noise regularisation~\citep{hua-etal-2021-noise, wu-etal-2022-noisytune, chen-etal-2023-ptp}, that were shown to be effective techniques capable of improving overall performance and mitigating randomness. Our method, which is illustrated in Figure~\ref{fig:method-illustration} and in Algorithm~\ref{alg:method}, comprises two main components: 1) \textbf{Delayed Ensemble}; and 2) \textbf{Noisy Interpolation}.

\begin{algorithm}[!tbh]
\caption{Delayed Ensemble with Noisy Interpolation} \label{alg:method}
\begin{algorithmic}[1]
\begin{footnotesize}
\REQUIRE 

$var_{noise}$: noise variance

$noise_{start}$: steps before adding noise

$noise_{end}$: steps after which no noise is added

$ensemble_{start}$: steps before creating final ensemble

$steps_{noisy}$: number of steps with noisy models

$steps_{regular}$: number of steps with single aggregated model

$\lambda$: scaling factor for noise

$M$: pretrained model 

$N$: number of models in ensemble

\STATE Train model $M$ for $noise_{start}$ steps
\STATE Set $steps = noise_{start}$
\WHILE{$steps < noise_{end}$}
    \FORALL{$n$ in $1, 2, ..., N$}
        \STATE $M_{n} = M + Noise(0, var_{noise}) * \lambda_{steps}$
    \ENDFOR
    \STATE Train $M, M_{1}, ..., M_{N}$ for $steps_{noisy}$ 
    \STATE $M = average(M, M_{1}, ..., M_{N})$
    \STATE Train $M$ for $steps_{regular}$
    \STATE $steps = steps + steps_{noisy} + steps_{regular}$
\ENDWHILE
\STATE Train $M$ for $ensemble_{start} - noise_{end}$
\FORALL{$n$ in $1, 2, ..., N$}
        \STATE $M_{n} = M + Noise(0, var_{noise}) * \lambda_{ensemble}$
\ENDFOR
\STATE Train $M, M_{1}, ..., M_{N}$ for the remaining steps
\STATE Use hard voting of the final ensemble $M, M_{1}, ..., M_{N}$
\end{footnotesize}
\end{algorithmic}
\end{algorithm}

\paragraph{Delayed Ensemble (DE).} The main idea is to exploit the benefit of ensembling multiple models, while reducing the computation cost of obtaining such an ensemble. It was previously shown, that independently training multiple models (initialised with different random seeds) and aggregating their predictions represents an effective regularisation technique capable of reducing the deviation in the results and significantly improving performance. In addition, \citet{gueta-etal-2023-knowledge} show that linear interpolation of multiple models in the weight space leads to a better performing model than any of the individual ones used for its aggregation. However, training multiple models with different set of weights introduces a significant increase in computation costs. To reduce the computation cost, while keeping as many of the benefits as possible, we propose the \textbf{Delayed Ensemble} that creates the ensemble by adding noise to a single trained model. Instead of initialising multiple models and training them to full, we initialise and train only a single model $M$. The model is trained for $ensemble_{start}$ steps (E in Figure~\ref{fig:method-illustration}), and afterwards is used to create $N$ new models ($M_1, M_2, ..., M_N$) by perturbing its parameters by adding noise as follows:
\begin{equation}
    M_{n} = M + Noise(0, var_{noise}) * \lambda_{ensemble}
\end{equation}
where $Noise(0,var)$ represents Gaussian noise with $0$ mean and variance of size $var$, and $\lambda_{ensemble}$ represents a noise scaling factor for the ensemble. The models are trained for the remainder of the steps, and the prediction is obtained using hard voting (point F in Figure~\ref{fig:method-illustration}).

\paragraph{Noisy Interpolation (NI).} Besides allowing us to create the ensemble from a single trained model, adding noise to the model before, during or even after training was found to be an effective regularisation by itself that can improve the overall performance, generalisability and mitigate the effects of randomness in training~\citep{hua-etal-2021-noise, wu-etal-2022-noisytune, hua2023improving}. At the same time, training the model that is a result of the linear interpolation of multiple models is more effective and leads to better performance than the further training of the individual models used for aggregation, especially when the interpolated model is in close proximity to the optimal set of parameters~\citep{gueta-etal-2023-knowledge}. Based on these findings, we propose a repeated \textbf{Noisy Interpolation}, which creates and interpolates the ensemble multiple times during training. After initialising the model $M$ (A in Figure~\ref{fig:method-illustration}), we train it for $noise_{start}$ steps (B in Figure~\ref{fig:method-illustration}). Afterwards, we create $N$ new models ($M_1, M_2, ..., M_N$) by perturbing the parameters of model $M$ by adding noise (C1 in Figure~\ref{fig:method-illustration}), similarly to \textit{Delayed Ensemble}. However, we use a different scaling factor, $\lambda_{steps}$, which performs inverse scaling based on the number of steps (i.e., in later stages of the training, the introduced noise is smaller). These noisy models are then trained for $steps_{noisy}$ steps (ending in C2 in Figure~\ref{fig:method-illustration}) and are aggregated together using uniform interpolation (C3 in Figure~\ref{fig:method-illustration}). The aggregated model is then trained for $steps_{regular}$ steps (ending in C4 in Figure~\ref{fig:method-illustration}). The whole process of creating and interpolating ensembles is repeated until $noise_{end}$ steps of training are finished (D in Figure~\ref{fig:method-illustration}). Afterwards, the aggregated model is trained until the end.

\paragraph{Full Method (DENI)} is the combination of the two components, i.e., running \textit{Noisy Interpolation} for large part of the training before creating final \textit{Delayed Ensemble} near the end.

\paragraph{Hyperparameters.} As we define multiple hyperparameters for the proposed method that can affect its mitigation effectiveness and efficiency, we provide recommendations on how to set them up for optimal performance. For the most important parameters (that represent the trade-off between mitigation and computations costs) we run a thorough hyperparameter sensitivity analysis in Appendix~\ref{sec:hyperparameter-sensitivity}. For the remaining hyperparameters we provide suggestions based on heuristics and our preliminary experiments. Although there are multiple possibilities for the $Noise()$ function, we have observed that using the Gaussian noise provides the best results. In addition, we suggest to add noise only to the newly initialised parameters. Adding noise to all layers makes the method significantly more sensitive to the other hyperparameters (especially $var_{noise}$ and $steps_{noisy}$) -- we observed that finding the optimal setup in such case becomes complicated, as adding slightly larger noise completely breaks the model (i.e., performing worse than random), while adding slightly smaller noise leads to negligible mitigation effects. Finally, similar to~\citet{wu-etal-2022-noisytune}, we found that scaling the noise based on the standard deviation of the model parameters provides the most benefit (again reducing hyperparameter sensitivity). At the same time, scaling the noise based on the number of steps is beneficial for the \textit{Noisy Interpolation}, but not \textit{Delayed Ensemble}. As such, we define the scaling factors in the following way: $\lambda_{steps} = std(W_{i}) * \frac{1}{steps}$ and $\lambda_{ensemble} = std(W_{i})$, where $std(Wi)$ represents the standard deviation of the parameter to which we are adding the noise. Overall, the DENI method is not as sensitive to the hyperparameter setup, even though the number of parameters is larger. In majority of the cases, the optimal parameters can be determined based on heuristics, while slight change leads to only a small difference in mitigation effectiveness, but only if using scaled noise and only for the newly initialised model parameters.

\paragraph{Reducing Memory Requirements and Inference Cost.} Although we mainly focus on the computation cost of training the models for the ensemble, the proposed strategy has a positive impact on memory requirements and the inference cost as well. As the noise is added only to the newly initialised parameters and the models in ensemble are created from a single model, majority of the parameters can be shared between the models in the ensemble. Thanks to the weight sharing (similar to~\citet{chang-etal-2023-multi, liang-etal-2022-camero}), the memory requirements and inference cost can be reduced significantly, being only slightly higher than using a single model (only the newly initialised parameters need to be duplicated, while reusing the single forward pass through the majority of the model as the newly initialised parameters are last layers in the network). However, the weight sharing strongly benefit depends on which parameters are perturbed -- if adding noise to all parameters or to parameters that are at the start of the model, the possibilities for weight sharing, and its benefits, would be limited.

\section{Experimental Results and Findings}

\paragraph{Baselines.} First, we use two basic baselines without any mitigation: 1) \textbf{Default}, where the model is fine-tuned only on the limited data (representing the lower bound of performance); and 2) \textbf{All Data}, where the model is fine-tuned on all the samples available in the dataset (representing the upper bound of performance). Second, we compare with other representative approaches that introduce a change into the optimisation process, specifically: 1) \textbf{Best Practices} as reported by \citet{mosbach2021on} and \citet{zhang2021revisiting}, which includes using scheduling, warm-up, optimiser with bias correction and training for longer; 2) \textbf{Ensemble}, where we concurrently train 10 models with different initialisation and use hard voting to aggregate their predictions; 3) \textbf{Noise$_{Input}$}, where we periodically add noise to the input embeddings during training; 4) \textbf{Noise$_{Weights}$}, where we periodically add noise to the parameters during training; 5) \textbf{Stochastic Weight Averaging (SWA)}~\citep{izmailov2018averaging, khurana-etal-2021-emotionally, lu-etal-2022-improving} that performs an equal average of the weights traversed by the optimiser with modified learning rate schedule; and 6) \textbf{Mixout}~\citep{Lee2020mixout} that stochastically mixes the parameters of two models instead of dropout. In addition, we also compare with \textbf{Augment N}~\citep{cegin-etal-2023-chatgpt, cegin2024effects}, which uses a large language model to paraphrase each sample $N$ times to extend the training data (we report results only for the best performing number of paraphrases, Augment $1$ and $2$). Finally, we compare our proposed method (\textbf{Our$_{DENI}$}), its individual components (\textbf{Our$_{DE}$} and \textbf{Our$_{NI}$}) and its extended version \textbf{Delayed Ensemble with Noisy Interpolation and Augmented Labelled Samples (Our$_{DENIALS}$}) that combines DENI with Augment 1. More details for each of the strategies are reported in Appendix~\ref{app:mitigation_strategies_setup}.

\paragraph{Datasets.} The experiments are conducted on 7 text classification datasets composed of different tasks with different number of classes. We focus on 3 binary classification datasets from the GLUE benchmark~\cite{wang-etal-2018-glue}: \textbf{SST2}~\cite{socher-etal-2013-recursive} for sentiment classification, \textbf{CoLA}~\cite{warstadt-etal-2019-neural} for determining the grammatical acceptability of a sentence, and \textbf{MRPC}~\cite{dolan-brockett-2005-automatically} for determining the semantic equivalence relationship between two sentences. In addition, we use 4 multi-class text datasets: \textbf{AG News}~\cite{zhang2015agnews} for news classification, \textbf{TREC}~\cite{voorhees2000trec} for question classification, \textbf{DBPedia}~\cite{lehmann2015dbpedia} for topic classification and \textbf{SNIPS}~\cite{coucke2018snips} for intent classification.

\paragraph{Experimental Setup.} As there is no consensus what constitutes a low-resource setting (e.g., for some it is having less than 1000 labelled samples~\citep{chen-etal-2022-revisiting, zhang2021revisiting, gueta-etal-2023-knowledge, chen-etal-2023-ptp} and for others less than 10 000~\citep{hua-etal-2021-noise}), we opted to use only 1000 labelled samples from each dataset for training for the main set of experiments and explore different sizes of available training data in an experiment presented in Section~\ref{sec:size_change}. Each experiment is repeated 20 times with different random seed that affects the initialisation, order of samples and model randomness. For each experiment, we report a mean F1 macro score and standard deviation over these repeated runs. Further experimental setup details are reported in Appendix~\ref{app:general_setup}.

\paragraph{Models.} Each experiment is run using the BERT~\citep{devlin-etal-2019-bert}, RoBERTa~\citep{liu2019roberta} and ALBERT~\citep{lan2020albert} base models. Besides regular fine-tuning, we also report results for representative parameter-efficient fine-tuning methods, specifically LoRA~\citep{hu2022lora}, IA3~\citep{liu2022ia3} and UniPELT~\citep{mao-etal-2022-unipelt} (which combines LoRA, Pfeiffer Adapter~\citep{pfeiffer-etal-2020-mad} and Prefix-Tuning~\citep{li-liang-2021-prefix}). Further details regarding hyperparameter setup are reported in Appendix~\ref{app:general_setup}.

\subsection{DENI Method on Full Fine-Tuning}

\begin{table*}
\begin{center}
\footnotesize
\begin{sc}
\tabcolsep=0.11cm
\begin{tabular}{lcccccccc}
\toprule
 BERT   & AG News       & TREC  & SNIPS & DBPedia       & SST2  & MRPC  & CoLA  & Cost \\ \midrule
\multicolumn{9}{c}{Full FineTuning} \\
Default & 84.95$_{0.482}$       & 90.00$_{0.682}$       & 97.99$_{0.109}$       & 98.75$_{0.047}$       & 88.27$_{0.230}$       & 62.73$_{1.497}$       & 75.11$_{0.662}$       & $1$ \\
All Data        & 88.65$_{0.343}$       & 95.66$_{0.579}$       & 98.86$_{0.066}$       & 99.21$_{0.030}$       & 95.15$_{0.078}$       & 68.86$_{0.893}$       & 79.56$_{0.440}$       & $5-55$ \\ \midrule
Best Practices  & 84.96$_{0.444}$       & 90.40$_{0.585}$       & 98.05$_{0.137}$       & 98.76$_{0.049}$       & 88.32$_{0.313}$       & 63.07$_{1.593}$       & 75.30$_{0.718}$       & $2$ \\
Ensemble        & 85.56$_{0.304}$       & 90.67$_{0.400}$       & 98.22$_{0.076}$       & 98.82$_{0.021}$       & \underline{89.05$_{0.079}$}       & 65.69$_{0.674}$       & 76.54$_{0.325}$       & $10$ \\
Noise$_{Input}$ & 85.07$_{0.444}$       & 89.43$_{0.775}$       & 98.09$_{0.114}$       & 98.77$_{0.043}$       & 88.25$_{0.285}$       & 62.44$_{1.499}$       & 75.05$_{0.817}$       & $1$ \\
Noise$_{Weights}$       & 85.25$_{0.671}$       & 90.20$_{0.860}$       & 98.03$_{0.146}$       & 98.75$_{0.061}$       & 88.04$_{0.266}$       & 62.68$_{1.765}$       & 75.06$_{0.699}$       & $1$ \\
SWA     & 85.40$_{0.364}$       & 90.24$_{0.632}$       & 98.06$_{0.122}$       & 98.76$_{0.060}$       & 88.54$_{0.211}$       & 63.50$_{1.751}$       & 75.26$_{0.572}$       & $1.75$ \\
Mixout  & 84.96$_{0.538}$       & 90.20$_{0.603}$       & 98.03$_{0.111}$       & 98.73$_{0.040}$       & 88.39$_{0.280}$       & 63.23$_{1.137}$       & 75.33$_{0.572}$       & $1$ \\ \midrule
Augment 1       & 85.25$_{0.542}$       & 90.36$_{0.945}$       & 98.19$_{0.142}$       & 98.82$_{0.082}$       & 88.66$_{0.286}$       & 63.01$_{1.544}$       & 75.56$_{0.443}$       & $2.4$ \\
Augment 2       & 85.28$_{0.433}$       & 90.45$_{0.676}$       & 98.10$_{0.143}$       & 98.72$_{0.076}$       & 88.63$_{0.327}$       & 62.40$_{1.514}$       & 74.58$_{0.426}$       & $3.6$ \\ \midrule
Our$_{DE}$      & 85.17$_{0.291}$       & 90.54$_{0.278}$       & 98.10$_{0.086}$       & \underline{98.76$_{0.019}$}       & 88.43$_{0.107}$       & 63.68$_{0.594}$       & 75.09$_{0.281}$       & $1.9$ \\
Our$_{NI}$      & 86.81$_{0.760}$       & 91.42$_{0.901}$       & 98.42$_{0.131}$       & 98.99$_{0.044}$       & 90.08$_{0.284}$       & 65.38$_{1.571}$       & 77.16$_{0.586}$       & $2.8$ \\
Our$_{DENI}$    & 87.67$_{0.261}$       & 92.04$_{0.377}$       & 98.65$_{0.099}$       & \textbf{99.07$_{0.023}$}       & 90.82$_{0.129}$       & 66.66$_{0.613}$       & 77.74$_{0.263}$       & $3.7$ \\
Our$_{DENIALS}$ & \underline{\textbf{88.18$_{0.196}$}}       & \underline{\textbf{92.27$_{0.236}$}}       & \underline{\textbf{98.73$_{0.075}$}}       & 99.03$_{0.040}$       & \textbf{91.22$_{0.132}$}       & \underline{\textbf{67.04$_{0.544}$}}       & \underline{\textbf{78.29$_{0.153}$}}       & $7.4$ \\
\bottomrule
\end{tabular}
\end{sc}
\end{center}
\caption{Comparison of the DENI method with existing mitigation strategies and baselines on full fine-tuning using BERT. The comparison is done in terms of overall performance, deviation and cost (normalised training steps). The highest performance is in \textbf{bold} and lowest deviation is \underline{underlined} (not considering \textit{All Data} baseline).}
\label{tab:main_results}
\end{table*}

In this section, our goal is to answer the following research question: \textit{\textbf{RQ1:} How does the DENI method perform in comparison to other mitigation strategies?} We compare the proposed method with existing mitigation strategies in terms of the overall performance, standard deviation and their computation cost. The computation cost is calculated using the number of training steps normalised by the steps of the \textit{Default} strategy (i.e., $cost=\frac{n\_steps_{strategy}}{n\_steps_{Default}}$). In addition, we explore what benefit the different components of the DENI method bring. The results from the comparison on full fine-tuning are presented for the BERT model in Table~\ref{tab:main_results}, with the full results in Appendix~\ref{app:full_results} and visualisation for the relation between performance, deviation and the cost in Appendix~\ref{app:cost-rel}.

\textbf{Strategies that significantly increase number of steps provide the most benefit.} The \textit{Ensemble} represents the most effective mitigation strategy, showing the highest increase in the overall performance (e.g., from $75.11$ to $76.54$ on CoLA dataset) and reduction in standard deviation (e.g., from $0.662$ to $0.325$ on CoLA dataset), with the deviation being often lower than the \textit{All Data} baseline. However, it also represent the most computationally expensive strategy (when not considering training on all data where the cost is dataset dependent). The \textit{Ensemble} strategy is closely followed by \textit{SWA}, \textit{Best Practices} and \textit{Augment N} that are less consistent and show lower benefit, but also require lower number of steps. \textbf{The \textit{Augment} strategy is beneficial only when using at most $2$ paraphrases per samples}, with higher number of paraphrases leading to significant decrease in performance (see Appendix~\ref{app:augment_full} for results for higher number of $N$). The remaining mitigation strategies show only negligible and dataset-dependent benefit.

\textbf{The \textit{DENI} method leads to higher performance and lower standard deviation while requiring lower number of steps than the \textit{Ensemble} strategy}. In comparison to the best performing mitigation strategy (\textit{Ensemble}), we observe a statistically significant increase of $0.25-2.11$ percentage points in performance (p-value of 7e-5 using Mann-Whitney U test), with the deviation staying similar or even lower in specific cases. At the same time, the \textit{DENI} method requires only 37\% of the training steps. As such, it represents an effective and efficient mitigation strategy.

\begin{table*}[tbh!]
\begin{center}
\footnotesize
\begin{sc}
\tabcolsep=0.11cm
\begin{tabular}{lccccccc}
\toprule
BERT        & AG News       & TREC  & SNIPS & DBPedia       & SST2  & MRPC  & CoLA \\ \midrule
\multicolumn{8}{c}{LoRA} \\
Default & 85.10$_{0.488}$       & 89.68$_{0.770}$       & 98.06$_{0.142}$       & 98.72$_{0.077}$       & 87.95$_{0.315}$       & 61.70$_{6.377}$       & 74.80$_{0.907}$ \\
Ensemble        & 86.32$_{0.266}$       & 91.58$_{0.371}$       & 98.32$_{0.125}$       & 98.86$_{0.030}$       & 89.17$_{0.109}$       & 66.59$_{0.753}$       & 77.02$_{0.533}$ \\
Our$_{DENI}$    & 87.86$_{0.305}$       & 92.39$_{0.377}$       & 98.67$_{0.084}$       & 99.05$_{0.041}$       & 90.74$_{0.122}$       & 67.49$_{0.554}$       & 77.66$_{0.425}$ \\
Our$_{DENIALS}$ & 87.79$_{0.244}$       & 92.24$_{0.204}$       & 98.65$_{0.065}$       & 99.04$_{0.038}$       & 91.03$_{0.206}$       & 66.79$_{0.671}$       & 78.24$_{0.041}$ \\ \midrule
\multicolumn{8}{c}{IA3} \\
Default & 83.83$_{0.829}$       & 88.42$_{1.197}$       & 97.59$_{0.259}$       & 98.56$_{0.105}$       & 87.60$_{0.216}$       & 62.72$_{0.979}$       & 73.56$_{0.963}$ \\
Ensemble        & 85.39$_{0.327}$       & 90.54$_{0.304}$       & 98.16$_{0.082}$       & 98.78$_{0.034}$       & 88.55$_{0.092}$       & 64.54$_{0.767}$       & 75.99$_{0.440}$ \\
Our$_{DENI}$    & 86.74$_{0.279}$       & 90.84$_{0.238}$       & 98.40$_{0.117}$       & 99.03$_{0.041}$       & 90.15$_{0.121}$       & 65.00$_{0.473}$       & 76.70$_{0.295}$ \\
Our$_{DENIALS}$ & 87.08$_{0.574}$       & 92.26$_{0.127}$       & 98.42$_{0.128}$       & 99.04$_{0.042}$       & 90.35$_{0.224}$       & 67.02$_{0.378}$       & 78.30$_{0.037}$ \\ \midrule
\multicolumn{8}{c}{UniPELT} \\
Default & 84.97$_{0.582}$       & 87.13$_{1.316}$       & 97.63$_{0.194}$       & 98.56$_{0.262}$       & 87.82$_{0.233}$       & 63.77$_{0.798}$       & 74.91$_{0.403}$ \\
Ensemble        & 86.02$_{0.291}$       & 91.08$_{0.559}$       & 98.31$_{0.104}$       & 98.89$_{0.025}$       & 88.63$_{0.141}$       & 64.07$_{0.610}$       & 76.00$_{0.458}$ \\
Our$_{DENI}$    & 87.54$_{0.301}$       & 91.43$_{0.678}$       & 98.56$_{0.084}$       & 99.06$_{0.045}$       & 90.57$_{0.078}$       & 66.72$_{0.275}$       & 77.68$_{0.409}$ \\
Our$_{DENIALS}$ & 87.44$_{0.054}$       & 92.25$_{0.088}$       & 98.56$_{0.068}$       & 99.04$_{0.045}$       & 91.01$_{0.157}$       & 66.59$_{0.323}$       & 78.24$_{0.033}$ \\
\bottomrule
\end{tabular}
\end{sc}
\end{center}
\caption{Comparison of the best performing mitigation strategy, \textit{DENI} method and the \textit{Default} baseline for different parameter-efficient fine-tuning methods. The comparison is done in terms of overall performance and deviation.}
\label{tab:peft_picked_results}
\end{table*}

\begin{figure*}[tbh!]
    \centering
    \includegraphics[width=1\linewidth]{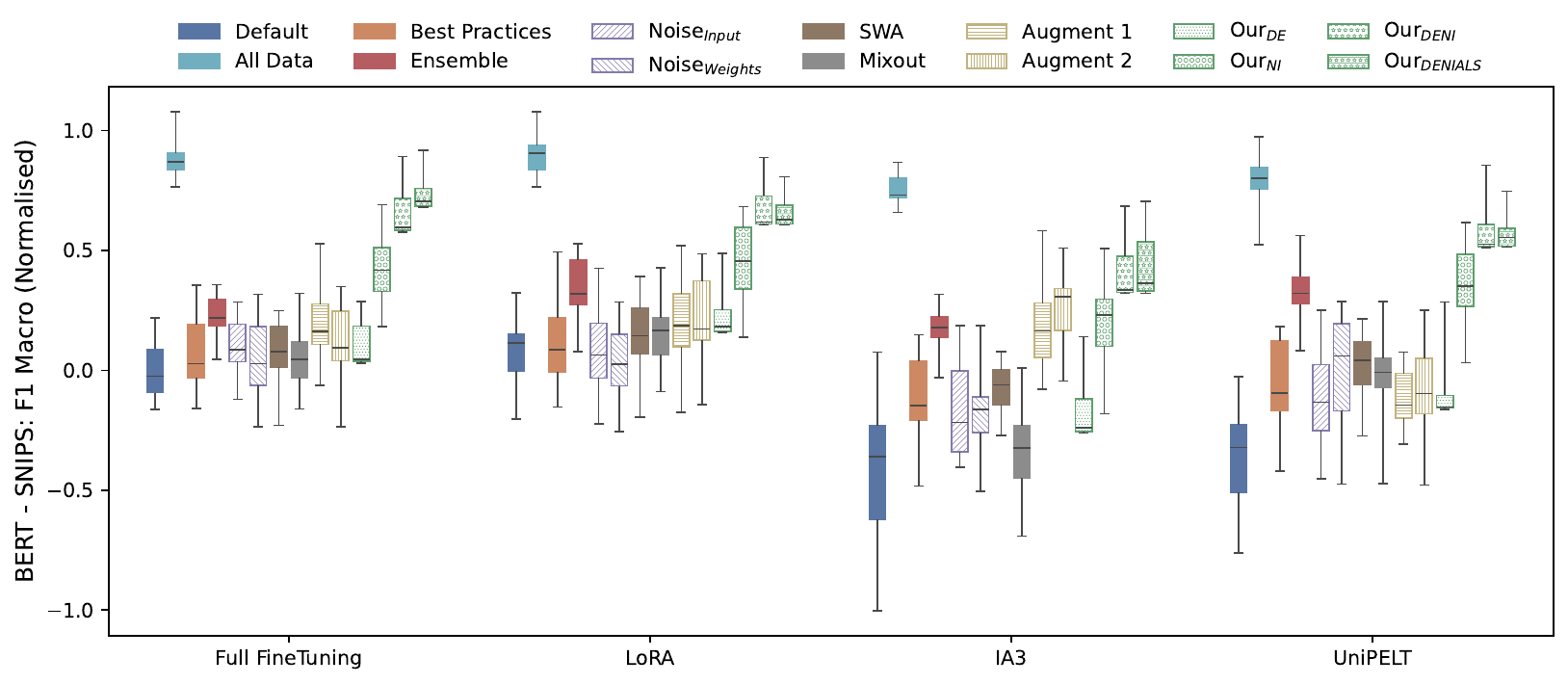}
    \caption{Benefit of mitigation strategies for the different fine-tuning methods using BERT on SNIPS dataset. The benefit is calculated as difference to the mean performance of the \textit{Default} baseline. The different mitigation strategies are beneficial for all fine-tuning methods, but with different overall benefit (e.g., \textit{Augment} on IA3). In addition, the \textit{DENI} method outperforms all mitigation strategies, leading to higher performance and lower deviation.}
    \label{fig:main-boxplot}
\end{figure*}

\textbf{The \textit{DENI} method components provide different benefits and complement each other.} The \textit{Delayed Ensemble} component leads to reduction of standard deviation, achieving deviation on par or lower than \textit{Ensemble}, but may not necessarily lead to increase in performance. For example on the MRPC dataset, even though the deviation is lower ($0.594$ as opposed to $0.674$), the performance is also significantly lower ($63.68$ as opposed to $65.69$). However, the performance is higher than the one from \textit{Default} baseline in all cases. On the other hand, the \textit{Noisy Interpolation} leads to significant increase in performance, at the cost of higher deviation in results. For example on the AG News dataset, the performance is increased to $87.67$ as opposed to $85.56$ from \textit{Ensemble} strategy, with the deviation also increasing to $0.760$ as opposed to $0.304$. In the full method, the components complement each other, suppressing the weaknesses and amplifying the strengths (e.g., reducing the deviation of NI also leads to higher performance). Finally, combining our proposed method with data augmentation may lead to further benefits, although the overall benefit is lower ($0.2 - 0.6$) and not statistically significant (p-value of 0.14 using Mann-Whitney U test) with a significant cost increase (double the training steps). As such, \textbf{combining mitigation strategies that target optimisation with augmentation strategies may lead to further increase} for all strategies.

\textbf{The effectiveness of the \textit{DENI} method and the different mitigation strategies is consistent across different models.} On the RoBERTa and ALBERT model, we still observe similar findings for the \textit{DENI} method and the majority of mitigation strategies, even though the effect of mitigation in absolute number may be different. In addition, the benefit of specific mitigation strategies grows on different models (\textit{SWA} or \textit{Augment} for ALBERT).

\subsection{Mitigation Strategies for PEFT Methods}
\label{sec:peft-experiments}

In this section, we aim to answer the following research question: \textit{\textbf{RQ2:} What is the benefit of mitigation strategies for parameter-efficient fine-tuning (PEFT) methods?} We compare the mitigation strategies and the \textit{DENI} method across different PEFT methods and observe how the increase in overall performance and the reduction in standard deviation changes. The results for the BERT model on the best performing mitigation strategies is in Table~\ref{tab:peft_picked_results}, for all strategies on the SNIPS dataset in Figure~\ref{fig:main-boxplot}, with the full results in Appendix~\ref{app:full_results}.

\textbf{Combination of the PEFT methods with the mitigation strategies can often lead to more effective mitigation.} Compared to full fine-tuning combination of \textit{Ensemble} with LoRA or UniPELT leads to higher increase in performance ($86.32$ for LoRA and $86.02$ for UniPELT compared to $85.56$ for full fine-tuning on TREC) and reduction in standard deviation ($0.266$ and $0.291$ compared to $0.304$). Similar behaviour can be observed on the \textit{Default} baseline as well, where with lower number of trained parameters, the performance or the standard deviation is on par, or even better, than when training all parameters. For example, we observe deviation of $0.403$ on CoLA dataset with UniPELT as compared to $0.662$ for full fine-tuning. As such, when faced with limited labelled data \textbf{the instability of training can be addressed by reducing the number of trainable parameters.}

\textbf{The \textit{Augment} mitigation strategy provides more benefit when reducing the number of trainable parameters.} The IA3 method, which trains approximately 0.5\% parameters, benefits significantly more from using augmented data. For example, we observe an increase in performance of up to $2$ percentage points on the MRPC dataset for the \textit{DENIALS} method over the \textit{DENI} method. At the same time, the \textit{Augment N} often outperforms the \textit{Ensemble} in terms of performance when using the IA3 method, but not necessarily in terms of deviation. The performance benefit of data augmentation is lower for other PEFT method, specifically up to $0.6$ percentage points for LoRA (CoLA dataset) and up to $0.8$ percentage points for UniPELT (SST2 dataset). Similarly, using augmented data leads to a significantly larger performance increase for ALBERT and lower increase for RoBERTa. Finally, \textbf{using augmented data in combination with the \textit{DENI} method often leads to lower deviation in results across all tuning methods}, especially for PEFT. For example, using UniPELT the deviation drops from $0.301$ to $0.054$ on AG News dataset, or from $0.678$ to $0.088$ on TREC dataset. \textbf{Similar results can be observed across different models for all of the datasets and tuning methods.}

\subsection{Behaviour on Different Sizes of Dataset}
\label{sec:size_change}

\begin{figure}
    \centering
    \includegraphics[width=1\linewidth]{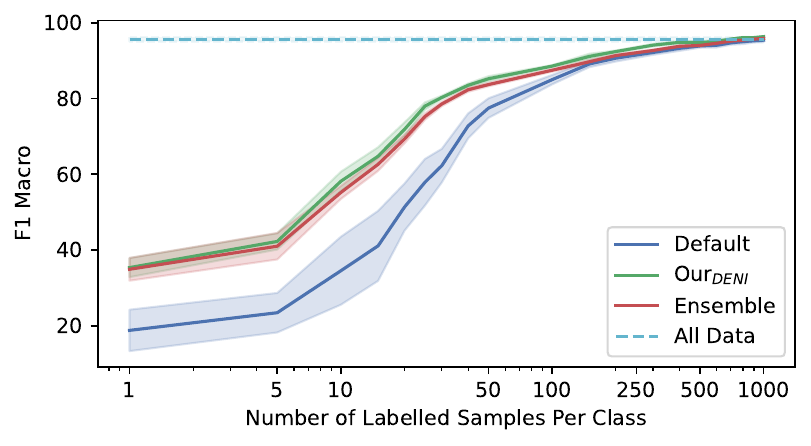}
    \caption{Mitigation effectiveness across different dataset sizes for the BERT model on TREC dataset. Benefit of mitigation strategies is higher on lower number of shots and gradually decrease with more shots.}
    \label{fig:size-change}
\end{figure}

Our goal in this section is to answer the following research question: \textit{\textbf{RQ3:} How does the number of available labelled samples (shots) affect the benefit and effectiveness of mitigation strategies?} We vary the number of available samples from 1 per class up to 1000 per class and compare the \textit{Ensemble} and \textit{DENI} method with the \textit{Default} and \textit{All Data} baselines. The results of this ablation study for the BERT model on the TREC dataset presented in Figure~\ref{fig:size-change} (the number of samples ranging from 6 in 1-shot setting up to 6000 in 1000-shot setting).

\textbf{Mitigation strategies provide larger benefit on low number of samples.} The difference in performance between the mitigation strategies and the \textit{Default} baseline is as high as $20$ percentage points ($80\%$ compared to $60\%$ when using $30$ shots), with significantly lower standard deviation ($0.453$ compared to $4.34$). As we increase the number of shots, the benefit gradually decreases. However, \textbf{the mitigation strategies provide their benefit even when using all data for training}, increasing the performance by up to $1$ percentage point and reducing deviation by up to $75\%$ (from $0.448$ to $0.116$).

\textbf{The \textit{DENI} method outperforms \textit{Ensemble} across all dataset sizes.} The performance difference oscillates between $0.4-3$ percentage points across the different shots. Similarly, the deviation stays the same, or even lower for the \textit{DENI} method, especially on lower number of shots (e.g., $2.095$ compared to $3.418$ when using 5 shots).

\section{Conclusions}

In this work, we have proposed a novel method for effective and efficient mitigation of the instability in fine-tuning of pre-trained language models based on ensembling, model interpolation and noise regularisation. We compare the method with prior mitigation strategies, showing it leads to stronger mitigation, with lower computation cost. In addition, we show that the mitigation strategies provide more benefit to the PEFT methods, such as LoRA, IA3 or UniPELT. Combining mitigation strategies with PEFT can often lead to higher performance and lower deviation than their combination with full fine-tuning. Moreover, we show that data augmentation provides more benefit when limiting the number of trainable parameters, such as using IA3 or training smaller models.

\section*{Acknowledgements}

This work was partially supported by \textit{VIGILANT}, a project funded by the European Union under the Horizon Europe under the Contract No. \href{https://doi.org/10.3030/101073921}{101073921}, the \textit{Central European Digital Media Observatory 2.0 (CEDMO 2.0)}, a project funded by the European Union under the Contract No. 101158609, and the \textit{MODERMED}, a project funded by the Slovak Research and Development Agency under GA No. APVV-22-0414.

This work was supported by the Ministry of Education, Youth and Sports of the Czech Republic through the e-INFRA CZ (ID:90254).

\section*{Limitations}
\label{sec:limitations}

The setting we focus on in the main experiments may obscure the overall benefit and effectiveness of different mitigation strategies. First, it was previously observed that smaller models with lower number of parameters show lower sensitivity to the effects of randomness when working with limited labelled data. As we use smaller models, namely base versions of BERT, RoBERTa and ALBERT, the benefit of our proposed method as well as other mitigation strategies may be lower than if we used larger models. In addition, we are quite conservative with what we consider as low resource setting and treat availability of 1000 labelled samples to already represent such. However, as we observed in the experiments, the performance is already quite high and the deviation quite low on some datasets even with this number of labelled samples. In addition, in the experiments from Section~\ref{sec:size_change}, we observe that the benefit of mitigation strategies on significantly lower number of samples per class is significantly higher. As such, this may obscure the real benefit of different mitigation strategies. However, our choices in both cases are based on previous works. As discussed in the Experimental Setup, many previous works consider 1000 labelled samples to already be low resource setting. At the same time, many of the mitigation strategies were designed and evaluated on high-resource setting with full datasets. Finally, we choose the smaller models in order to provide more extensive analysis and comparison, while limiting the impact (in terms of generated CO2) as much as possible.

Even though a large number of mitigation strategies exist, we focus on a smaller set of representative strategies. This is especially evident with augmentation strategies, where we focus only on augmentation using the paraphrasing from pretrained large language models. However, when choosing the strategies for comparison, we specifically focused on selecting the most representative ones, i.e., those that achieved the best mitigation (in terms of performance and standard deviation) in related work (and in some cases in our pilot experiments) from different groups of mitigation strategies. As such, even though we could increase the number strategies we compare to, it would not provide any additional findings. At the same time, our goal is to limit the potential negative impact of our work as much as possible and therefore have focused only on this set of samples.

Finally, we do not consider prompt-based parameter-efficient fine-tuning approaches (P-tuning~\citep{liu2023gpt}, Prefix-tuning~\citep{li-liang-2021-prefix} or Prompt-tuning~\citep{lester-etal-2021-power}), even though they are one of the most popular PEFT methods that are currently used for many pretrained large language models. However, we specifically focus on fine-tuning with typical classification models (that use classification head at the end and do not perform classification through prompting), and as such the prompt-based PEFT methods would not provide any significant benefit as they were specifically designed for training models for prompting and in-context learning. In addition, the UniPELT method already uses Prefix-tuning (which may be the most relevant for typical classification models) as one of the aggregated methods -- however, we have observed that this PEFT method performed the worst in our experiments and was especially sensitive to the hyperparameter setup (both for the models and for the mitigation strategies).

\bibliography{custom}

\appendix

\section{Ethical Considerations and Impact Statement}
\label{sec:ethics}
The experiments in this paper work with publicly available benchmark dataset GLUE (SST2, MRPC and CoLA from the benchmark), and publicly available datasets AG News, TREC, SNIPS and DB-Pedia, citing the original authors. As we were not able to determine the license for all of the tasks and datasets used, we opted to use them in as limited form as possible, adhering to the terms of use (no annotation of the test set) for the GLUE benchmark dataset and applying it to other datasets as well. As the datasets are commonly used, we do not check them for any identifiable information or offensive content, assuming the publicly available classification benchmark datasets do not contain such content (as it was already removed by the authors of the datasets). We do not work with any personally identifiable information or offensive content and perform no crowdsourcing for further data annotation. In addition, we are not aware of any potential ethical harms or negative societal impacts of our work, apart from the ones related to the advancement of the field of Machine Learning and Learning with Limited Labelled Data (mainly the use of computation resources, consuming energy and generating CO2). Finally, we follow the license terms for all the models we use -- all models and datasets allow their use as part of research. As we perform only classification and do not release the predictions, we generate no potentially biased or offensive content.

\paragraph{Impact Statement: CO2 Emissions Related to Experiments}
\label{compute-used}

The experiments presented in this paper used significant compute resources as we train multiple models (3) over multiple random seeds (20), for different mitigation strategies and baselines (17), fine-tuning methods (4) and datasets (7). Overall, all the experiments (including preliminary experiments for which we do not report results in this paper) were conducted using a private infrastructure, which has a carbon efficiency of 0.432 kgCO$_2$eq/kWh. A cumulative of approximately 1000 hours of computation was performed on hardware of type A100 PCIe 40GB (TDP of 250W). Total emissions are estimated to be 108 kgCO$_2$eq of which 0 percent were directly offset. These estimations were conducted using the \href{https://mlco2.github.io/impact#compute}{MachineLearning Impact calculator} presented in \citet{lacoste2019quantifying}. 
Whenever possible, we tried to reduce the compute resources used as much as possible. We evaluate on smaller models (base versions instead of large). In addition, we evaluate and compare our method only to a set of representative mitigation strategies, determined based on prior works and our preliminary experiments. Finally, the estimate does not include the cost of generating the paraphrases as the information about the efficiency for the large language model behind API is not available. To limit the impact of the augmentation, we decide to run it only a single time on a set of labelled samples that are used throughout the training.

\section{Hyperparameter Sensitivity Analysis}
\label{sec:hyperparameter-sensitivity}

\begin{figure*}
    \centering
    \includegraphics[width=1\linewidth]{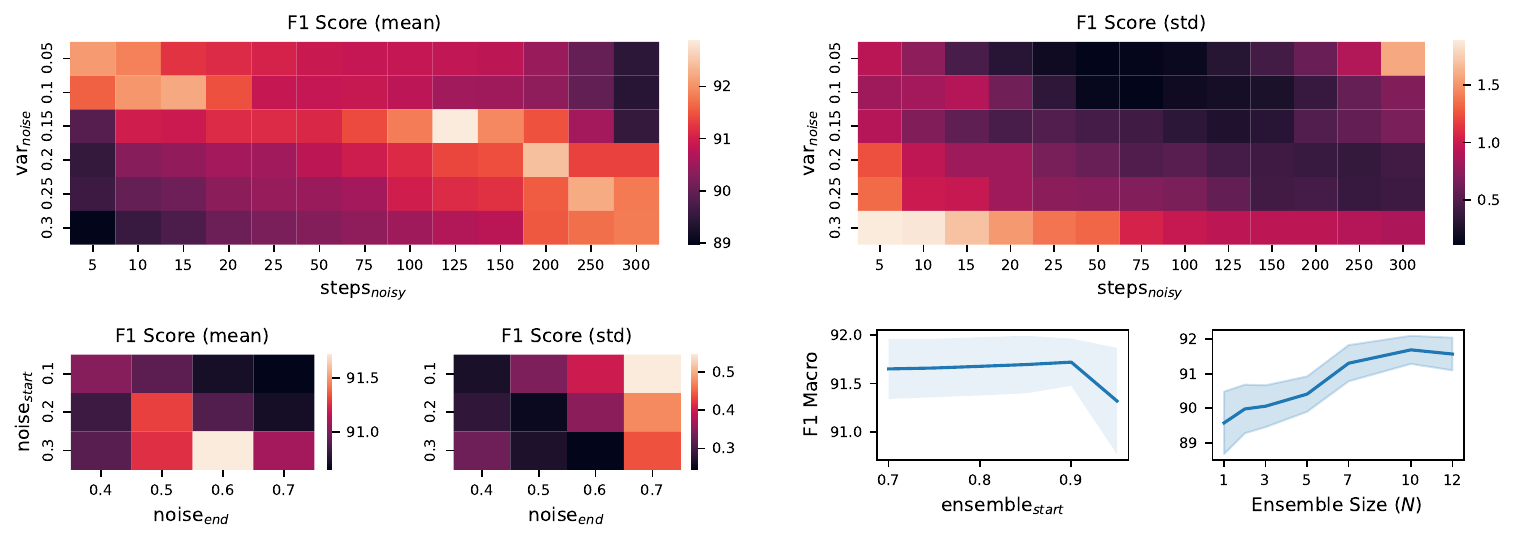}
    \caption{The effect of hyperparameter setup on the mitigation effectiveness of the DENI method, based on hyperparameter search for the BERT model on TREC dataset. The hyperparameters that affect each other are grouped together.}
    \label{fig:hyperparamas}
\end{figure*}

In this section, we explore the sensitivity of the \textit{DENI} method to the hyperparameter setup. We focus on the hyperparameters that represents the trade-off between mitigation effectiveness and computation costs. The hypeparameters that directly affect each other are grouped together as follows: 1) the size of noise we are adding ($var_{noise}$) together with how often the noise is added ($steps_{noisy}$); 2) number of steps after which we start ($noise_{start}$) and stop ($noise_{end}$) adding noise; 3) number of steps after which the final ensemble is created ($ensemble_{start}$); and 4) number of models in the ensemble ($N$). The results of this analysis are presented in Figure~\ref{fig:hyperparamas} for BERT on TREC dataset.

\textbf{The number of models in ensemble shows highest influence.} Increasing the number of models that are used in the ensemble leads to significant increase in mitigation effectiveness (increasing performance and reducing deviation), but also a significant increase in the cost. However, after a certain number ($10$), the improvement of mitigation strength is negligible. As such, we recommend using $8-10$ models in the ensemble, as it represents the optimal trade-off between computation and mitigation.

\textbf{When increasing the size of noise, we also need to add it less often (and vice versa) to get optimal performance}, in terms of both performance and deviation. Small noise should be added often, while large noise should be added only occasionally. On the other hand, adding large noise often leads to large decrease in overall performance, while adding small noise only sporadically leads to no mitigation. We find that the optimal value for both factors is at their midpoint -- adding noise of medium size ($0.15$) every so often ($125$ steps). However, this only holds when using the scaling factor. \textbf{When foregoing the scaling factor, we observe a significant sensitivity to the size of noise}.

\textbf{Noise-adding noise should be started later in the training.} At the start of optimisation, we observe a higher sensitivity of the model to the added noise, negatively affecting the mitigation effectiveness (i.e., leading to smaller performance increase and smaller decrease in deviation, often even being counterproductive). The sensitivity to the added noise slowly decreases as the training progresses (also due to the learning rate and scaling factor). As such, we observe that the optimal value for the $noise_{start}$ is after 30\% of the training. At the same time, \textbf{adding noise for a longer period of time increases the overall deviation in results, while increasing the overall performance only until a certain point.} For example, starting the noise early and adding it for 60\% of training leads to a large increase in deviation (and decrease in performance). Overall, optimal value for $noise_{end}$ is 30\% of training steps after the $noise_{start}$ value, regardless of its value. \textbf{Finally, we observe negligible impact of the $ensemble_{start}$ hyperparameter.} The only requirement is to leave enough steps of optimisation for the model to deal with the added noise that can break the learned knowledge (i.e., leaving enough steps for the different models to converge to their optimal parameters). As such, it is enough to create the ensemble after 90\% of training steps are already done, as creating the ensemble after this point leads to significant decrease mitigation (although if using larger number of steps, the start of ensemble may be delayed even further).

\textbf{Overall, the hyperparameter sensitivity of the \textit{DENI} method is not as significant.} In majority of the cases, the optimal parameters can be determined based on heuristics (e.g., adding large noise often breaks the models), while slight deviation from the optimal values does not leave to a significant changes in the mitigation effectiveness. However, this holds true only when using the recommended hyperparameter setup introduced in Section~\ref{sec:DENI} (using the scaling factor based on standard deviation of the individual parameters in the model and adding noise only to the newly initialised parameters).

\section{Experimental Setup: Further Details}
\label{app:experimental_setup_details}

\subsection{General Experimental Setup}
\label{app:general_setup}

For the experiments, we are using English only datasets from the GLUE~\citep{wang-etal-2018-glue} benchmark suite and other publicly available datasets. The datasets we use from GLUE benchmark are: 1) SST2~\citep{socher-etal-2013-recursive}, binary dataset with up to 68 000 labelled samples; 2) CoLA~\citep{warstadt-etal-2019-neural}, binary dataset with up to 10 000 labelled samples; and 3) MRPC~\citep{dolan-brockett-2005-automatically}, binary dataset with up to 6 000 labelled samples. The remaining datasets are all multi-class datasets, specifically: 1) Ag News~\cite{zhang2015agnews} with 4 classes and up to 120 000 labelled samples; 2) TREC~\cite{voorhees2000trec} with 6 classes and up to 6 000 labelled samples; 3) SNIPS~\citep{coucke2018snips} with 7 classes and up to 15 000 labelled samples; and 4) DBPedia~\citep{lehmann2015dbpedia} with 14 classes and up to 630 000 labelled samples. All datasets are split into training and testing using 80-20 split. In addition, the training dataset is split into training and validation set, with the validation set used for hyperparameter tuning. When selecting the 1 000 labelled samples for training for the main experiments (or smaller/larger number of samples for the size change experiments), we use uniform split, i.e., we select the same number of labelled samples for each class.

Each experiment is repeated 20 times, each time with different random seeds that influences the randomness originating from model initialisation, data shuffling and non-deterministic operations in the model (e.g., dropout). Each such repeat uses the same set of training and testing data, mainly in order to make the data augmentation easier (i.e., we can use the augmentation only single time on the set of labelled samples without worrying about introducing an information leak, such as using for training samples that are paraphrases of the ones in the test or validation set).

For the experiments, we use the base versions of 3 representative pretrained language models (obtained from HuggingFace~\citep{wolf2019huggingface}), specifically: 1) BERT~\citep{devlin-etal-2019-bert}\footnote{\url{https://huggingface.co/bert-base-uncased}} with 110M parameters; 2) RoBERTa~\citep{liu2019roberta}\footnote{\url{https://huggingface.co/roberta-base}} with 130M parameters; and 3) ALBERT~\citep{lan2020albert}\footnote{\url{https://huggingface.co/albert/albert-base-v2}} with 12M parameters. We add a Dropout layer with a drop rate of 0.3 followed by fully-connected classification layer on top of each of these models. Each model is trained for 10 epochs using Adam optimiser, learning rate of 1e-5 and batch size of 8 (with further modifications to these hyperparameters based on the mitigation strategy used, more information in Appendix~\ref{app:mitigation_strategies_setup}).

Besides full fine-tuning, where we train all parameters of the models, we also perform the experiments with parameter-efficient fine-tuning (PEFT) methods, specifically: 1) LoRA~\citep{hu2022lora}, applied to all attention layers, with rank $64$, alpha $64$, dropout $0.1$, with bias for all layers and using rank stabilisation~\citep{kalajdzievski2023rank}, that trains approximately $2\%$ of the parameters; 2) IA3~\citep{liu2022ia3}, applied to all atention and linear layers, that trains approximately $0.5\%$ of parameters; and 3) UniPELT~\citep{mao-etal-2022-unipelt}, which combines LoRA (same as above), Pfeiffer Adapter~\citep{pfeiffer-etal-2020-mad}, with reduction factor of $16$, and Prefix-Tuning~\citep{li-liang-2021-prefix}), with prefix-length of $25$, that trains approximately $10\%$ of the parameters. As specified in the Limitations section, we do not focus on any prompt-based PEFT methods (P-tuning~\citep{liu2023gpt}, Prefix-tuning~\citep{li-liang-2021-prefix} or Prompt-tuning~\citep{lester-etal-2021-power}), as these were not designed for optimising models with classification heads (and in our preliminary experiments, the prompt-based PEFT methods achieved significantly lower performance). Each of the PEFT methods has slightly different hyperparameters for training, as we have observed a significant sensitivity of the individual PEFT methods to the hyperparameter setup (with LoRA showing lowest sensitivity and UniPELT showing highest sensitivity). Specifically, we use following setup: 1) for LoRA we use learning rate of 1e-4 for BERT and ALBERT models and 1e-5 for RoBERTa model; 2) for IA3 we use learning rate of 75e-4; and 3) for UniPELT we use learning rate of 1e-4 for BERT, 25e-5 for RoBERTa and 6e-5 for ALBERT.

All of the hyperparameters, for all the combination of model and fine-tuning approach, are set using a separate hyperparameter optimisation using the validation data. This hyperparameter optimisation is done in a two-level fashion. First, the optimisation is run using large differences in the hyperparameter values, to find the approximate set of hyperparameters that should provide good performance on the given dataset. In the second step, we explore the hyperparameter space around these approximate hyperparameters, to find the optimal set of parameters. In addition, the performance in the hyperparameter search is evaluated from 5 repeated runs (changing the model initialisation, data order and model non-determinism, but not the labelled training or validation data), taking their average into consideration. When choosing the hyperparameter values in the first level, we draw inspiration from related work, using the optimal parameters reported in papers that propose, or use these approaches (such as~\citep{dodge2020fine, mccoy-etal-2020-berts, mosbach2021on, sellam_multiberts_2022}. However, we also search through additional hyperparameter values besides those reported in related works to better explore the parameter space and obtain as precise results from the investigation as possible.

For the size change experiments (Section~\ref{sec:size_change}) the experimental setup is slightly different. Instead of choosing the 1 000 labelled samples, we instead choose $N$ samples per class (called $N$-shots) and change the value of $N$, going from a low value up to the large part of the dataset. Specifically, we cover following $N$ values: 1, 5, 10, 15, 20, 25, 30, 40, 50, 100, 150, 200, 300, 400, 500, 600, 700, 800, 900 and 1000. Due to the large computation cost of the size change experiment, the experiment is done only for the full fine-tuning of the BERT model on the TREC dataset. The remaining experimental setup is kept the same as for the main experiments.

Similarly, the experimental setup for the hyperparameter sensitivity analysis (Appendix~\ref{sec:hyperparameter-sensitivity}) is slightly different. Due to the large number of combinations, we limit the number of repeated runs only to 5. In addition, we run the sensitivity analysis only for the full fine-tuning of the BERT model on the TREC dataset. The hyperparameters and their values we search through are: 1) $var_{noise}$ with values of 0.05, 0.1, 0.15, 0.2, 0.25 and 0.3; 2) $step_{noisy}$ with values of 5, 10, 15, 20, 25, 50, 75, 100, 125, 150, 200, 250 and 300; 3) $noise_{start}$ with values of 0.1, 0.2 and 0.3; 4) $noise_{end}$ with values of 0.4, 0.5, 0.6 and 0.7; 5) $ensemle_{start}$ with values of 0.7, 0.75, 0.8, 0.85, 0.9 and 0.95; and 6) ensemble size with values of 1, 2, 3, 5, 7, 9, 10 and 12.  The remaining experimental setup is kept the same as for the main experiments.

\subsection{Mitigation Strategies Setup}
\label{app:mitigation_strategies_setup}

\begin{table*}[!tbh]
\begin{center}
\begin{small}
\begin{tabularx}{\textwidth}{@{}p{0.2\linewidth}p{0.75\linewidth}@{}}
\toprule
\textbf{Dataset} & \textbf{Prompt}  \\ \midrule

AG News, TREC, SNIPS, DBPedia, SST2 & Rephrase an original question or statement 10 times. Format your responses like: 1. rephrase 1, 2. rephrase 2, ... , 10. rephrase 10. Original phrase: "\{\}". \\ \midrule

MRPC   & Paraphrase text consisting of two sentences in the format "Sentence 1: ... ; Sentence 2:..." Paraphrases both sentences 10 times. Format your responses in a JSON format. The text: "\{\}" \\ \midrule

CoLA   & Rephrase an original grammatically correct/incorrect text 10 times. Ensure that the rephrases are grammatically correct/incorrect. Format your responses like: 1. rephrase 1, 2. rephrase 2, ... , 10. rephrase 10. Original text: "\{\}". \\ 
\bottomrule
\end{tabularx}
\end{small}
\end{center}
\caption{Prompts used for generating the paraphrases for the different datasets. The MRPC and CoLA dataset use a modified prompt to guarantee that the generated paraphrases are valid. For the CoLA dataset, the prompt guarantees that the paraphrase has the same class for the sample (either grammatically acceptable or not). For the MRPC dataset, the prompt guarantees that the two sentences are paraphrased at once, reducing the number of queries needed otherwise.}
\label{tab:prompt-format}
\end{table*}

In this section, we provide further details regarding the hyperparameter setup for the individual mitigation strategies and how this affects their cost (represented as the number of steps). The hyperparameter setup for each of the methods, except for the baselines (\textit{Default} and \textit{Full}) and our method is determined using a hyperparameter search (with the same setup as for individual models, i.e., using average over 5 repeated runs on validation data).

The \textit{Default} baseline uses the setup as specified in Appendix~\ref{app:general_setup}, i.e., training for 10 epochs on 1 000 samples, with batch size of 8 and Adam optimiser. As such, the number of steps for each dataset is 1250, which we use as the normalisation number for the remaining mitigation strategies (i.e., having normalised cost value of $1$). 

The \textit{All Data} baseline uses all the labelled samples available in the dataset, with the same setup as \textit{Default} baseline. The exception are the AG News and DBPedia dataset, which are downsampled to use up to 55 000 samples. As such, the number of steps (and the cost of this baseline) is dataset dependent. For the  TREC and MRPC datasets, the cost is $5$ times the one of \textit{Default}, for the CoLA the cost is $10$, for SNIPS $11.5$ and for the AG News, SST2 and DBPedia the cost is $55$ times the \textit{Default}.

The \textit{Best Practices} mitigation strategy uses the AdamW optimiser with bias correction, warmup for 10\% of training step followed by linear scheduler and increases the number of training epoch to 20. As such, the cost is $2$ times the one in \textit{Default} baseline.

The \textit{Ensemble} mitigation strategy uses the same setup as the \textit{Default} baselines, but trains 10 models with different initialisation and data order. As such, the cost is 10 times the one in \textit{Default} baseline.

The \textit{Noise$_{Input}$} mitigation strategy introduces a Gaussian noise with the variance of 0.15 every 25 steps into the input embeddings of the model. Similarly, the \textit{Noise$_{Weights}$} mitigation strategy introduces a Gaussian noise with the variance of 0.15 (using scaling based on parameter standard deviation and number of steps) every 25 steps to all the randomly initialised parameters of the model (i.e., only the classification head or also the parameters added by PEFT method). As this does not modify the number of steps, the cost is the same as \textit{Default} baseline.

For the \textit{Stochastic Weight Averaging (SWA)} mitigation strategy we start the averaging after 25\% of training steps using the same small learning rate as used for the model (as recommended by~\citet{lu-etal-2022-improving}). As such, for the 75\% of training, we perform optimisation on 2 models and so the cost is $1.75$ time the one in \textit{Default} baseline.

The \textit{Mixout} mitigation strategy replaces the Dropout layers with mix probability of $0.9$ as recommended by~\citet{Lee2020mixout}. As it does not introduce any increase in the number of steps, the cost is the same as the one in \textit{Default} baseline.

For the \textit{Augment N} mitigation strategy, we follow the experimental setup from~\citet{cegin-etal-2023-chatgpt, cegin2024effects}. A pre-trained large language model is used to paraphrase all the training samples 10 times and the generated paraphrases are then use as additional samples for training. For paraphrasing we use the ChatGPT 3.5 (version \textit{gpt-3.5-turbo-0125}) model, with temperature value of 1 and other parameters set to default values. The prompts used to generate the paraphrases are in Table~\ref{tab:prompt-format}. For MRPC dataset, we formatted the augmented phrases in JSON for easier parsing. For CoLA dataset, we noticed that the resulting paraphrases for the grammatically incorrect sentences were in some cases (~10\% of the time) grammatically correct, meaning the label was incorrect. We did not do any additional manual filtering though (as it would be costly) and used the method as is with the all of the collected data. To reduce the required computation (and other) costs of the augmentation, we run it only a single time, asking the large language model to generate 10 paraphrases at once. When selecting what paraphrases to use for different $N$, we always select from top ones, i.e., when selecting $1$ paraphrase, we always take the first generated by the model, or when selecting $3$, we always take top three paraphrases. In addition, we increase the number of epochs to 12, as we have observed this significantly improves the performance. As the mitigation strategy introduces new samples and we increase the number of steps, the resulting cost of the strategy can be calculated as $1.2 * (N + 1)$.

Finally, for our method, we set the hyperparameters based on the hyperparameter search (Appendix~\ref{sec:hyperparameter-sensitivity}), superficially, we use $var_{noise}$ set to 0.15, $steps_{noisy}$ set to 125, $noise_{start}$ set to 30\% of training steps; $noise_{end}$ set to 60\% of the training steps; $ensemble_{start}$ set to 90\% of the training steps; ensemble size of 10; and the remaining parameters as recommended in Section~\ref{sec:DENI} (i.e., using Gaussian noise introduced only to newly initialised parameters, same number of steps for $steps_{noisy}$ and $steps_{regular}$, using scaled noise based on the parameters and also on number of already done steps for the Noisy Interpolation part). In addition, for the \textit{DENIALS} configuration, we use the \textit{Augment 1} to generate the new samples. We calculate the cost for the \textit{DENI} method in following way. For the Delayed Ensemble (DE), a single model is trained for 90\% of the training steps (cost of $0.9$) and 10 models are trained for the remaining training steps (cost of $10 * 0.1$) resulting in a cost of $1.9$. For the Noisy Interpolation part a single model is trained for 70\% of the training steps (cost of $0.7$), and the noise is added for 30\% of training steps, from which we train the ensemble for 10\% of training (cost of $1$), single model for additional 10\% of training (cost of $0.1$) and another ensemble for additional 10\% of training (cost of $1$), resulting in cost of $2.8$. The cost of the Delayed Ensemble with Noisy Interpolation (DENI) is the combination of the Noisy Interpolation ($2.8$) and the cost for the final ensemble ($0.9$) resulting in cost of $3.7$. Finally for the Delayed Ensemble with Noisy Interpolation and Augmented Labelled Samples (DENIALS) we double the number of steps and so the cost is calculated as $(N + 1) * 3.7$, resulting in cost of $7.4$.

\section{Additional Results: Relationship Between Performance, Deviation and Normalised Cost}
\label{app:cost-rel}

\begin{figure*}[!tbh]
    \centering
    \includegraphics[width=1\linewidth]{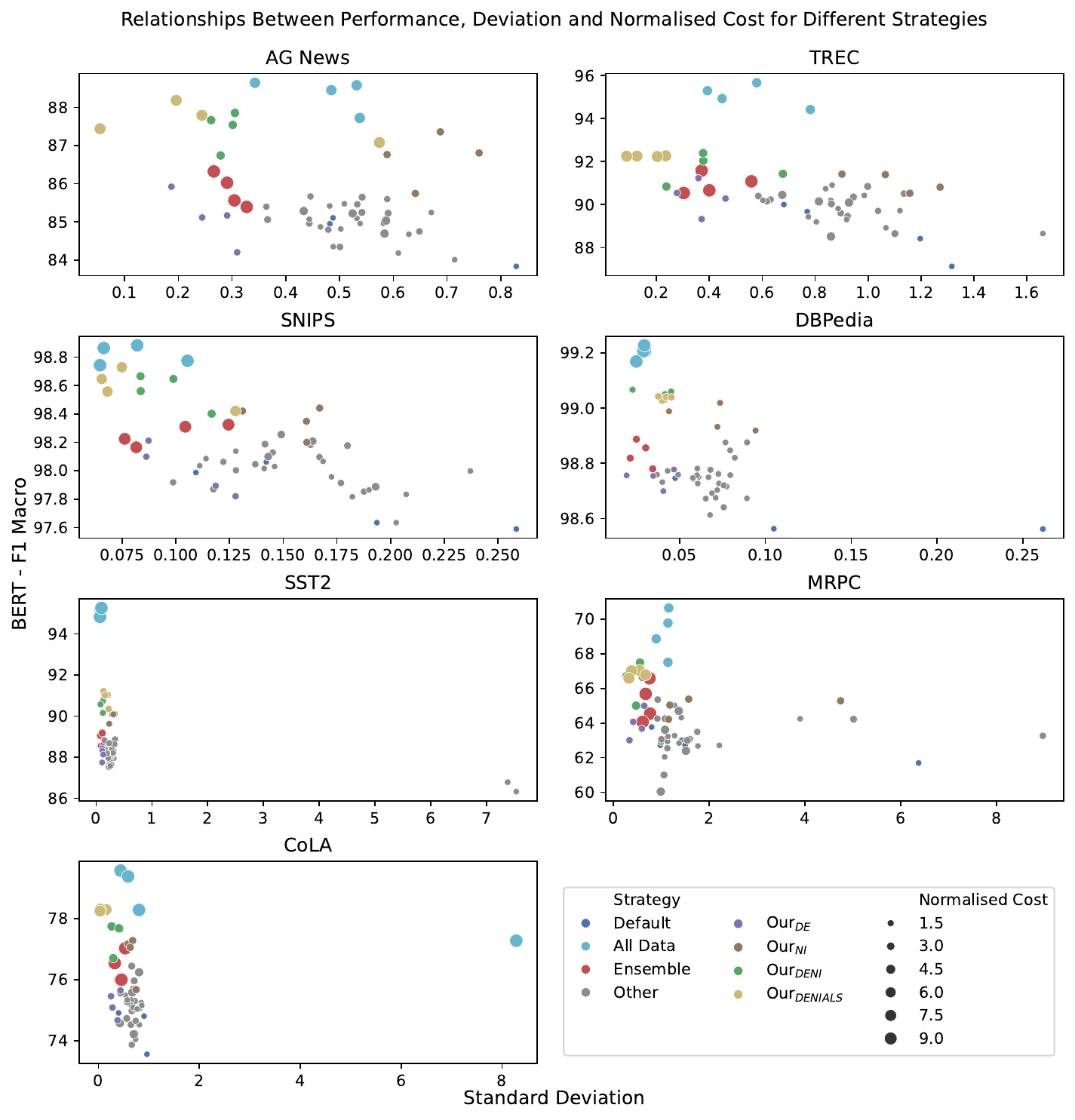}
    \caption{The relationship between performance, standard deviation and the normalised cost for the different mitigation strategies. The results are presented for the BERT model and all fine-tuning methods on all datasets. For the sake of better readability, we highlight only the most efficient mitigation strategies and baselines (highest performance and lowest standard deviation), greying out all the remaining ones. We can observe that the \textit{DENI} method provides the most effective mitigation, with the performance and standard deviation being on par or better than other strategies, while requiring only a fraction of their computation costs.}
    \label{fig:bert-bubble-cost}
\end{figure*}

In this Appendix, we provide a visualisation of the relationship between the performance, standard deviation and the normalised cost for the different mitigation strategies. The visualisation serves to allow for better evaluation and comparison for the benefit of the mitigation strategies, which also takes cost into consideration. The relationship is visualised in Figure~\ref{fig:bert-bubble-cost} for the BERT model for all fine-tuning methods across all datasets.

We observe that the mitigation strategies that provide the highest benefit, increasing performance and reducing standard deviation, also introduce increase in the computation cost. When taking the cost into consideration, the \textit{DENI} method (without the augmented labelled samples) appears the most efficient, as it provides similar (and often higher) performance, similar (and often lower) standard deviation, while requiring only half of the computation costs. Similarly, the \textit{DENI} method provides higher performance and similar (and often lower) standard deviation than the \textit{Ensemble} mitigation strategy, while requiring only a 1/3 of the computation cost. An additional interesting comparison is with the \textit{All Data} baseline, where the \textit{DENI} method, but often also \textit{Ensemble} mitigation strategy, appear close to the baseline, often outperforming it in terms standard deviation.

\section{Additional Results: \textit{Augment N} With Higher Number of Paraphrases}
\label{app:augment_full}

\begin{figure*}[t]
    \centering
    \includegraphics[width=1\linewidth]{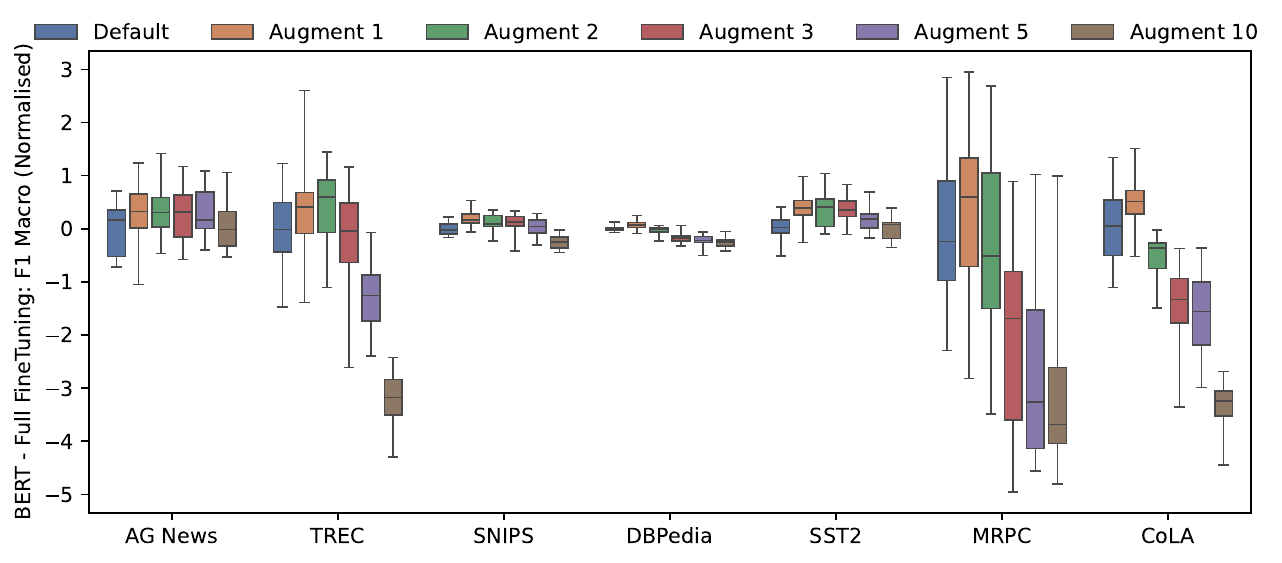}
    \caption{The performance of the Augment mitigation strategy for the different values of $N$ for BERT using full fine-tuning across all datasets. The reported performance is normalised using the mean value of the \textit{Default} baselines for each of the given datasets (i.e., subtracting the mean from all values).}
    \label{fig:bert-fft-augment}
\end{figure*}

\begin{figure*}[t]
    \centering
    \includegraphics[width=1\linewidth]{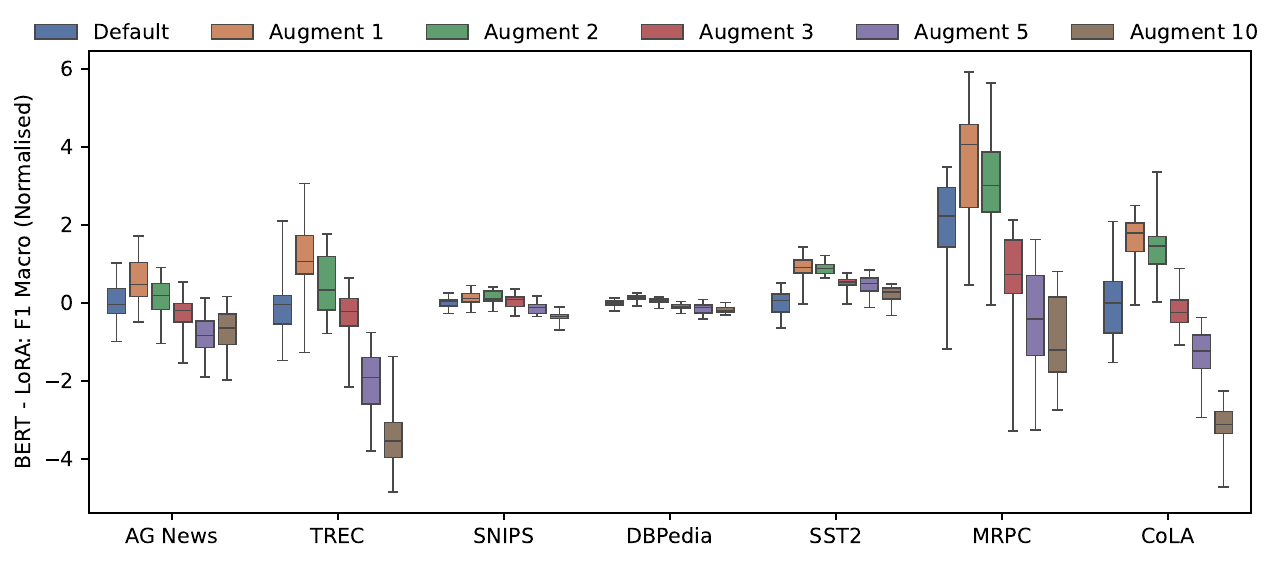}
    \caption{The performance of the Augment mitigation strategy for the different values of $N$ for BERT using LoRA across all datasets. The reported performance is normalised using the mean value of the \textit{Default} baselines for each of the given datasets (i.e., subtracting the mean from all values).}
    \label{fig:bert-lora-augment}
\end{figure*}

\begin{figure*}[t]
    \centering
    \includegraphics[width=1\linewidth]{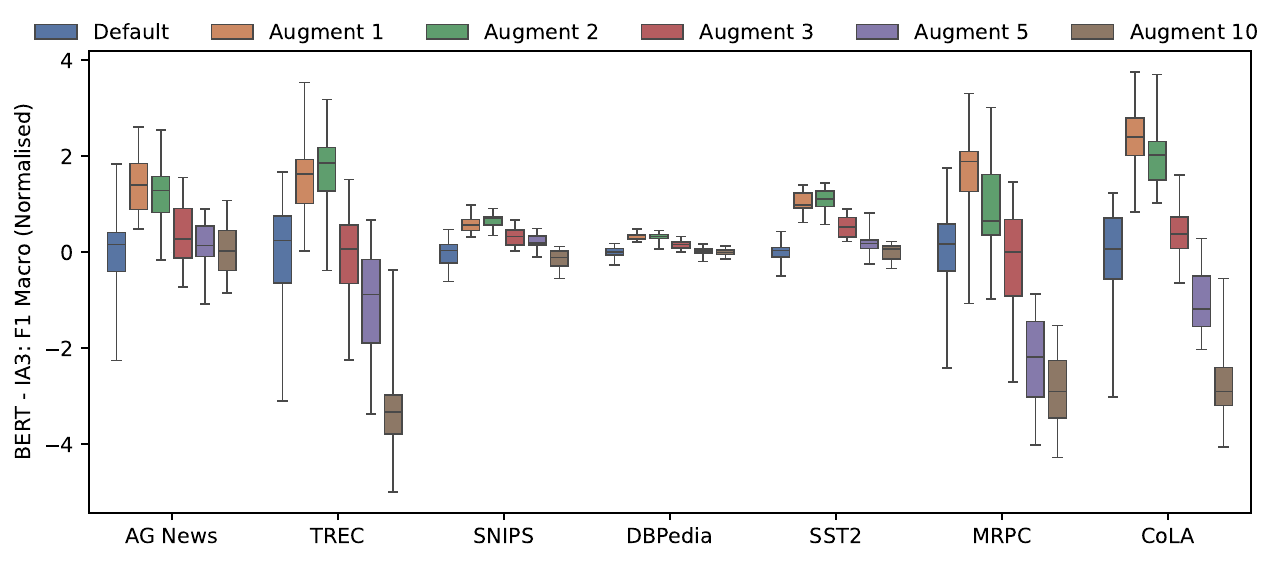}
    \caption{The performance of the Augment mitigation strategy for the different values of $N$ for BERT using IA3 across all datasets. The reported performance is normalised using the mean value of the \textit{Default} baselines for each of the given datasets (i.e., subtracting the mean from all values).}
    \label{fig:bert-ia3-augment}
\end{figure*}

\begin{figure*}[t]
    \centering
    \includegraphics[width=1\linewidth]{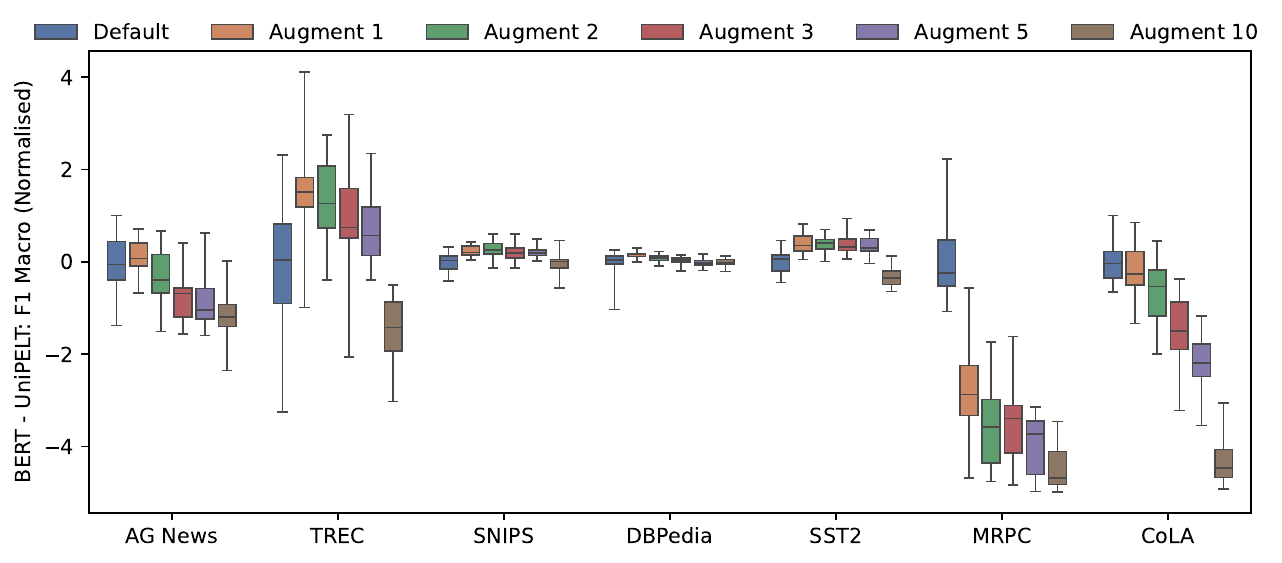}
    \caption{The performance of the Augment mitigation strategy for the different values of $N$ for BERT using UniPELT across all datasets. The reported performance is normalised using the mean value of the \textit{Default} baselines for each of the given datasets (i.e., subtracting the mean from all values).}
    \label{fig:bert-unipelt-augment}
\end{figure*}

\begin{figure*}[t]
    \centering
    \includegraphics[width=1\linewidth]{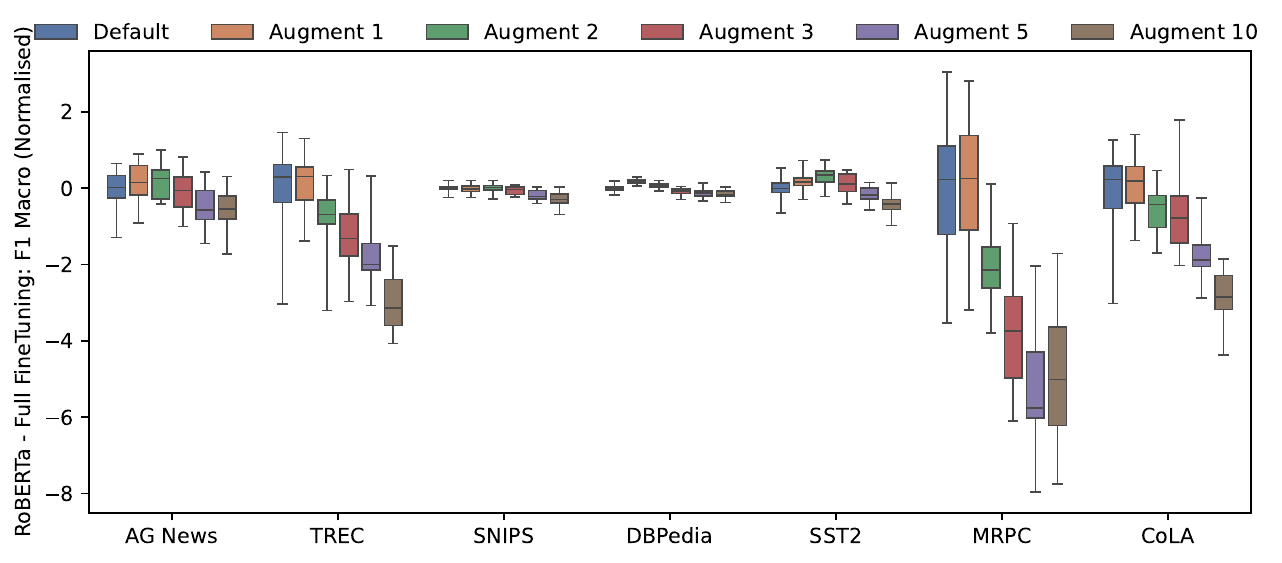}
    \caption{The performance of the Augment mitigation strategy for the different values of $N$ for RoBERTa using full fine-tuning across all datasets. The reported performance is normalised using the mean value of the \textit{Default} baselines for each of the given datasets (i.e., subtracting the mean from all values).}
    \label{fig:roberta-fft-augment}
\end{figure*}

\begin{figure*}[t]
    \centering
    \includegraphics[width=1\linewidth]{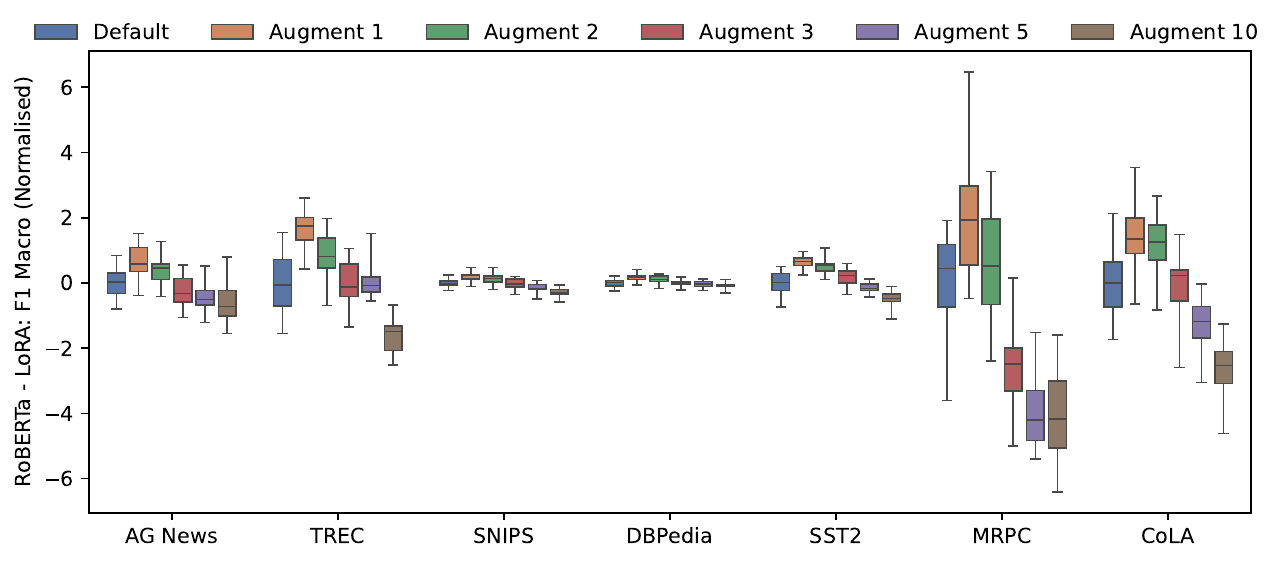}
    \caption{The performance of the Augment mitigation strategy for the different values of $N$ for RoBERTa using LoRA across all datasets. The reported performance is normalised using the mean value of the \textit{Default} baselines for each of the given datasets (i.e., subtracting the mean from all values).}
    \label{fig:roberta-lora-augment}
\end{figure*}

\begin{figure*}[t]
    \centering
    \includegraphics[width=1\linewidth]{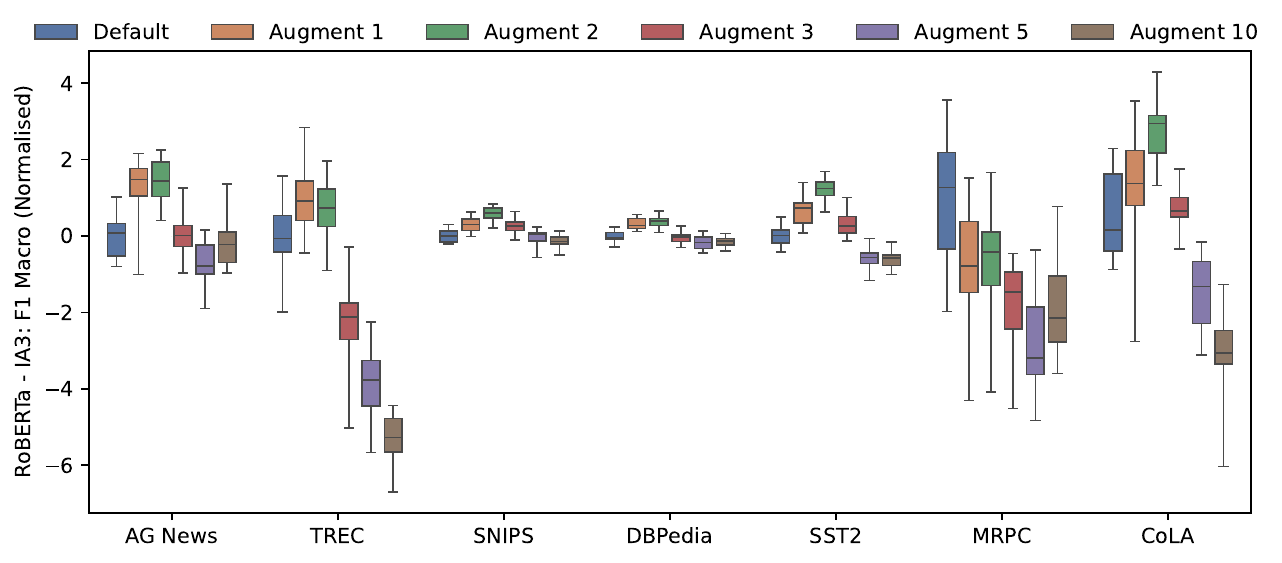}
    \caption{The performance of the Augment mitigation strategy for the different values of $N$ for RoBERTa using IA3 across all datasets. The reported performance is normalised using the mean value of the \textit{Default} baselines for each of the given datasets (i.e., subtracting the mean from all values).}
    \label{fig:roberta-ia3-augment}
\end{figure*}

\begin{figure*}[t]
    \centering
    \includegraphics[width=1\linewidth]{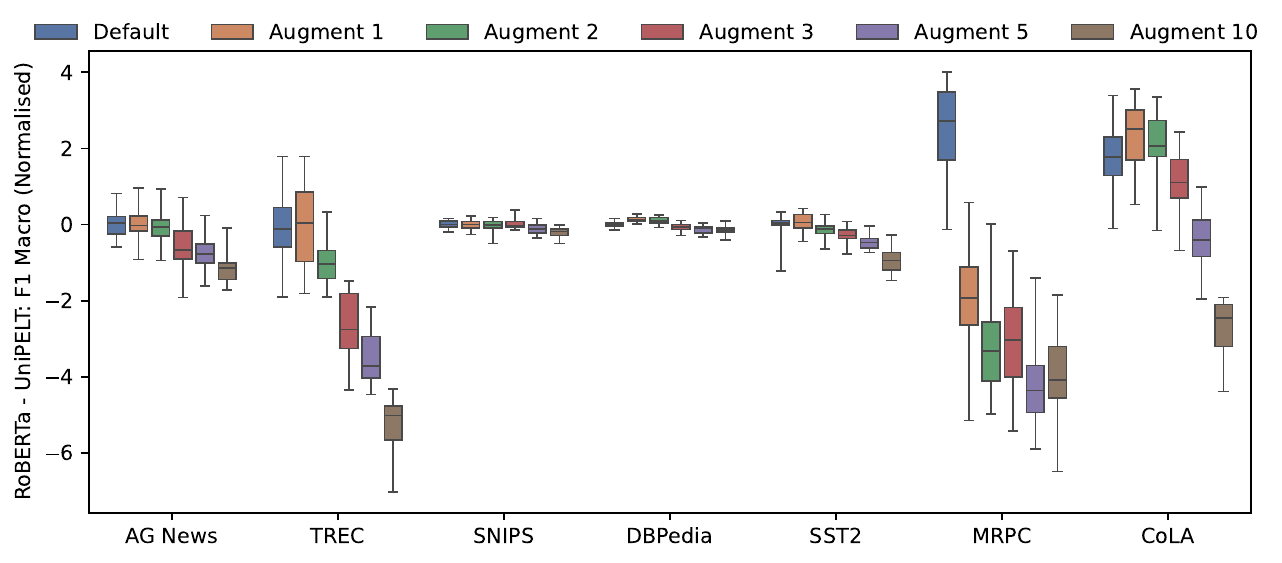}
    \caption{The performance of the Augment mitigation strategy for the different values of $N$ for RoBERTa using UniPELT across all datasets. The reported performance is normalised using the mean value of the \textit{Default} baselines for each of the given datasets (i.e., subtracting the mean from all values).}
    \label{fig:roberta-unipelt-augment}
\end{figure*}

\begin{figure*}[t]
    \centering
    \includegraphics[width=1\linewidth]{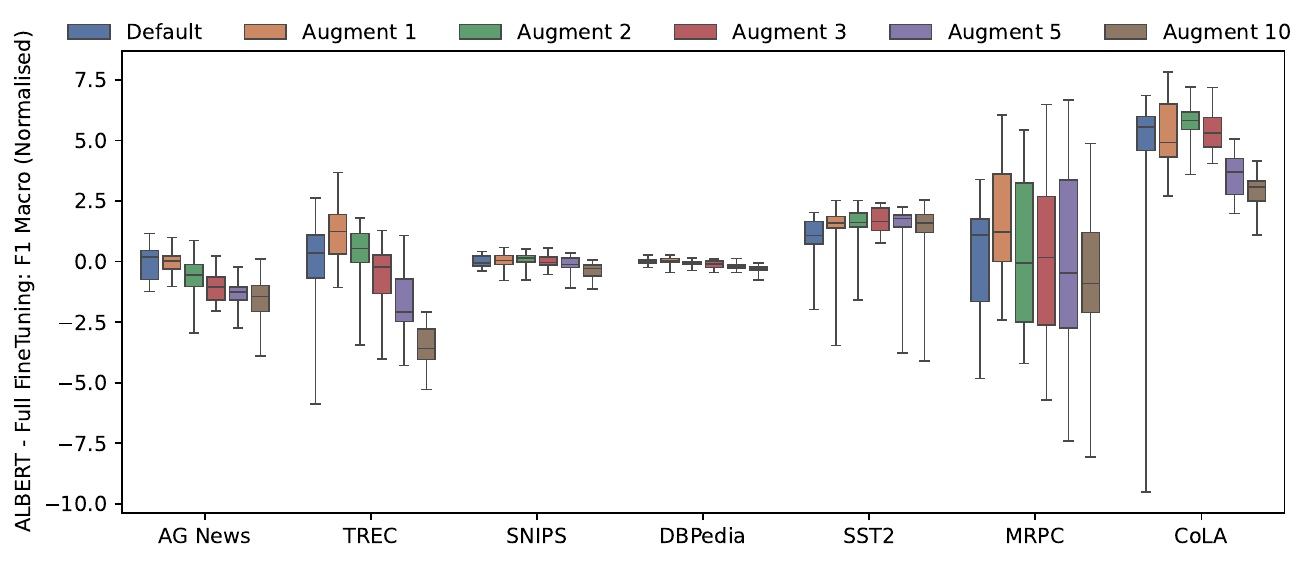}
    \caption{The performance of the Augment mitigation strategy for the different values of $N$ for ALBERT using full fine-tuning across all datasets. The reported performance is normalised using the mean value of the \textit{Default} baselines for each of the given datasets (i.e., subtracting the mean from all values).}
    \label{fig:albert-fft-augment}
\end{figure*}

\begin{figure*}[t]
    \centering
    \includegraphics[width=1\linewidth]{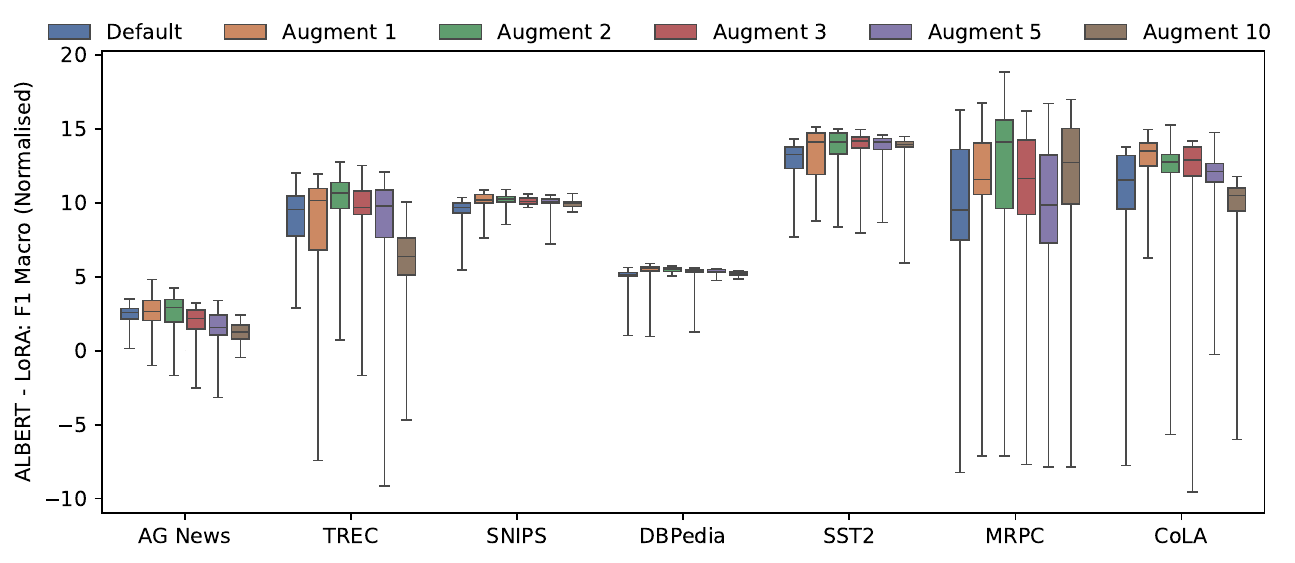}
    \caption{The performance of the Augment mitigation strategy for the different values of $N$ for ALBERT using LoRA across all datasets. The reported performance is normalised using the mean value of the \textit{Default} baselines for each of the given datasets (i.e., subtracting the mean from all values).}
    \label{fig:albert-lora-augment}
\end{figure*}

\begin{figure*}[t]
    \centering
    \includegraphics[width=1\linewidth]{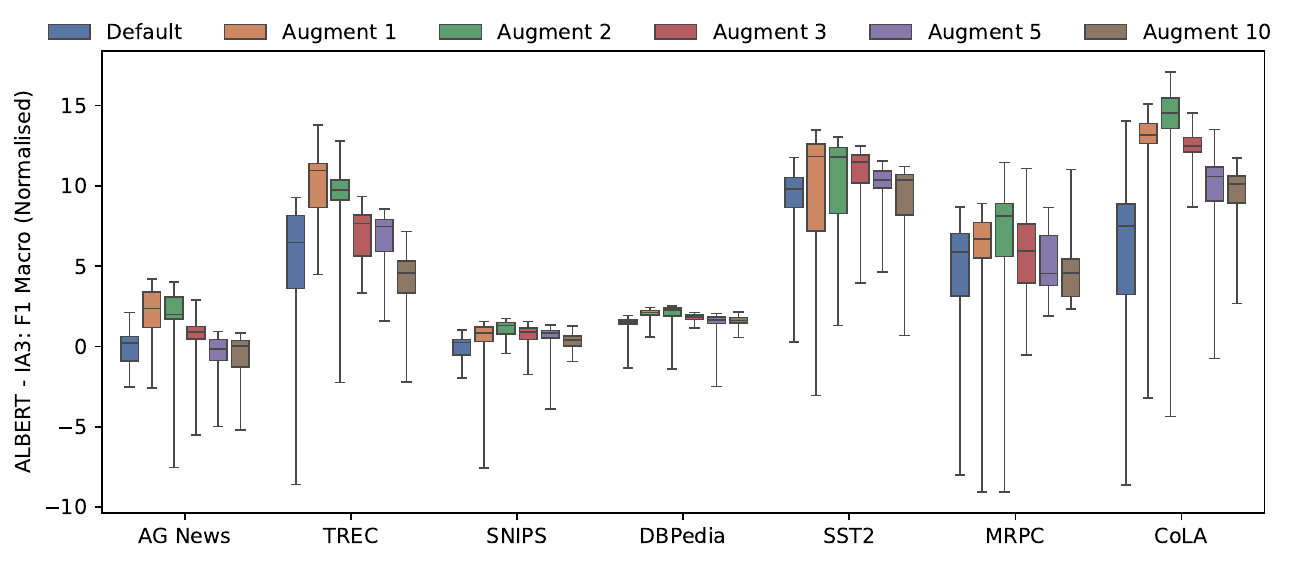}
    \caption{The performance of the Augment mitigation strategy for the different values of $N$ for ALBERT using IA3 across all datasets. The reported performance is normalised using the mean value of the \textit{Default} baselines for each of the given datasets (i.e., subtracting the mean from all values).}
    \label{fig:albert-ia3-augment}
\end{figure*}

\begin{figure*}[t]
    \centering
    \includegraphics[width=1\linewidth]{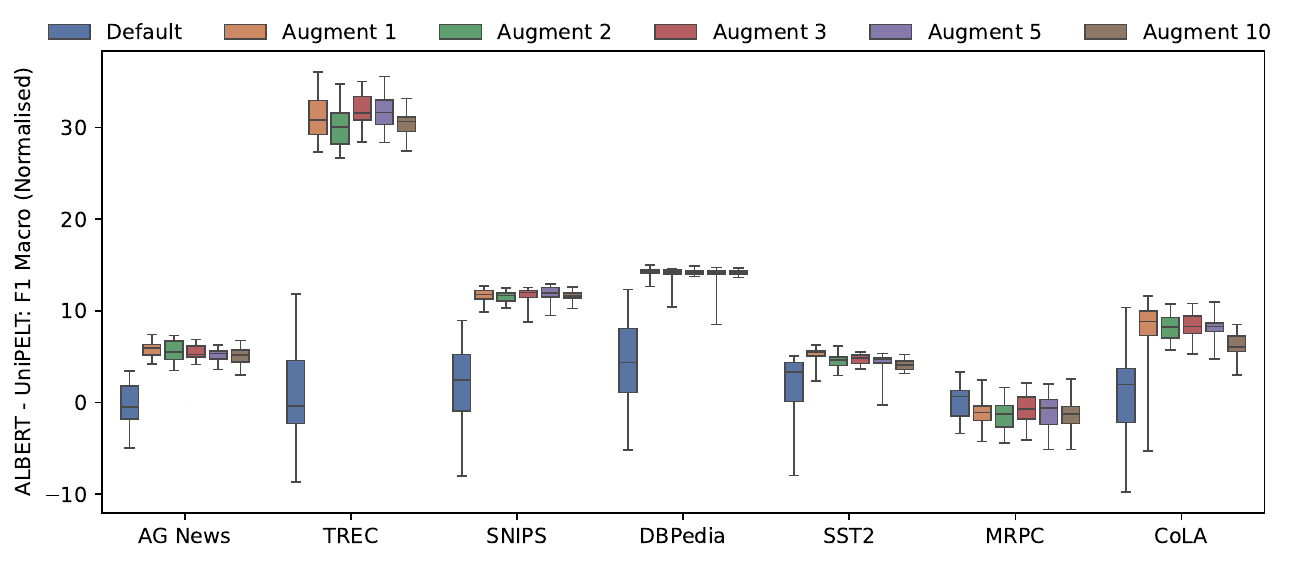}
    \caption{The performance of the Augment mitigation strategy for the different values of $N$ for ALBERT using UniPELT across all datasets. The reported performance is normalised using the mean value of the \textit{Default} baselines for each of the given datasets (i.e., subtracting the mean from all values).}
    \label{fig:albert-unipelt-augment}
\end{figure*}

In this Appendix, we provide the results for the \textit{Augment N} mitigation strategy with additional values for $N$, specifically $3$, $5$ and $10$. The results are reported for all the models and datasets and are normalised by the mean value of \textit{Default} baseline for each of the given datasets. For the BERT model, the results are reported in Figure~\ref{fig:bert-fft-augment} for full fine-tuning, Figure~\ref{fig:bert-lora-augment} for LoRA, Figure~\ref{fig:bert-ia3-augment} for IA3 and Figure~\ref{fig:bert-unipelt-augment} for UniPELT. For the RoBERTa model, the results are reported in Figure~\ref{fig:roberta-fft-augment} for full fine-tuning, Figure~\ref{fig:roberta-lora-augment} for LoRA, Figure~\ref{fig:roberta-ia3-augment} for IA3 and Figure~\ref{fig:roberta-unipelt-augment} for UniPELT. For the ALBERT model, the results are reported in Figure~\ref{fig:albert-fft-augment} for full fine-tuning, Figure~\ref{fig:albert-lora-augment} for LoRA, Figure~\ref{fig:albert-ia3-augment} for IA3 and Figure~\ref{fig:albert-unipelt-augment} for UniPELT.

For the majority of the cases, the highest performance increase is observed when using only 1 or 2 paraphrases for each of the labelled samples (the performance reported in the main content of the paper). Increasing the number of paraphrases, the performance gradually decreases, achieving lowest performance with 10 paraphrased samples. The reason for this decrease can be twofold. First, we run the paraphrasing only a single time, generating 10 paraphrases at once and selecting from the top. As the model tries to generate 10 paraphrases at once, the later ones may be significantly different, using more obscure words. As such, adding such paraphrases into the training may confuse the model as the distribution of words is significantly different (and may even represent noise for the model). Second, with the higher number of paraphrases, the model may start to lose its generalisability as it is trained on very similar samples (i.e., paraphrases) and thus start overfitting the data distribution which is not representative of the test data. However, we observe some exceptions, where using full set of paraphrases does not lower the performance.

In addition, we observe that the benefit of data augmentation is dependent on the model and the fine-tuning approach used. When limiting the number of trainable parameters, either by using the IA3 PEFT method (which trains approximately 0.5\% of the parameters) or using the ALBERT model (which has only 12M parameters), we observe a significantly higher benefit of the data augmentation. In specific cases, using the augmented data can lead to a performance higher than from any other mitigation strategy, such as using ALBERT model trained with IA3 method on CoLA dataset (see Table~\ref{tab:results-albert}).

Finally, we can observe that in specific cases, the augmented data cause significant problems to the models. Specifically, we observe a significant decrease in performance for the UniPELT method on the MRPC dataset across all models. In addition, we observe that the augmented data on MRPC dataset show higher standard deviation across all models and fine-tuning approaches. The reason for both may be due to the characteristics of the MRPC dataset -- the task of determining the semantic similarity between 2 sentences. Although we specifically design the prompt to paraphrase both sentences at once, which should keep their semantic similarity, the large language model may not produce correct paraphrases in this case (e.g., introducing a significant changes, making the sentences too dissimilar). Such augmented data may cause problems to the models and especially to UniPELT, which we observed to be the most brittle fine-tuning approach.

\clearpage
\section{Additional Results: All Datasets, Models and Fine-Tuning Approaches}
\label{app:full_results}

\begin{table*}
\begin{center}
\footnotesize
\begin{sc}
\tabcolsep=0.15cm
\begin{tabular}{lccccccc}
\toprule
BERT    & AG News       & TREC  & SNIPS & DBPedia       & SST2  & MRPC  & CoLA \\ \midrule
\multicolumn{8}{c}{Full FineTuning} \\
Default & 84.95$_{0.482}$       & 90.00$_{0.682}$       & 97.99$_{0.109}$       & 98.75$_{0.047}$       & 88.27$_{0.230}$       & 62.73$_{1.497}$       & 75.11$_{0.662}$ \\
All Data        & 88.65$_{0.343}$       & 95.66$_{0.579}$       & 98.86$_{0.066}$       & 99.21$_{0.030}$       & 95.15$_{0.078}$       & 68.86$_{0.893}$       & 79.56$_{0.440}$ \\
Best Practices  & 84.96$_{0.444}$       & 90.40$_{0.585}$       & 98.05$_{0.137}$       & 98.76$_{0.049}$       & 88.32$_{0.313}$       & 63.07$_{1.593}$       & 75.30$_{0.718}$ \\
Ensemble        & 85.56$_{0.304}$       & 90.67$_{0.400}$       & 98.22$_{0.076}$       & \underline{98.82$_{0.021}$}       & 89.05$_{0.079}$       & 65.69$_{0.674}$       & 76.54$_{0.325}$ \\
Noise$_{Input}$ & 85.07$_{0.444}$       & 89.43$_{0.775}$       & 98.09$_{0.114}$       & 98.77$_{0.043}$       & 88.25$_{0.285}$       & 62.44$_{1.499}$       & 75.05$_{0.817}$ \\
Noise$_{Weights}$       & 85.25$_{0.671}$       & 90.20$_{0.860}$       & 98.03$_{0.146}$       & 98.75$_{0.061}$       & 88.04$_{0.266}$       & 62.68$_{1.765}$       & 75.06$_{0.699}$ \\
SWA     & 85.40$_{0.364}$       & 90.24$_{0.632}$       & 98.06$_{0.122}$       & 98.76$_{0.060}$       & 88.54$_{0.211}$       & 63.50$_{1.751}$       & 75.26$_{0.572}$ \\
Mixout  & 84.96$_{0.538}$       & 90.20$_{0.603}$       & 98.03$_{0.111}$       & 98.73$_{0.040}$       & 88.39$_{0.280}$       & 63.23$_{1.137}$       & 75.33$_{0.572}$ \\
Augment 1       & 85.25$_{0.542}$       & 90.36$_{0.945}$       & 98.19$_{0.142}$       & 98.82$_{0.082}$       & 88.66$_{0.286}$       & 63.01$_{1.544}$       & 75.56$_{0.443}$ \\
Augment 2       & 85.28$_{0.433}$       & 90.45$_{0.676}$       & 98.10$_{0.143}$       & 98.72$_{0.076}$       & 88.63$_{0.327}$       & 62.40$_{1.514}$       & 74.58$_{0.426}$ \\
Our$_{DE}$      & 85.17$_{0.291}$       & 90.54$_{0.278}$       & 98.10$_{0.086}$       & \underline{98.76$_{0.019}$}       & 88.43$_{0.107}$       & 63.68$_{0.594}$       & 75.09$_{0.281}$  \\
Our$_{NI}$      & 86.81$_{0.760}$       & 91.42$_{0.901}$       & 98.42$_{0.131}$       & 98.99$_{0.044}$       & 90.08$_{0.284}$       & 65.38$_{1.571}$       & 77.16$_{0.586}$   \\
Our$_{DENI}$    & 87.67$_{0.261}$       & 92.04$_{0.377}$       & 98.65$_{0.099}$       & \textbf{99.07$_{0.023}$}       & 90.82$_{0.129}$       & 66.66$_{0.613}$       & 77.74$_{0.263}$  \\
Our$_{DENIALS}$ & \underline{\textbf{88.18$_{0.196}$}}       & \underline{\textbf{92.27$_{0.236}$}}       & \underline{\textbf{98.73$_{0.075}$}}       & 99.03$_{0.040}$       & \textbf{91.22$_{0.132}$}       & \underline{\textbf{67.04$_{0.544}$}}       & \underline{\textbf{78.29$_{0.153}$}} \\ \midrule
\multicolumn{8}{c}{LoRA} \\
Default & 85.10$_{0.488}$       & 89.68$_{0.770}$       & 98.06$_{0.142}$       & 98.72$_{0.077}$       & 87.95$_{0.315}$       & 61.70$_{6.377}$       & 74.80$_{0.907}$ \\
All Data        & 88.45$_{0.485}$       & 95.29$_{0.393}$       & 98.88$_{0.082}$       & 99.21$_{0.029}$       & 95.16$_{0.086}$       & 70.64$_{1.156}$       & 77.27$_{8.283}$ \\
Best Practices  & 85.46$_{0.533}$       & 90.51$_{1.135}$       & 98.10$_{0.167}$       & 98.74$_{0.058}$       & 88.13$_{0.263}$       & 65.36$_{0.925}$       & 74.97$_{0.662}$ \\
Ensemble        & 86.32$_{0.266}$       & 91.58$_{0.371}$       & 98.32$_{0.125}$       & \underline{98.86$_{0.030}$}       & \underline{89.17$_{0.109}$}       & 66.59$_{0.753}$       & 77.02$_{0.533}$ \\
Noise$_{Input}$ & 85.42$_{0.482}$       & 90.42$_{0.986}$       & 98.07$_{0.169}$       & 98.76$_{0.037}$       & 88.16$_{0.285}$       & 65.02$_{1.277}$       & 75.48$_{0.545}$ \\
Noise$_{Weights}$       & 85.09$_{0.532}$       & 90.74$_{0.840}$       & 98.02$_{0.141}$       & 98.73$_{0.061}$       & 88.14$_{0.280}$       & 64.25$_{3.900}$       & 75.25$_{0.851}$ \\
SWA     & 85.67$_{0.446}$       & 90.33$_{0.948}$       & 98.13$_{0.145}$       & 98.75$_{0.058}$       & 86.78$_{7.381}$       & 63.26$_{8.973}$       & 74.65$_{0.739}$ \\
Mixout  & 85.47$_{0.509}$       & 90.91$_{0.864}$       & 98.14$_{0.128}$       & 98.68$_{0.071}$       & 88.21$_{0.306}$       & 64.31$_{1.418}$       & 75.15$_{0.860}$ \\
Augment 1       & 85.65$_{0.542}$       & 90.84$_{0.998}$       & 98.18$_{0.163}$       & 98.85$_{0.079}$       & 88.87$_{0.339}$       & 64.23$_{5.017}$       & 76.44$_{0.662}$ \\
Augment 2       & 85.22$_{0.524}$       & 90.15$_{0.814}$       & 98.21$_{0.164}$       & 98.78$_{0.068}$       & 88.82$_{0.149}$       & 64.70$_{1.366}$       & 76.24$_{0.808}$ \\
Our$_{DE}$      & \underline{85.92$_{0.187}$}       & 91.23$_{0.359}$       & 98.21$_{0.087}$       & 98.75$_{0.035}$       & 88.30$_{0.111}$       & 65.00$_{0.644}$       & 75.65$_{0.439}$ \\
Our$_{NI}$      & 87.36$_{0.688}$       & 91.39$_{1.065}$       & 98.44$_{0.167}$       & 98.93$_{0.072}$       & 90.10$_{0.332}$       & 65.29$_{4.747}$       & 77.28$_{0.680}$ \\
Our$_{DENI}$    & \textbf{87.86$_{0.305}$}       & \textbf{92.39$_{0.377}$}       & \textbf{98.67$_{0.084}$}       & \textbf{99.05$_{0.041}$}       & 90.74$_{0.122}$       & \underline{\textbf{67.49$_{0.554}$}}       & 77.66$_{0.425}$ \\
Our$_{DENIALS}$ & 87.79$_{0.244}$       & \underline{92.24$_{0.204}$}       & \underline{98.65$_{0.065}$}       & 99.04$_{0.038}$       & \textbf{91.03$_{0.206}$}       & 66.79$_{0.671}$       & \underline{\textbf{78.24$_{0.041}$}} \\ \midrule
\multicolumn{8}{c}{IA3} \\
Default & 83.83$_{0.829}$       & 88.42$_{1.197}$       & 97.59$_{0.259}$       & 98.56$_{0.105}$       & 87.60$_{0.216}$       & 62.72$_{0.979}$       & 73.56$_{0.963}$ \\
All Data        & 87.72$_{0.538}$       & 94.93$_{0.449}$       & 98.74$_{0.065}$       & 99.17$_{0.025}$       & 94.83$_{0.069}$       & 67.51$_{1.139}$       & 78.28$_{0.806}$ \\
Best Practices  & 84.34$_{0.501}$       & 89.47$_{0.923}$       & 97.85$_{0.188}$       & 98.67$_{0.065}$       & 87.58$_{0.225}$       & 62.92$_{0.994}$       & 74.53$_{0.646}$ \\
Ensemble        & 85.39$_{0.327}$       & 90.54$_{0.304}$       & \underline{98.16$_{0.082}$}       & \underline{98.78$_{0.034}$}       & \underline{88.55$_{0.092}$}       & 64.54$_{0.767}$       & 75.99$_{0.440}$ \\
Noise$_{Input}$ & 84.18$_{0.609}$       & 89.30$_{0.919}$       & 97.83$_{0.207}$       & 98.61$_{0.068}$       & 87.51$_{0.223}$       & 62.05$_{1.070}$       & 74.52$_{0.812}$ \\
Noise$_{Weights}$       & 84.35$_{0.488}$       & 89.20$_{0.805}$       & 97.82$_{0.182}$       & 98.67$_{0.089}$       & 87.63$_{0.277}$       & 63.02$_{1.426}$       & 74.26$_{0.739}$ \\
SWA     & 84.79$_{0.479}$       & 90.15$_{0.618}$       & 97.92$_{0.099}$       & 98.78$_{0.060}$       & 88.19$_{0.196}$       & 62.54$_{1.127}$       & 73.87$_{0.663}$ \\
Mixout  & 84.01$_{0.714}$       & 88.66$_{1.660}$       & 97.63$_{0.203}$       & 98.70$_{0.072}$       & 87.56$_{0.253}$       & 62.71$_{2.212}$       & 74.05$_{0.738}$ \\
Augment 1       & 85.23$_{0.590}$       & 90.03$_{0.861}$       & 98.18$_{0.180}$       & 98.88$_{0.077}$       & 88.65$_{0.229}$       & 64.25$_{1.079}$       & 75.96$_{0.710}$ \\
Augment 2       & 85.03$_{0.587}$       & 90.10$_{0.928}$       & 98.25$_{0.149}$       & 98.88$_{0.089}$       & 88.68$_{0.234}$       & 63.61$_{1.078}$       & 75.58$_{0.672}$ \\
Our$_{DE}$      & 84.20$_{0.309}$       & 89.33$_{0.371}$       & 97.82$_{0.128}$       & 98.70$_{0.040}$       & 87.74$_{0.107}$       & \underline{63.01$_{0.335}$}       & 74.67$_{0.381}$ \\
Our$_{NI}$      & 85.74$_{0.641}$       & 90.53$_{1.158}$       & 98.20$_{0.161}$       & 98.92$_{0.094}$       & 89.62$_{0.233}$       & 64.22$_{1.152}$       & 75.67$_{0.749}$ \\
Our$_{DENI}$    & \underline{86.74$_{0.279}$}       & 90.84$_{0.238}$       & 98.40$_{0.117}$       & 99.03$_{0.041}$       & 90.15$_{0.121}$       & 65.00$_{0.473}$       & 76.70$_{0.295}$ \\
Our$_{DENIALS}$ & \textbf{87.08$_{0.574}$}       & \underline{\textbf{92.26$_{0.127}$}}       & \textbf{98.42$_{0.128}$}       & \textbf{99.04$_{0.042}$}       & \textbf{90.35$_{0.224}$}       & \textbf{67.02$_{0.378}$}       & \underline{\textbf{78.30$_{0.037}$}} \\ \midrule
\multicolumn{8}{c}{UniPELT} \\
Default & 84.97$_{0.582}$       & 87.13$_{1.316}$       & 97.63$_{0.194}$       & 98.56$_{0.262}$       & 87.82$_{0.233}$       & 63.77$_{0.798}$       & 74.91$_{0.403}$ \\
All Data        & 88.58$_{0.532}$       & 94.42$_{0.782}$       & 98.77$_{0.105}$       & 99.23$_{0.029}$       & 95.26$_{0.095}$       & 69.77$_{1.139}$       & 79.37$_{0.590}$ \\
Best Practices  & 84.75$_{0.649}$       & 89.61$_{0.897}$       & 97.91$_{0.177}$       & 98.76$_{0.073}$       & 88.01$_{0.285}$       & 63.06$_{1.003}$       & 75.32$_{0.642}$ \\
Ensemble        & 86.02$_{0.291}$       & 91.08$_{0.559}$       & 98.31$_{0.104}$       & \underline{98.89$_{0.025}$}       & 88.63$_{0.141}$       & 64.07$_{0.610}$       & 76.00$_{0.458}$ \\
Noise$_{Input}$ & 84.67$_{0.629}$       & 88.93$_{1.068}$       & 97.86$_{0.190}$       & 98.75$_{0.067}$       & 87.82$_{0.247}$       & 62.94$_{1.128}$       & 75.16$_{0.652}$ \\
Noise$_{Weights}$       & 84.87$_{0.464}$       & 89.72$_{1.120}$       & 98.00$_{0.237}$       & 98.73$_{0.073}$       & 86.32$_{7.537}$       & 63.27$_{1.278}$       & 75.09$_{0.687}$ \\
SWA     & 85.60$_{0.589}$       & 89.71$_{1.038}$       & 98.00$_{0.128}$       & 98.76$_{0.080}$       & 88.37$_{0.163}$       & 64.26$_{0.919}$       & 75.69$_{0.685}$ \\
Mixout  & 84.81$_{0.502}$       & 89.82$_{0.887}$       & 97.96$_{0.173}$       & 98.69$_{0.069}$       & 87.95$_{0.217}$       & 62.84$_{1.371}$       & 75.04$_{0.770}$ \\
Augment 1       & 85.06$_{0.366}$       & 88.65$_{1.101}$       & 97.87$_{0.118}$       & 98.72$_{0.077}$       & 88.21$_{0.200}$       & 61.00$_{1.055}$       & 74.73$_{0.564}$ \\
Augment 2       & 84.69$_{0.584}$       & 88.52$_{0.860}$       & 97.89$_{0.193}$       & 98.64$_{0.076}$       & 88.18$_{0.192}$       & 60.04$_{0.990}$       & 74.22$_{0.702}$ \\
Our$_{DE}$      & 85.11$_{0.244}$       & 90.28$_{0.461}$       & 97.89$_{0.119}$       & 98.78$_{0.047}$       & 88.13$_{0.128}$       & 64.07$_{0.412}$       & 75.46$_{0.250}$ \\
Our$_{NI}$      & 86.76$_{0.588}$       & 90.81$_{1.272}$       & 98.35$_{0.161}$       & 99.02$_{0.074}$       & 90.08$_{0.309}$       & 65.04$_{1.179}$       & 77.05$_{0.631}$ \\
Our$_{DENI}$    & \textbf{87.54$_{0.301}$}       & 91.43$_{0.678}$       & \textbf{98.56$_{0.084}$}       & \textbf{99.06$_{0.045}$}       & \underline{90.57$_{0.078}$}       & \underline{\textbf{66.72$_{0.275}$}}       & 77.68$_{0.409}$ \\
Our$_{DENIALS}$ & \underline{87.44$_{0.054}$}       & \underline{\textbf{92.25$_{0.088}$}}       & \underline{\textbf{98.56$_{0.068}$}}       & 99.04$_{0.045}$       & \textbf{91.01$_{0.157}$}       & 66.59$_{0.323}$       & \underline{\textbf{78.24$_{0.033}$}} \\
\bottomrule
\end{tabular}
\end{sc}
\end{center}
\caption{Comparison of the DENI method with existing mitigation strategies and baselines on full fine-tuning, LoRA, IA3 and UniPELT using BERT model. The comparison is done in terms of overall performance and deviation. The highest performance for each fine-tuning method is in \textbf{bold} and lowest deviation is \underline{underlined} (not considering \textit{All Data} baseline).}
\label{tab:results-bert}
\end{table*}

\begin{figure*}
    \centering
    \includegraphics[width=1\linewidth]{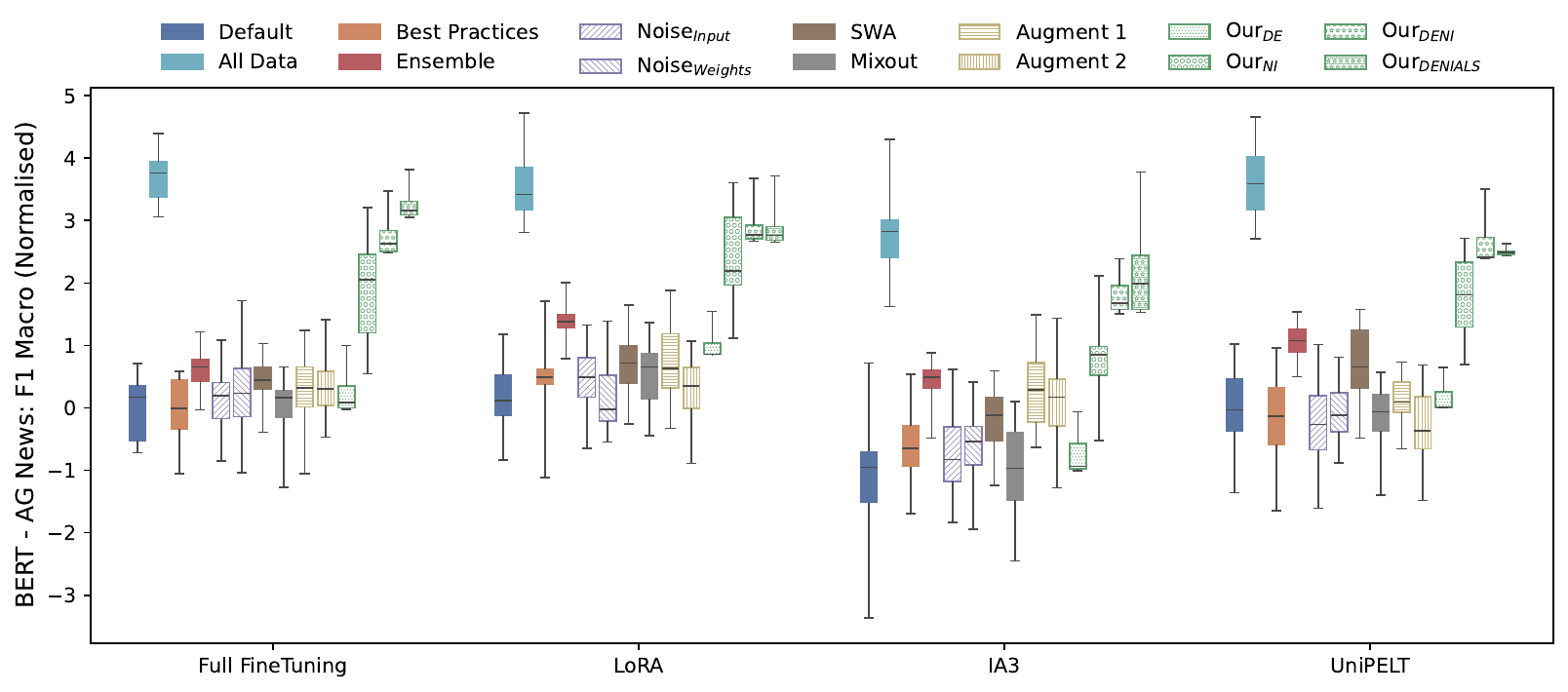}
    \caption{Benefit of mitigation strategies for the different fine-tuning methods using BERT on AG News dataset. The benefit is calculated as difference to the mean performance of the \textit{Default} baseline when using full fine-tuning. The different mitigation strategies are beneficial for all fine-tuning methods, but with different overall benefit (e.g., \textit{Augment} on IA3).}
    \label{fig:agnews-boxplot}
\end{figure*}

\begin{figure*}
    \centering
    \includegraphics[width=1\linewidth]{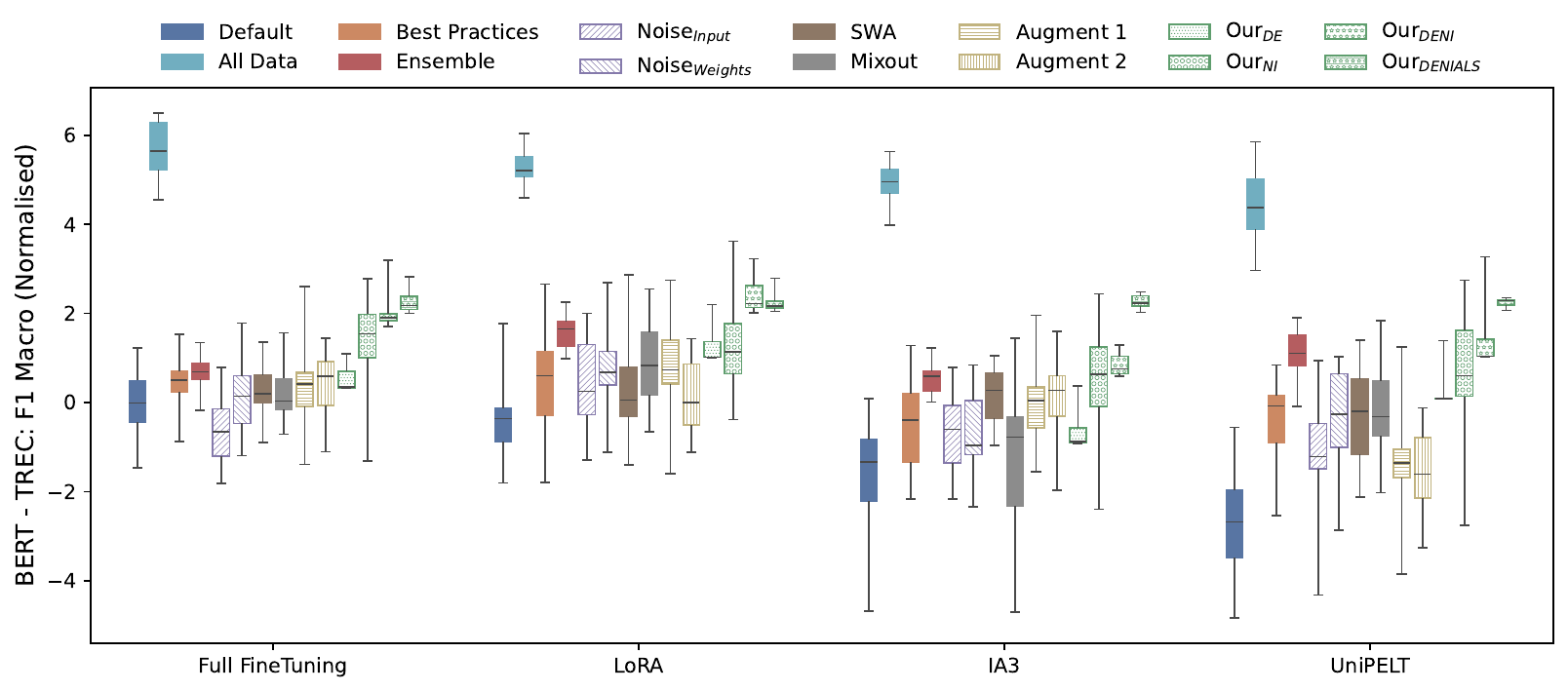}
    \caption{Benefit of mitigation strategies for the different fine-tuning methods using BERT on TREC dataset. The benefit is calculated as difference to the mean performance of the \textit{Default} baseline when using full fine-tuning. The different mitigation strategies are beneficial for all fine-tuning methods, but with different overall benefit (e.g., \textit{Augment} on IA3).}
    \label{fig:trec-boxplot}
\end{figure*}

\begin{figure*}
    \centering
    \includegraphics[width=1\linewidth]{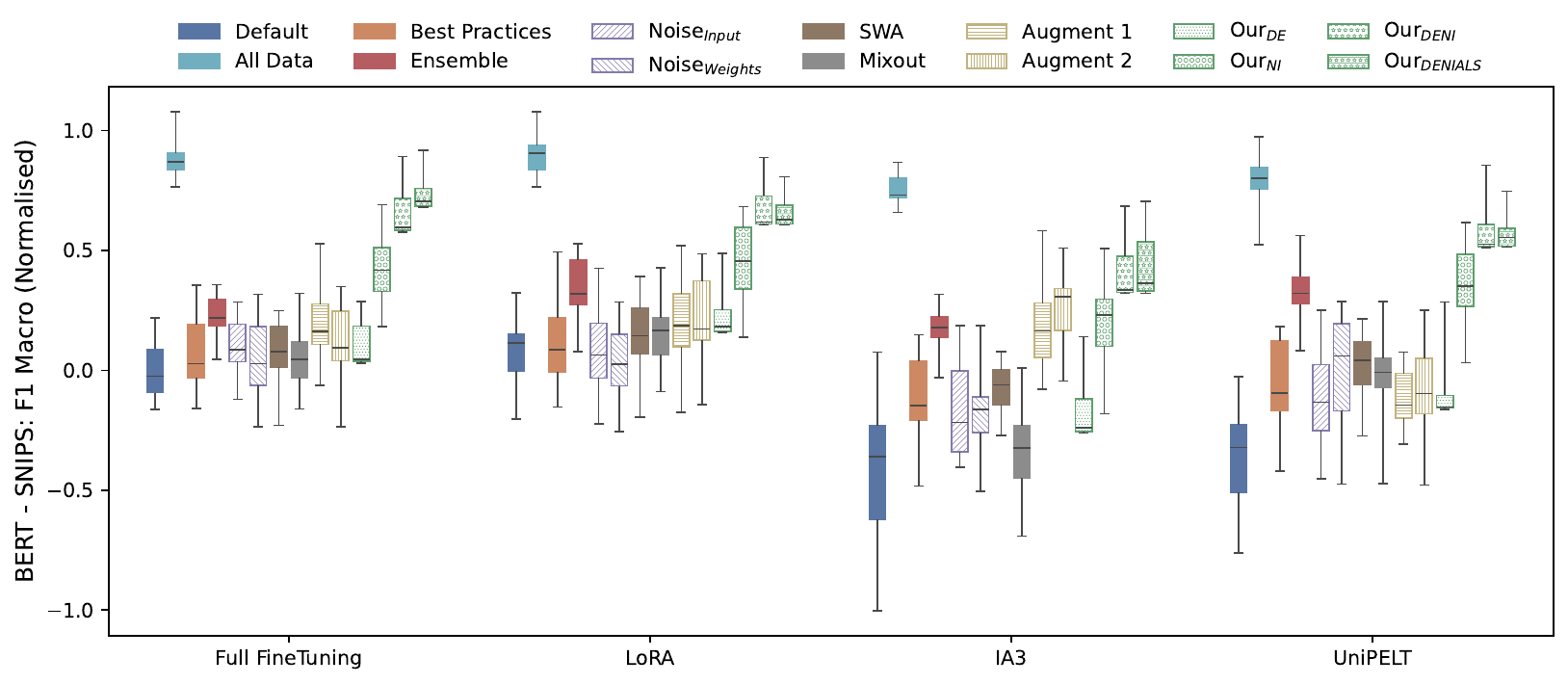}
    \caption{Benefit of mitigation strategies for the different fine-tuning methods using BERT on SNIPS dataset. The benefit is calculated as difference to the mean performance of the \textit{Default} baseline when using full fine-tuning. The different mitigation strategies are beneficial for all fine-tuning methods, but with different overall benefit (e.g., \textit{Augment} on IA3).}
    \label{fig:snips-boxplot}
\end{figure*}

\begin{figure*}
    \centering
    \includegraphics[width=1\linewidth]{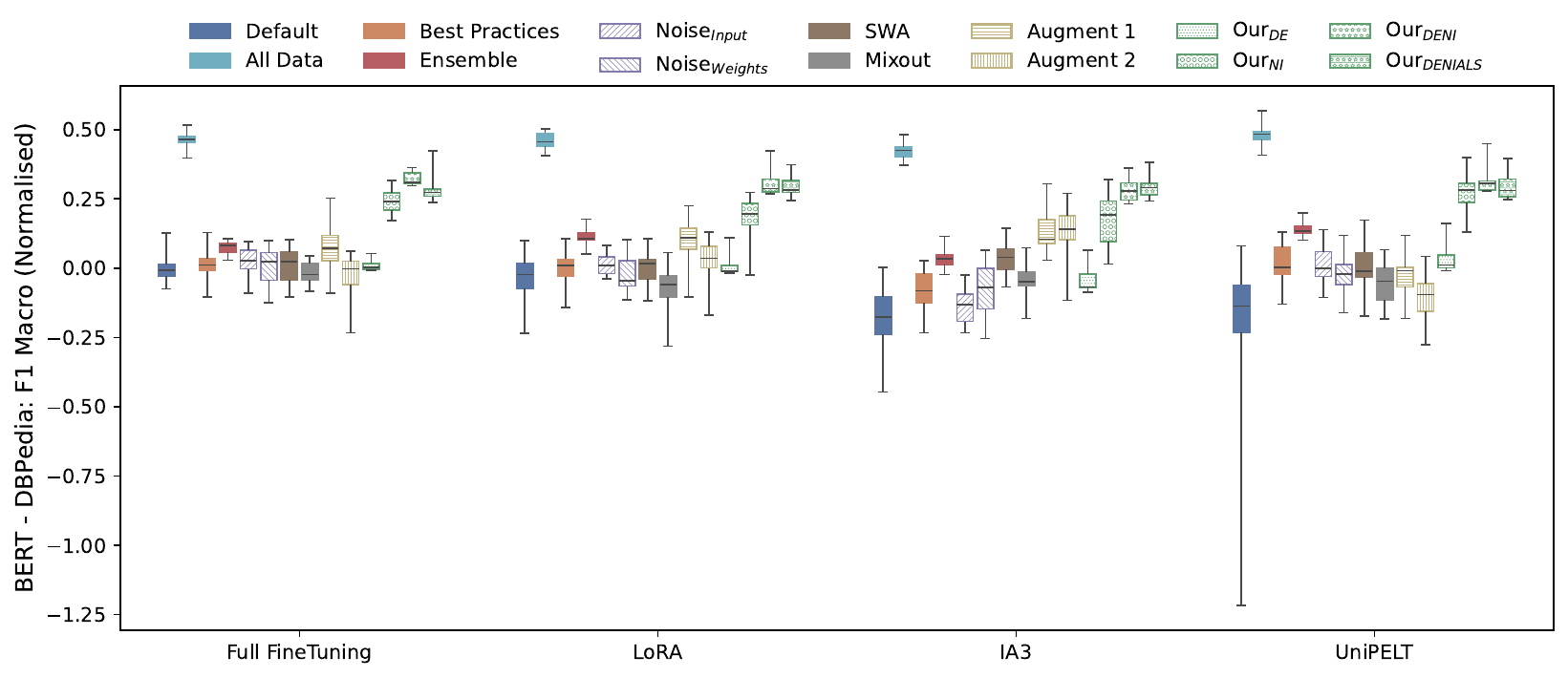}
    \caption{Benefit of mitigation strategies for the different fine-tuning methods using BERT on DBPedia dataset. The benefit is calculated as difference to the mean performance of the \textit{Default} baseline when using full fine-tuning. The different mitigation strategies are beneficial for all fine-tuning methods, but with different overall benefit (e.g., \textit{Augment} on IA3).}
    \label{fig:dbpedia-boxplot}
\end{figure*}

\begin{figure*}
    \centering
    \includegraphics[width=1\linewidth]{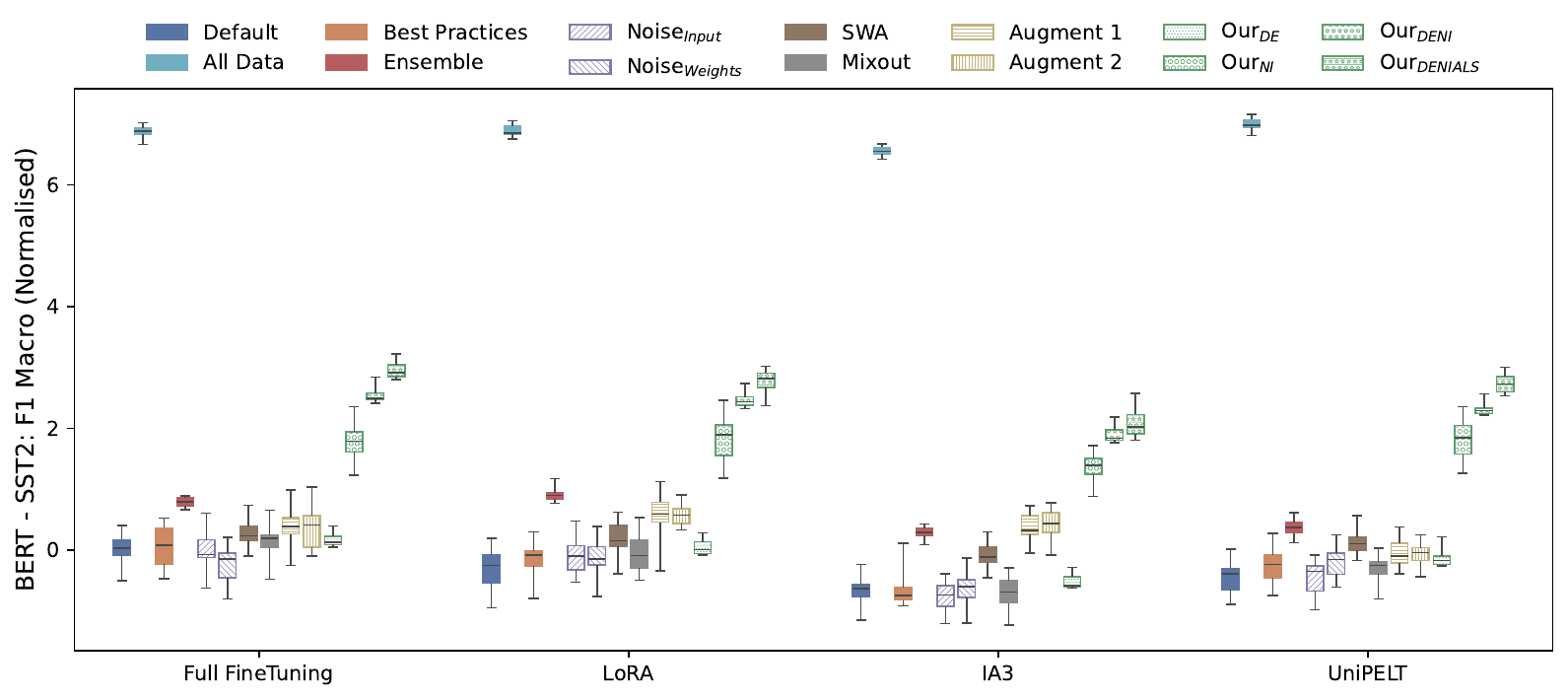}
    \caption{Benefit of mitigation strategies for the different fine-tuning methods using BERT on SST2 dataset. The benefit is calculated as difference to the mean performance of the \textit{Default} baseline when using full fine-tuning. The different mitigation strategies are beneficial for all fine-tuning methods, but with different overall benefit (e.g., \textit{Augment} on IA3).}
    \label{fig:sst2-boxplot}
\end{figure*}

\begin{figure*}
    \centering
    \includegraphics[width=1\linewidth]{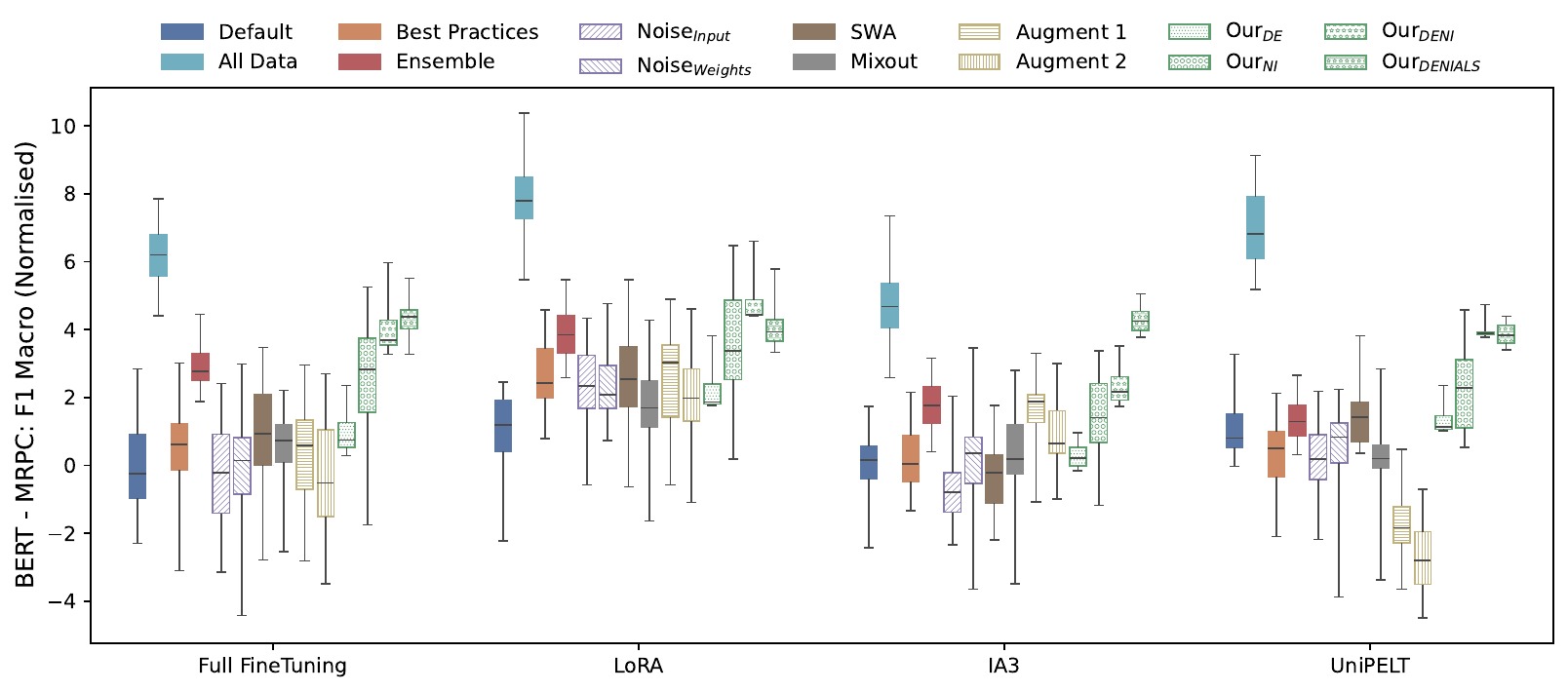}
    \caption{Benefit of mitigation strategies for the different fine-tuning methods using BERT on MRPC dataset. The benefit is calculated as difference to the mean performance of the \textit{Default} baseline when using full fine-tuning. The different mitigation strategies are beneficial for all fine-tuning methods, but with different overall benefit (e.g., \textit{Augment} on IA3).}
    \label{fig:mrpc-boxplot}
\end{figure*}

\begin{figure*}
    \centering
    \includegraphics[width=1\linewidth]{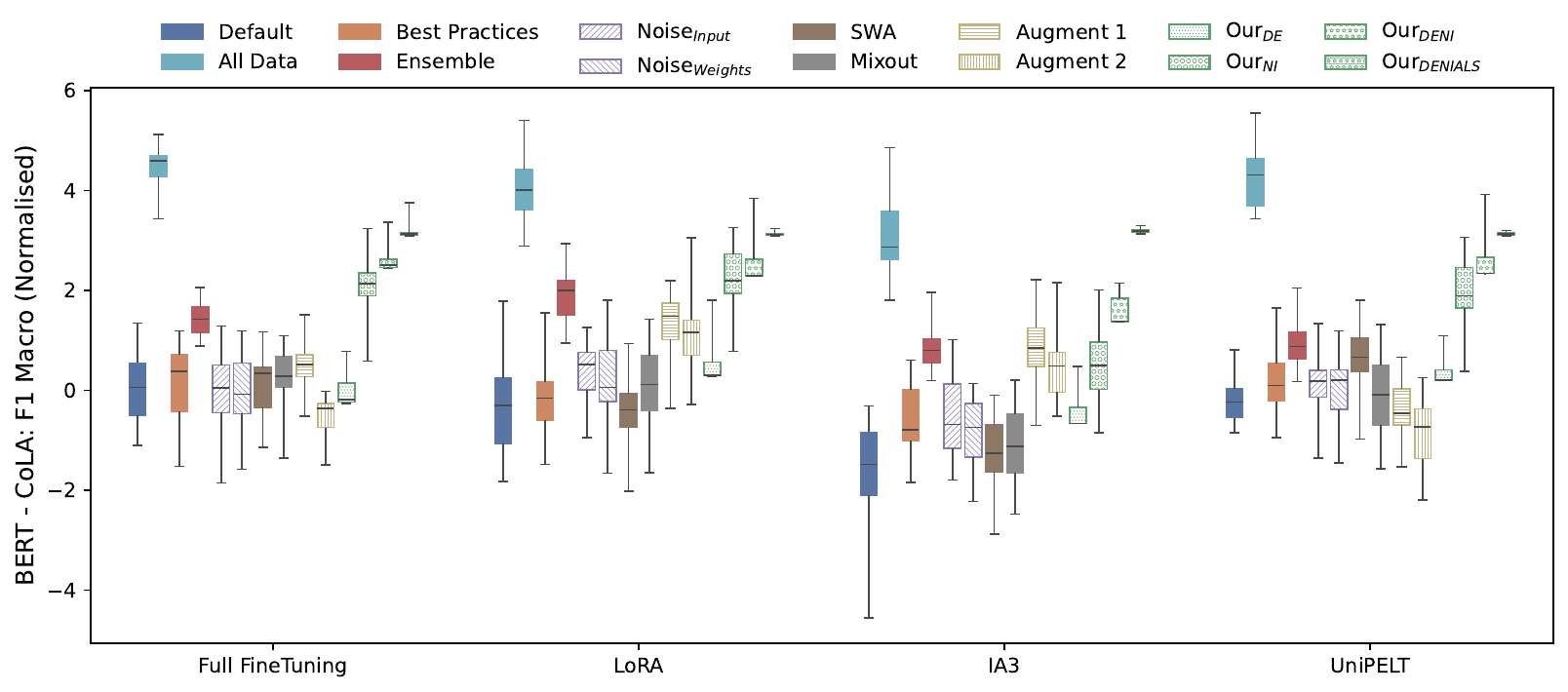}
    \caption{Benefit of mitigation strategies for the different fine-tuning methods using BERT on CoLA dataset. The benefit is calculated as difference to the mean performance of the \textit{Default} baseline when using full fine-tuning. The different mitigation strategies are beneficial for all fine-tuning methods, but with different overall benefit (e.g., \textit{Augment} on IA3).}
    \label{fig:cola-boxplot}
\end{figure*}

\begin{figure*}
    \centering
    \includegraphics[width=1\linewidth]{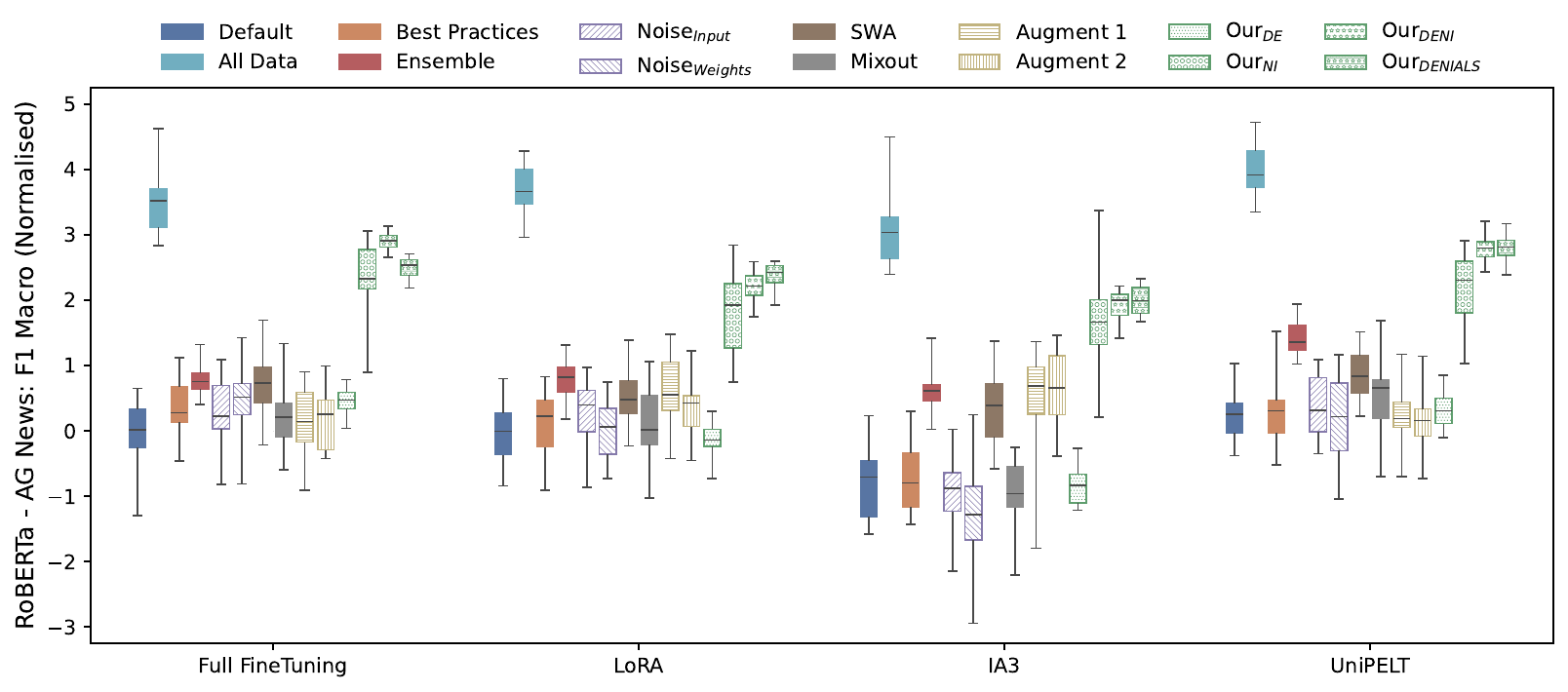}
    \caption{Benefit of mitigation strategies for the different fine-tuning methods using RoBERTa on AG News dataset. The benefit is calculated as difference to the mean performance of the \textit{Default} baseline when using full fine-tuning. The different mitigation strategies are beneficial for all fine-tuning methods, but with different overall benefit (e.g., \textit{Augment} on IA3).}
    \label{fig:roberta-agnews-boxplot}
\end{figure*}

\begin{figure*}
    \centering
    \includegraphics[width=1\linewidth]{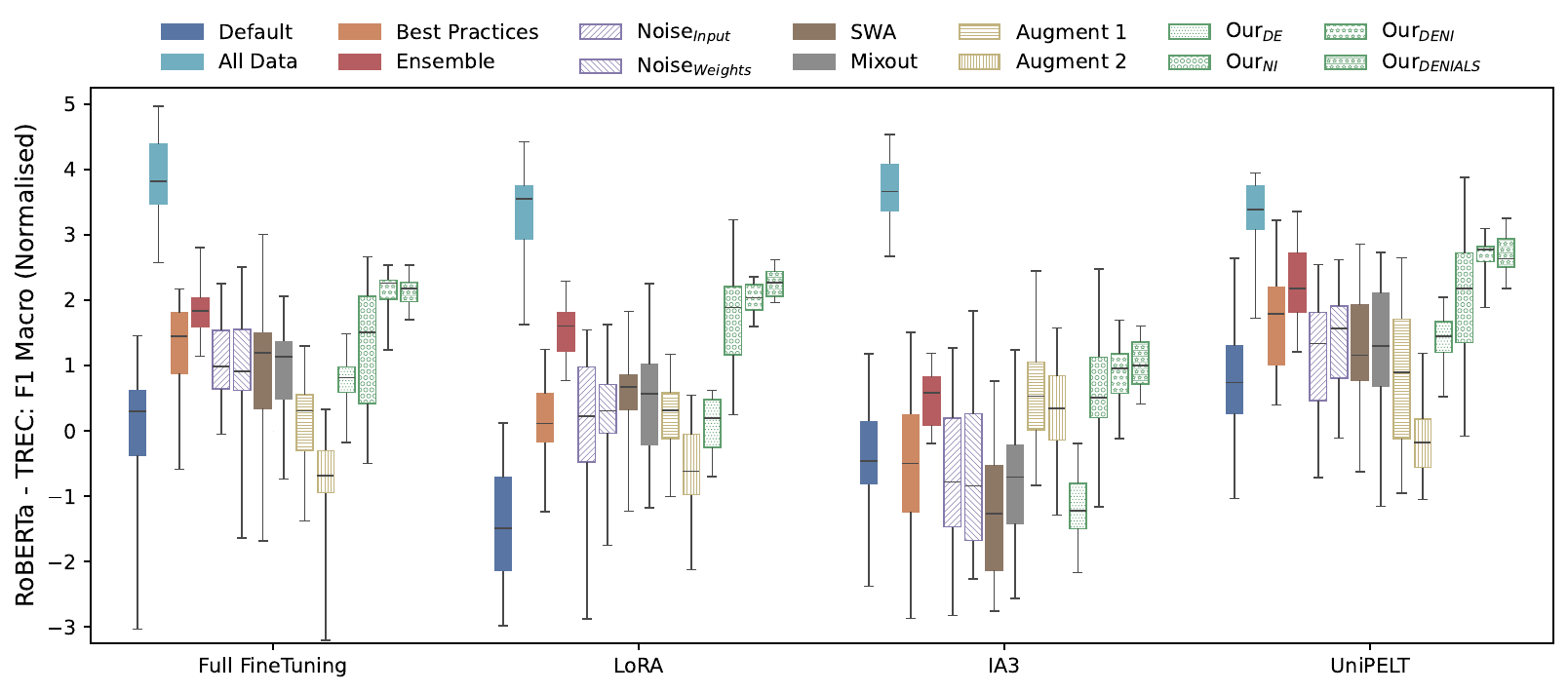}
    \caption{Benefit of mitigation strategies for the different fine-tuning methods using RoBERTa on TREC dataset. The benefit is calculated as difference to the mean performance of the \textit{Default} baseline when using full fine-tuning. The different mitigation strategies are beneficial for all fine-tuning methods, but with different overall benefit (e.g., \textit{Augment} on IA3).}
    \label{fig:roberta-trec-boxplot}
\end{figure*}

\begin{figure*}
    \centering
    \includegraphics[width=1\linewidth]{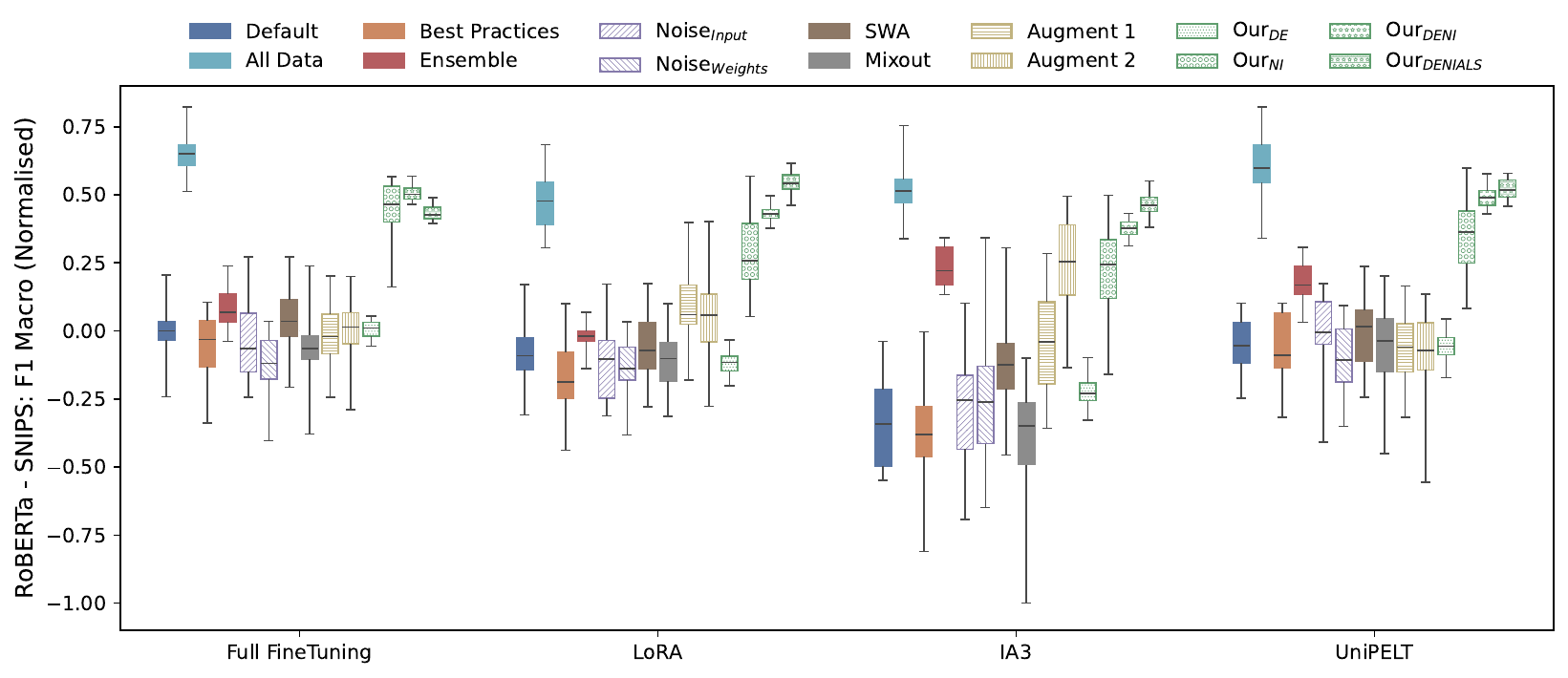}
    \caption{Benefit of mitigation strategies for the different fine-tuning methods using RoBERTa on SNIPS dataset. The benefit is calculated as difference to the mean performance of the \textit{Default} baseline when using full fine-tuning. The different mitigation strategies are beneficial for all fine-tuning methods, but with different overall benefit (e.g., \textit{Augment} on IA3).}
    \label{fig:roberta-snips-boxplot}
\end{figure*}

\begin{figure*}
    \centering
    \includegraphics[width=1\linewidth]{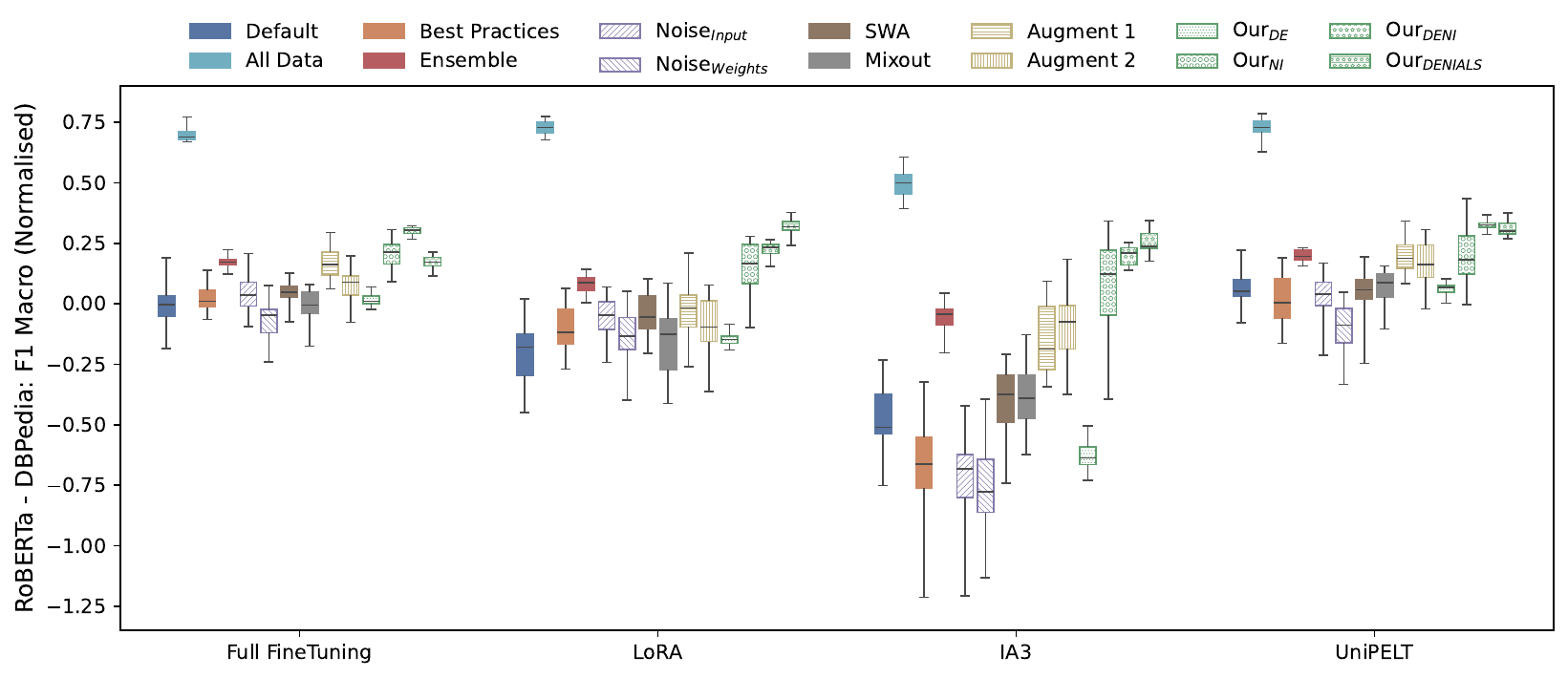}
    \caption{Benefit of mitigation strategies for the different fine-tuning methods using RoBERTa on DBPedia dataset. The benefit is calculated as difference to the mean performance of the \textit{Default} baseline when using full fine-tuning. The different mitigation strategies are beneficial for all fine-tuning methods, but with different overall benefit (e.g., \textit{Augment} on IA3).}
    \label{fig:roberta-dbpedia-boxplot}
\end{figure*}

\begin{figure*}
    \centering
    \includegraphics[width=1\linewidth]{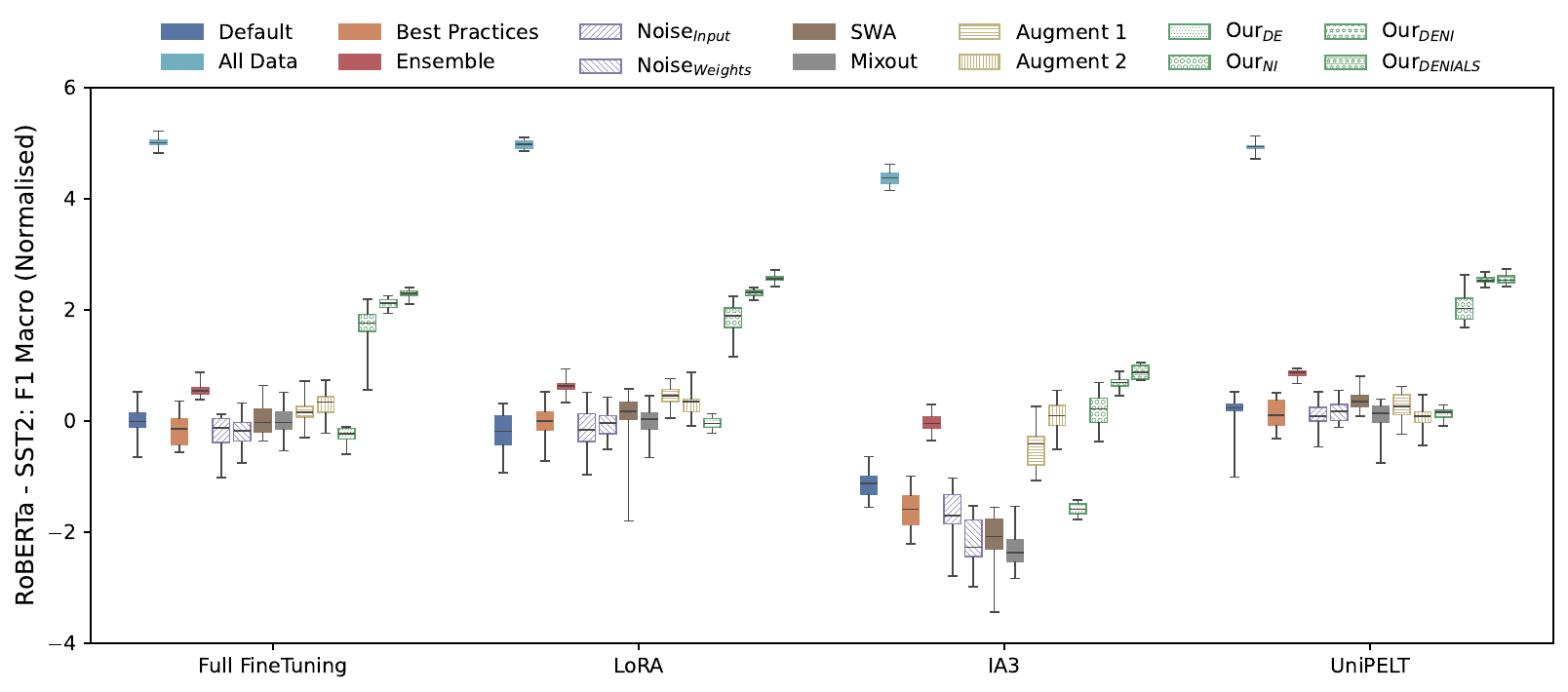}
    \caption{Benefit of mitigation strategies for the different fine-tuning methods using RoBERTa on SST2 dataset. The benefit is calculated as difference to the mean performance of the \textit{Default} baseline when using full fine-tuning. The different mitigation strategies are beneficial for all fine-tuning methods, but with different overall benefit (e.g., \textit{Augment} on IA3).}
    \label{fig:roberta-sst2-boxplot}
\end{figure*}

\begin{figure*}
    \centering
    \includegraphics[width=1\linewidth]{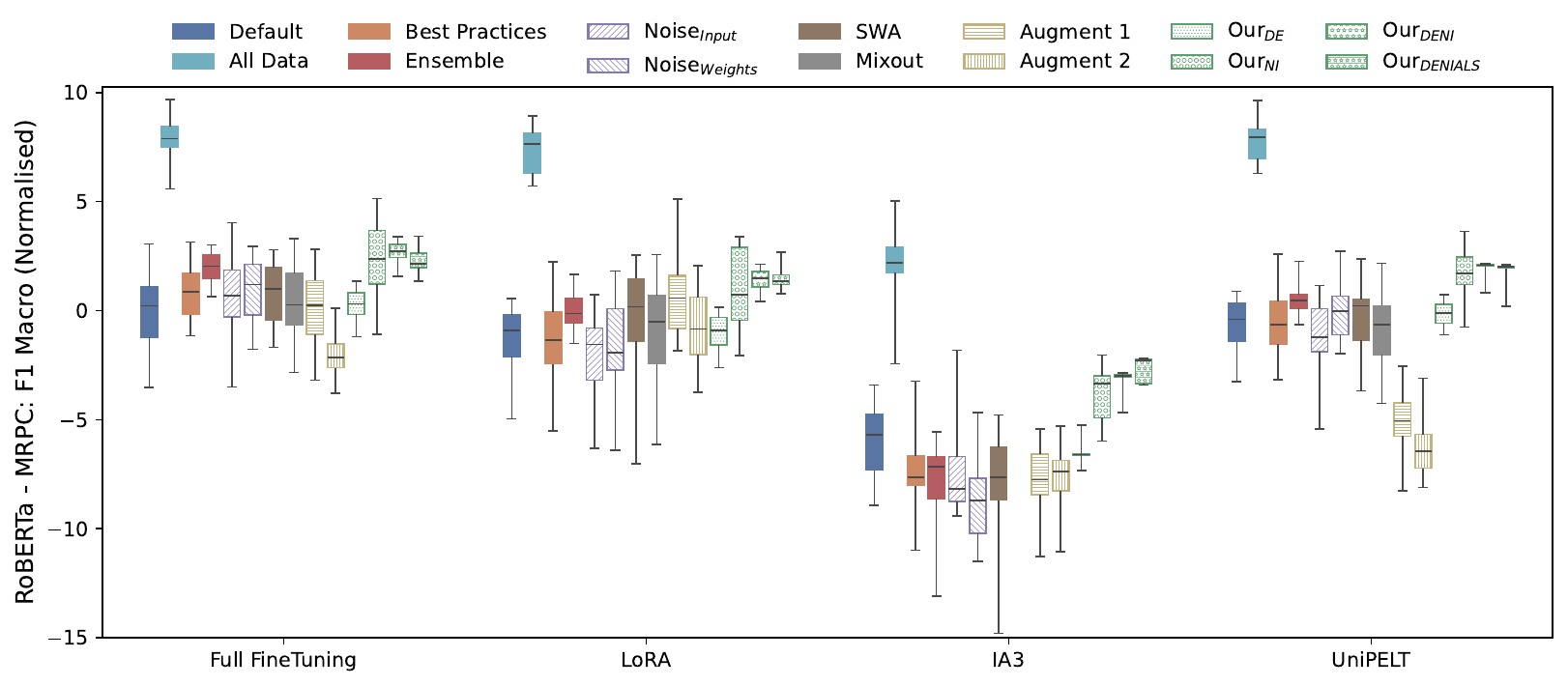}
    \caption{Benefit of mitigation strategies for the different fine-tuning methods using RoBERTa on MRPC dataset. The benefit is calculated as difference to the mean performance of the \textit{Default} baseline when using full fine-tuning. The different mitigation strategies are beneficial for all fine-tuning methods, but with different overall benefit (e.g., \textit{Augment} on IA3).}
    \label{fig:roberta-mrpc-boxplot}
\end{figure*}

\begin{figure*}
    \centering
    \includegraphics[width=1\linewidth]{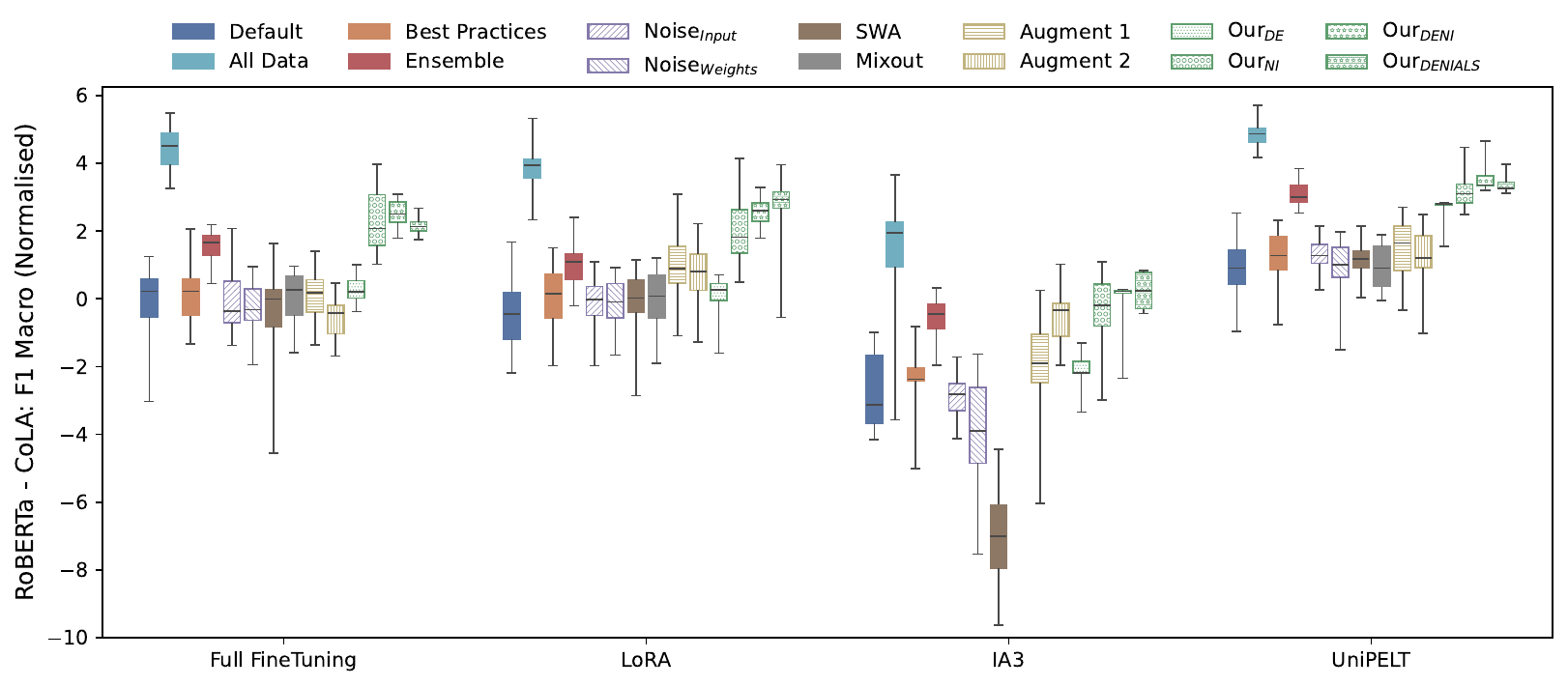}
    \caption{Benefit of mitigation strategies for the different fine-tuning methods using RoBERTa on CoLA dataset. The benefit is calculated as difference to the mean performance of the \textit{Default} baseline when using full fine-tuning. The different mitigation strategies are beneficial for all fine-tuning methods, but with different overall benefit (e.g., \textit{Augment} on IA3).}
    \label{fig:roberta-cola-boxplot}
\end{figure*}

\begin{figure*}
    \centering
    \includegraphics[width=1\linewidth]{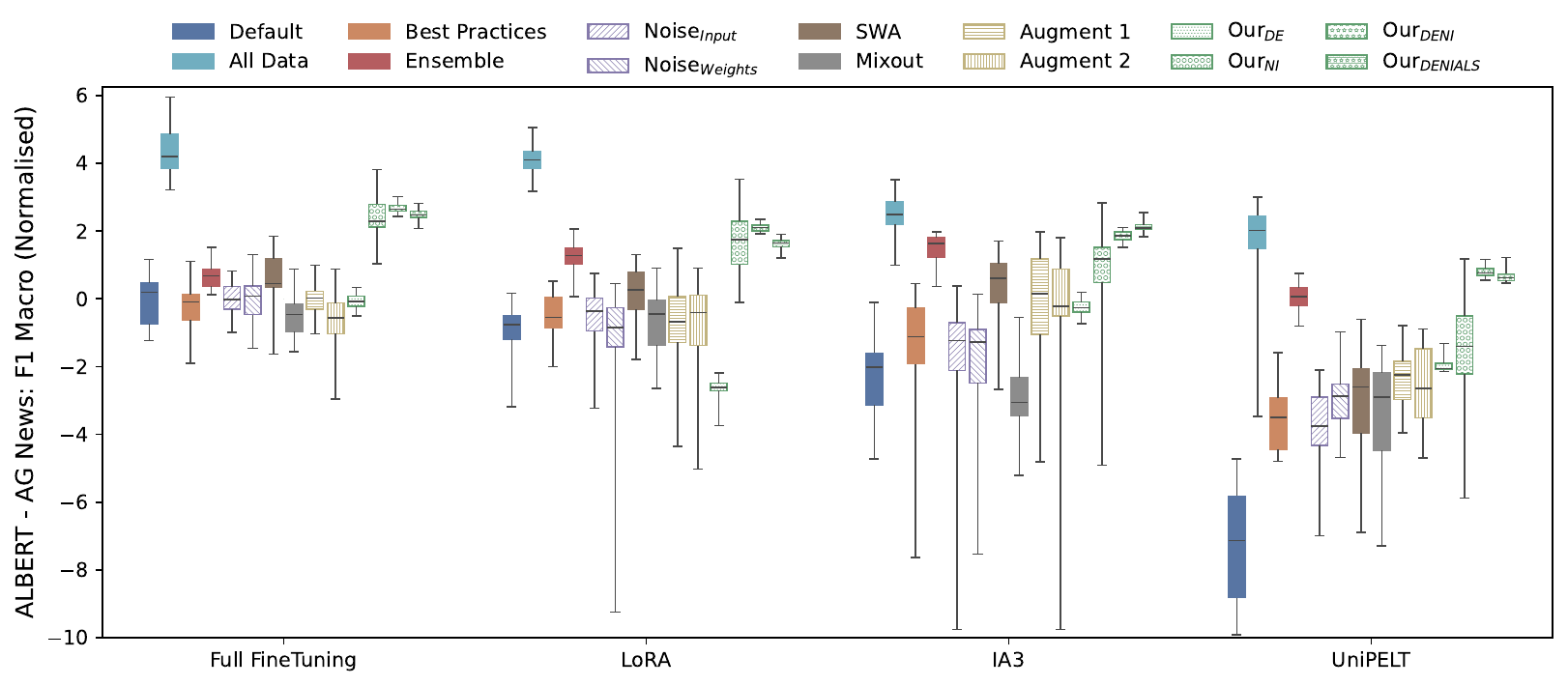}
    \caption{Benefit of mitigation strategies for the different fine-tuning methods using ALBERT on AG News dataset. The benefit is calculated as difference to the mean performance of the \textit{Default} baseline when using full fine-tuning. The different mitigation strategies are beneficial for all fine-tuning methods, but with different overall benefit (e.g., \textit{Augment} on IA3).}
    \label{fig:albert-agnews-boxplot}
\end{figure*}

\begin{figure*}
    \centering
    \includegraphics[width=1\linewidth]{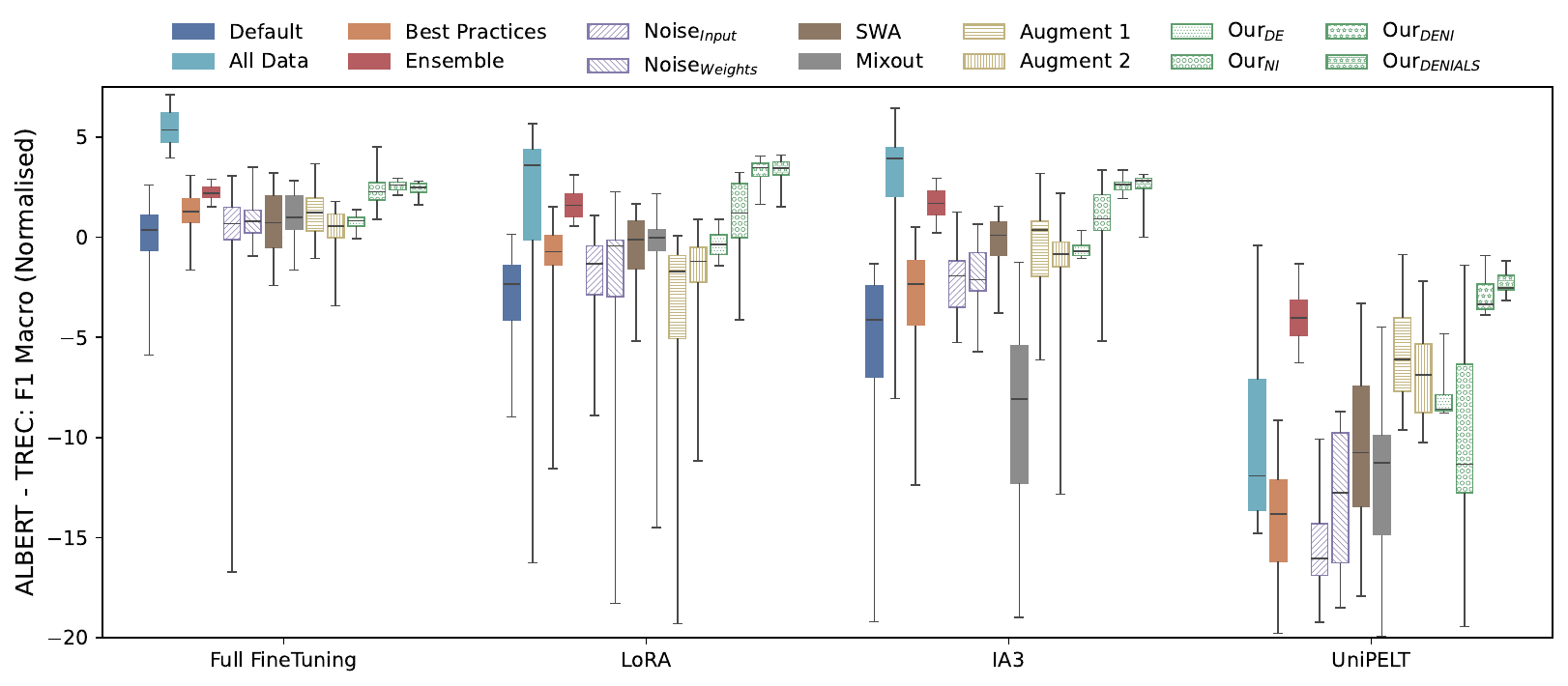}
    \caption{Benefit of mitigation strategies for the different fine-tuning methods using ALBERT on TREC dataset. The benefit is calculated as difference to the mean performance of the \textit{Default} baseline when using full fine-tuning. The different mitigation strategies are beneficial for all fine-tuning methods, but with different overall benefit (e.g., \textit{Augment} on IA3).}
    \label{fig:albert-trec-boxplot}
\end{figure*}

\begin{figure*}
    \centering
    \includegraphics[width=1\linewidth]{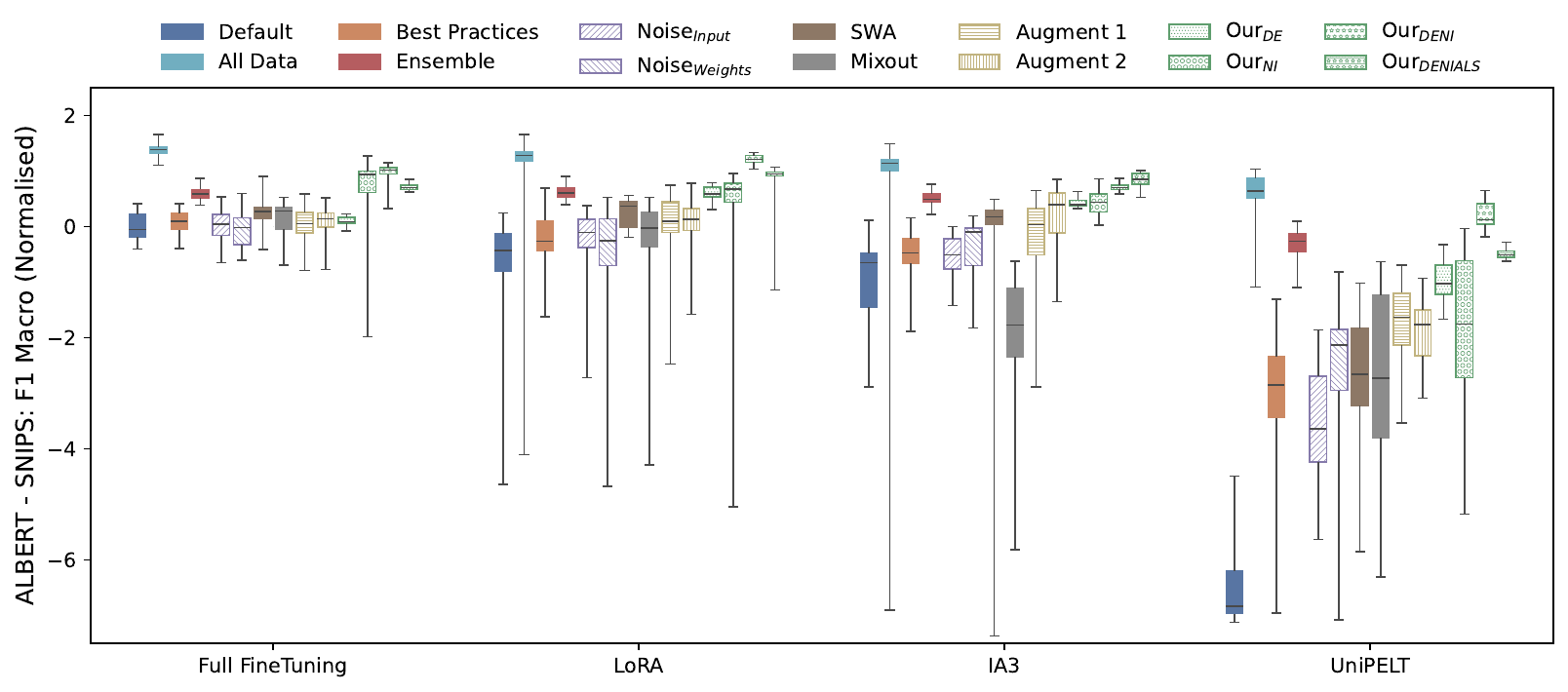}
    \caption{Benefit of mitigation strategies for the different fine-tuning methods using ALBERT on SNIPS dataset. The benefit is calculated as difference to the mean performance of the \textit{Default} baseline when using full fine-tuning. The different mitigation strategies are beneficial for all fine-tuning methods, but with different overall benefit (e.g., \textit{Augment} on IA3).}
    \label{fig:albert-snips-boxplot}
\end{figure*}

\begin{figure*}
    \centering
    \includegraphics[width=1\linewidth]{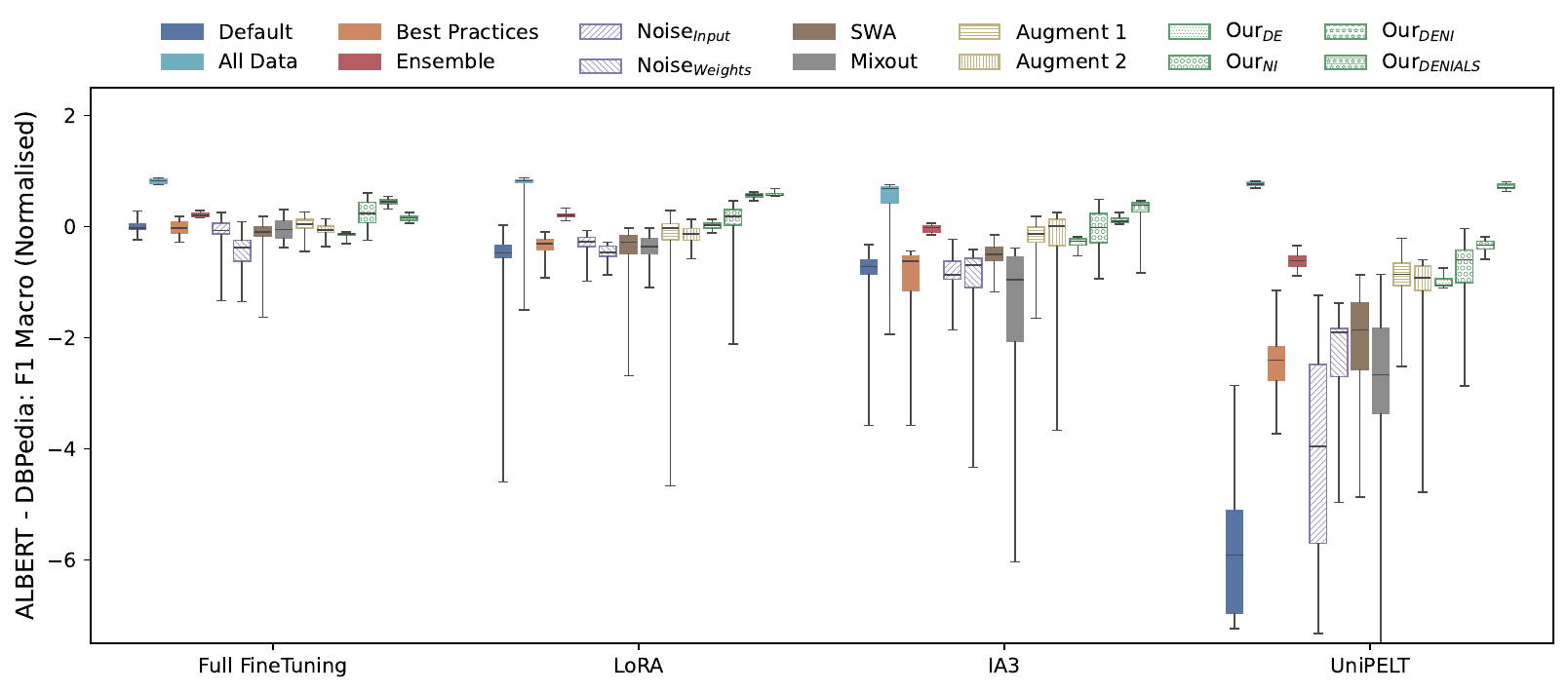}
    \caption{Benefit of mitigation strategies for the different fine-tuning methods using ALBERT on DBPedia dataset. The benefit is calculated as difference to the mean performance of the \textit{Default} baseline when using full fine-tuning. The different mitigation strategies are beneficial for all fine-tuning methods, but with different overall benefit (e.g., \textit{Augment} on IA3).}
    \label{fig:albert-dbpedia-boxplot}
\end{figure*}

\begin{figure*}
    \centering
    \includegraphics[width=1\linewidth]{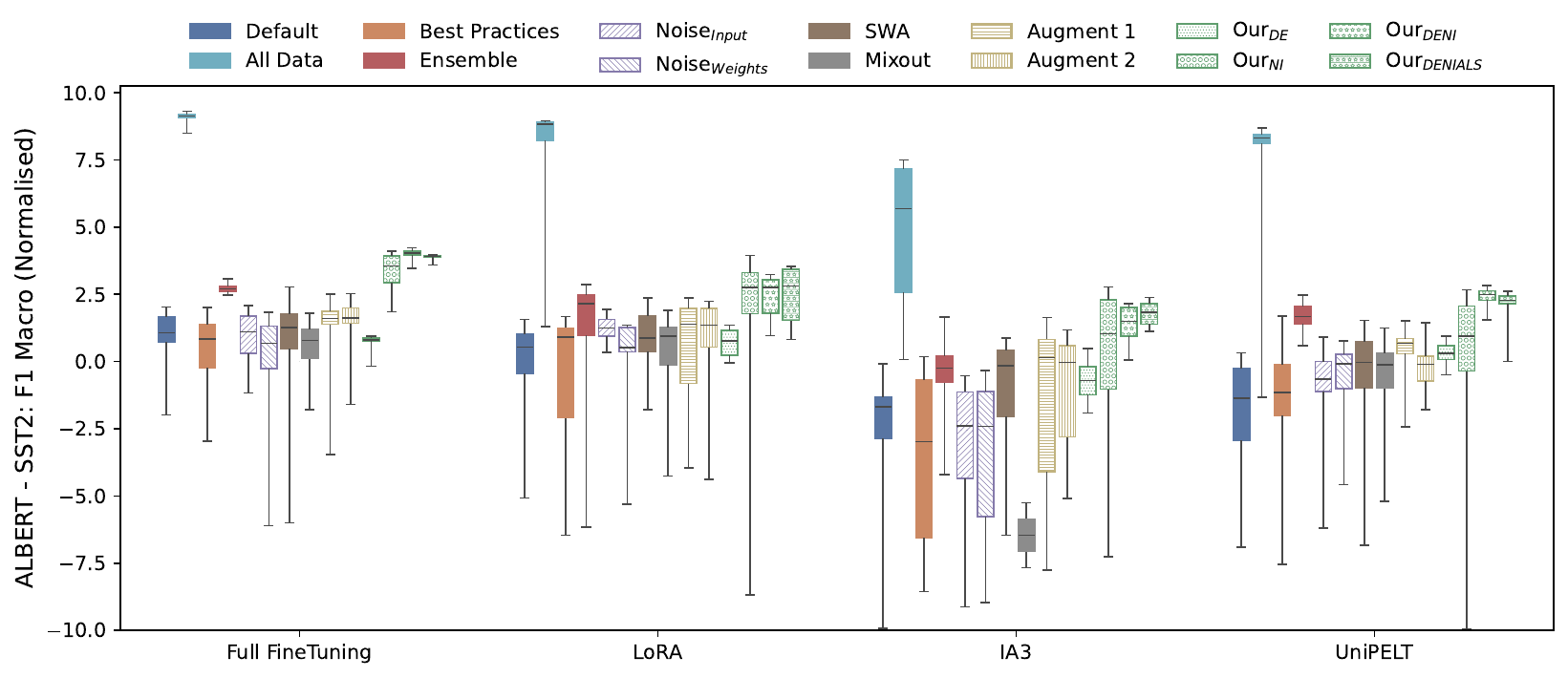}
    \caption{Benefit of mitigation strategies for the different fine-tuning methods using ALBERT on SST2 dataset. The benefit is calculated as difference to the mean performance of the \textit{Default} baseline when using full fine-tuning. The different mitigation strategies are beneficial for all fine-tuning methods, but with different overall benefit (e.g., \textit{Augment} on IA3).}
    \label{fig:albert-sst2-boxplot}
\end{figure*}

\begin{figure*}
    \centering
    \includegraphics[width=1\linewidth]{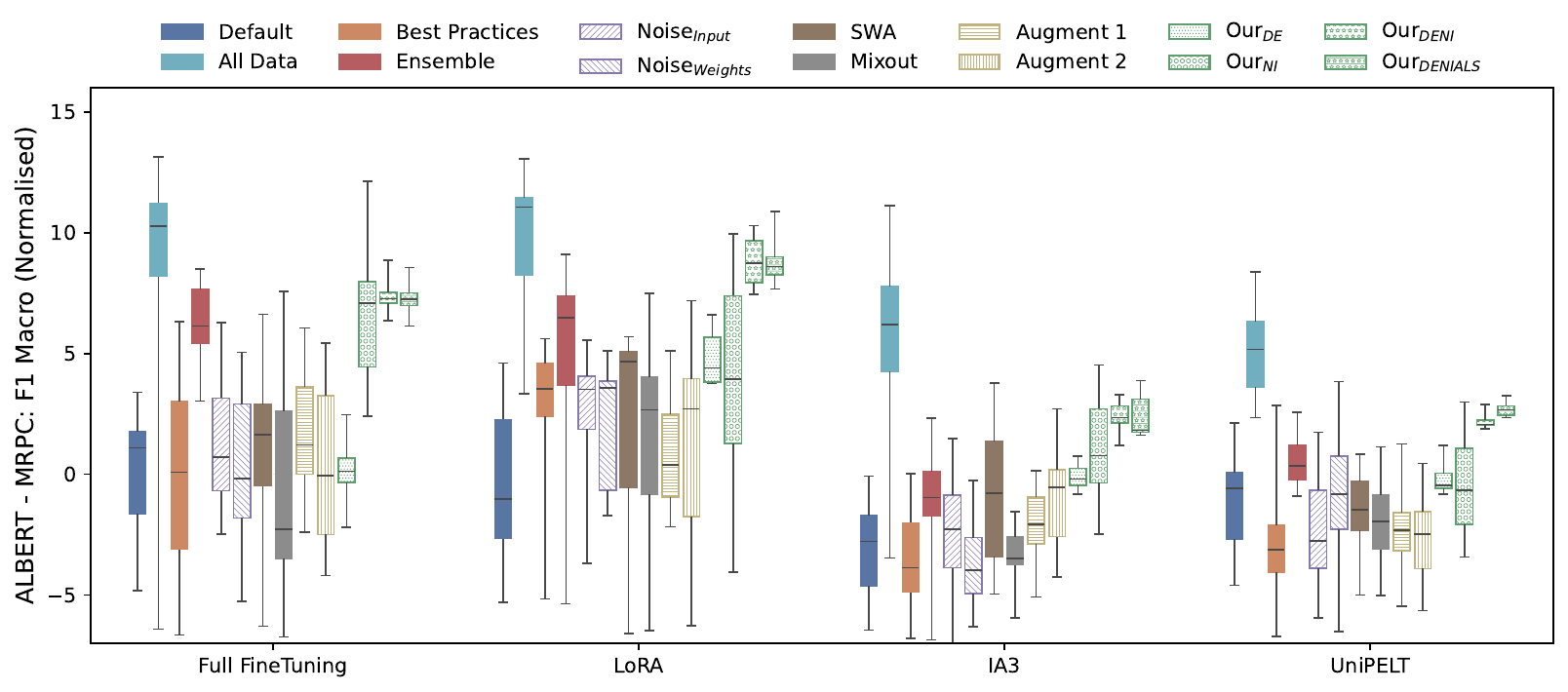}
    \caption{Benefit of mitigation strategies for the different fine-tuning methods using ALBERT on MRPC dataset. The benefit is calculated as difference to the mean performance of the \textit{Default} baseline when using full fine-tuning. The different mitigation strategies are beneficial for all fine-tuning methods, but with different overall benefit (e.g., \textit{Augment} on IA3).}
    \label{fig:albert-mrpc-boxplot}
\end{figure*}

\begin{figure*}
    \centering
    \includegraphics[width=1\linewidth]{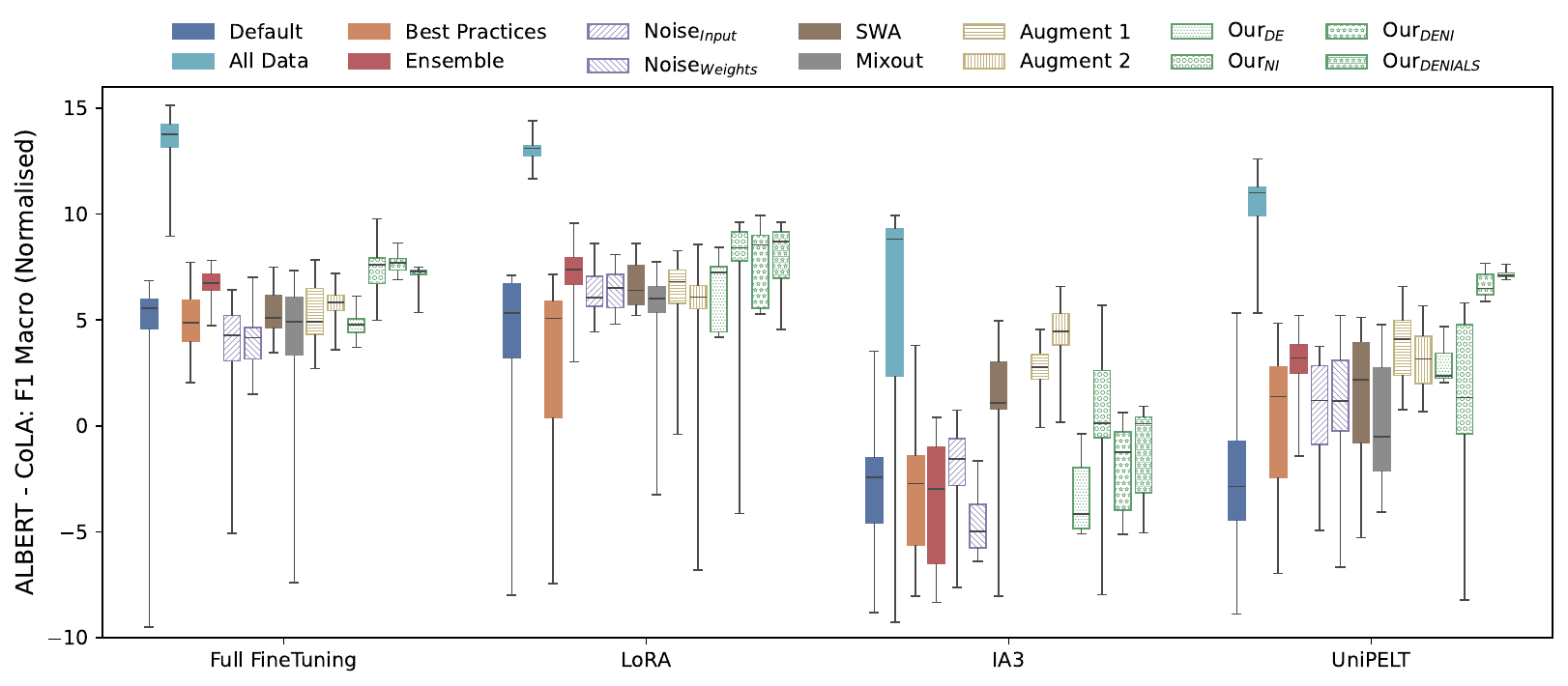}
    \caption{Benefit of mitigation strategies for the different fine-tuning methods using ALBERT on CoLA dataset. The benefit is calculated as difference to the mean performance of the \textit{Default} baseline when using full fine-tuning. The different mitigation strategies are beneficial for all fine-tuning methods, but with different overall benefit (e.g., \textit{Augment} on IA3).}
    \label{fig:albert-cola-boxplot}
\end{figure*}

In this Appendix we provide the results for each combination of dataset, model, mitigation strategy and fine-tuning approach. All the results are provided in Table~\ref{tab:results-bert} for the BERT model, in Table~\ref{tab:results-roberta} for the RoBERTa model and in Table~\ref{tab:results-albert} for the ALBERT model. In addition, we provide results in a form of figures (boxplots) separately for each dataset and model in a following way:
\begin{itemize}
    \item For BERT model in Figure~\ref{fig:agnews-boxplot} for AG News dataset, Figure~\ref{fig:trec-boxplot} for TREC dataset, Figure~\ref{fig:snips-boxplot} for SNIPS dataset, Figure~\ref{fig:dbpedia-boxplot} for DBPedia dataset, Figure~\ref{fig:sst2-boxplot} for SST2 dataset, Figure~\ref{fig:mrpc-boxplot} for MRPC dataset and Figure~\ref{fig:cola-boxplot} for CoLA dataset.
    \item For RoBERTa model in Figure~\ref{fig:roberta-agnews-boxplot} for AG News dataset, Figure~\ref{fig:roberta-trec-boxplot} for TREC dataset, Figure~\ref{fig:roberta-snips-boxplot} for SNIPS dataset, Figure~\ref{fig:roberta-dbpedia-boxplot} for DBPedia dataset, Figure~\ref{fig:roberta-sst2-boxplot} for SST2 dataset, Figure~\ref{fig:roberta-mrpc-boxplot} for MRPC dataset and Figure~\ref{fig:roberta-cola-boxplot} for CoLA dataset.
    \item For ALBERT model in Figure~\ref{fig:albert-agnews-boxplot} for AG News dataset, Figure~\ref{fig:albert-trec-boxplot} for TREC dataset, Figure~\ref{fig:albert-snips-boxplot} for SNIPS dataset, Figure~\ref{fig:albert-dbpedia-boxplot} for DBPedia dataset, Figure~\ref{fig:albert-sst2-boxplot} for SST2 dataset, Figure~\ref{fig:albert-mrpc-boxplot} for MRPC dataset and Figure~\ref{fig:albert-cola-boxplot} for CoLA dataset.
\end{itemize}

For the figures, we omit the failed runs in the visualisations (i.e., runs that show significantly lower performance for the given strategy or than the average of all the other strategies, for example achieving F1 score of 1\% instead of 80\% on TREC dataset). As such, specific mitigation strategies may be missing in the figures in some cases -- in such case, all the runs using the mitigation strategies resulted in failed runs, such as Mixout on MRPC or CoLA dataset with RoBERTa model trained using IA3, or the \textit{Default} baseline on TREC dataset with ALBERT model trained using UniPELT.

When it comes to the further models and dataset, we observe only minor differences in the results compared to the ones reported in the main body of the paper. We summarise the main findings and potential differences below.

On the different datasets, we observe different benefit for the individual mitigation strategies. For example, for the BERT model trained using UniPELT, the \textit{SWA} mitigation strategy achieves similar performance to the \textit{Ensemble} mitigation strategies on the AG News, MRPC or CoLA datasets, while on SNIPS, DBPedia or TREC it leads to lower benefit. Similar behaviour can be observed for other methods as well. The data augmentation provides significantly larger benefit when the number of trainable parameters is small, such as when using IA3 or the ALBERT model. For example, the \textit{Augment N} strategy outperforms the \textit{Ensemble} strategy in these cases. In addition, this has also the effect on our proposed method. The \textit{DENI} method (or \textit{DENIALS} method) achieves the highest performance increase and standard deviation reduction across all datasets, models and fine-tuning approaches, with the exception of RoBERTa and ALBERT trained using IA3 on CoLA dataset, where the \textit{Augment 2} strategy provides the most benefit. This can be explained by: 1) the benefit of augmented samples when the number of trainable parameters is small; and 2) we observe that the augmented samples for CoLA dataset appear to provide larger benefit across all models and fine-tuning approaches. As such, in these cases the benefit of adding further samples through augmentation outweighs benefit of regular mitigation strategies.

In addition, we observe different level of standard deviation for the different models, especially for the ALBERT model. As such, we also observe a different benefit of the individual mitigation strategies on these models as compared to BERT. For example, we observe higher benefit of the \textit{Stochastic Weight Averaging}, even outperforming the \textit{Ensemble} mitigation strategy and all of our proposed methods when using ALBERT model trained using IA3 on the CoLA dataset. At the same time, the \textit{Ensemble} strategy and the \textit{DENI} method provide more consistent and more significant reduction in the standard deviation on the ALBERT and RoBERTa models across the different datasets and fine-tuning approaches, while the majority of the remaining mitigation strategies often show significantly lower consistency, often not leading to any mitigation benefit over the \textit{Default} baseline. As such, we can better observe the efficiency and effectiveness of the different mitigation strategies on these cases -- with \textit{Ensemble} and our \textit{DENI} method being significantly more effective and efficient.

The different components of the \textit{DENI} method provide the same benefit across all the models, datasets and fine-tuning methods. The \textit{Delayed Ensemble (DE)} reduces the standard deviation, achieving deviation on par or often even lower than the one from \textit{Ensemble}. However, it also leads to lower performance than the \textit{Ensemble} method. Only in specific cases, such as BERT on the MRPC dataset trained using UniPELT or ALBERT on the CoLA dataset trained using IA3, we observe similar performance of the \textit{Delayed Ensemble} to the \textit{Ensemble} or other components of the \textit{DENI} method. The \textit{Noisy Interpolation (NI)} leads to an increase in the performance, but often at the cost of higher standard deviation. When combined in the \textit{DENI} method, the complementarity of these two components leads to a significant benefit, increasing the overall performance while reducing the standard deviation. Finally, combining the \textit{DENI} method with augmented labelled samples (to obtain the \textit{DENIALS} method) provides different benefit based on the dataset and the fine-tuning method used. In about half the cases (mainly when using full fine-tuning or UniPELT) the benefit of the \textit{DENI} method is higher, while in other (mainly when using LoRA or IA3) the \textit{DENIALS} method leads to higher performance and lower standard deviation. However, taking the cost of the method into consideration, the \textit{DENI} method can be viewed as more efficient, as it requires half of the computation cost, while producing similar increase in performance and reduction in standard deviation -- the difference in the performance and deviation in results between the \textit{DENI} and \textit{DENIALS} methods is not statistically significant (p-value of 0.14 using the Mann-Whitney U test; p-value of 0.82 using the Levene test).

To summarise, the \textit{DENI} method provides consistent benefit across different models, datasets and fine-tuning strategies, even in cases when the deviation in results is significantly higher or even negligible. At the same time, the other mitigation strategies we compare against (with the exception of \textit{Ensemble}) show lower consistency in their benefit. In some cases, using the mitigation strategies, especially \textit{DENIALS} or even \textit{DENI} leads to a performance that is approaching the one when using all data in the dataset, while also reducing the standard deviation. Finally, increasing the number of available labelled samples through augmentation has significantly higher benefit when the number of trainable parameters is low (using IA3 fine-tuning or ALBERT model), outperforming all the regular mitigation strategies that modify the optimisation process in specific cases.

\begin{table*}
\begin{center}
\footnotesize
\begin{sc}
\tabcolsep=0.15cm
\begin{tabular}{lccccccc}
\toprule
RoBERTa & AG News       & TREC  & SNIPS & DBPedia       & SST2  & MRPC  & CoLA \\ \midrule
\multicolumn{8}{c}{Full FineTuning} \\
Default & 85.77$_{0.433}$       & 90.98$_{0.979}$       & 98.42$_{0.092}$       & 98.36$_{0.082}$       & 90.14$_{0.270}$       & 65.55$_{1.654}$       & 77.16$_{0.934}$ \\
All Data        & 89.27$_{0.391}$       & 94.81$_{0.691}$       & 99.07$_{0.086}$       & 99.06$_{0.028}$       & 95.15$_{0.084}$       & 73.45$_{1.031}$       & 81.56$_{0.632}$ \\
Best Practices  & 86.12$_{0.418}$       & 92.25$_{0.733}$       & 98.36$_{0.124}$       & 98.38$_{0.059}$       & 89.99$_{0.288}$       & 66.43$_{1.276}$       & 77.27$_{0.870}$ \\
Ensemble        & 86.56$_{0.236}$       & 92.80$_{0.378}$       & 98.49$_{0.073}$       & 98.53$_{0.023}$       & 90.69$_{0.119}$       & 67.51$_{0.703}$       & 78.73$_{0.431}$ \\
Noise$_{Input}$ & 86.04$_{0.485}$       & 92.12$_{0.645}$       & 98.37$_{0.144}$       & 98.40$_{0.073}$       & 89.93$_{0.297}$       & 66.13$_{1.933}$       & 77.04$_{0.853}$ \\
Noise$_{Weights}$       & 86.23$_{0.547}$       & 91.90$_{0.954}$       & 98.29$_{0.111}$       & 98.29$_{0.083}$       & 89.94$_{0.280}$       & 66.56$_{1.326}$       & 76.92$_{0.678}$ \\
SWA     & 86.44$_{0.450}$       & 91.81$_{1.037}$       & 98.46$_{0.108}$       & 98.40$_{0.044}$       & 90.15$_{0.257}$       & 66.35$_{1.373}$       & 76.70$_{1.357}$ \\
Mixout  & 86.02$_{0.471}$       & 91.87$_{0.724}$       & 98.37$_{0.126}$       & 98.35$_{0.059}$       & 90.13$_{0.251}$       & 65.94$_{1.641}$       & 77.20$_{0.760}$ \\
Augment 1       & 85.91$_{0.524}$       & 91.09$_{0.708}$       & 98.40$_{0.104}$       & 98.52$_{0.067}$       & 90.30$_{0.226}$       & 65.54$_{1.723}$       & 77.25$_{0.656}$ \\
Augment 2       & 85.96$_{0.413}$       & 90.17$_{0.770}$       & 98.42$_{0.103}$       & 98.43$_{0.068}$       & 90.43$_{0.246}$       & 63.49$_{1.031}$       & 76.61$_{0.631}$ \\
Our$_{DE}$      & 86.24$_{0.192}$       & 91.76$_{0.335}$       & 98.42$_{0.031}$       & 98.37$_{0.024}$       & 89.89$_{0.138}$       & 65.82$_{0.693}$       & 77.40$_{0.350}$ \\
Our$_{NI}$      & 88.10$_{0.530}$       & 92.27$_{0.972}$       & 98.86$_{0.105}$       & 98.56$_{0.061}$       & 91.85$_{0.375}$       & 67.88$_{1.692}$       & 79.43$_{0.882}$ \\
Our$_{DENI}$    & \underline{\textbf{88.67$_{0.133}$}}       & \textbf{93.15$_{0.287}$}       & \underline{\textbf{98.92$_{0.027}$}}       & \underline{\textbf{98.66$_{0.018}$}}       & 92.25$_{0.085}$       & \underline{\textbf{68.23$_{0.483}$}}       & \textbf{79.69$_{0.365}$} \\
Our$_{DENIALS}$ & 88.26$_{0.150}$       & \underline{93.11$_{0.217}$}       & \underline{98.85$_{0.027}$}       & 98.53$_{0.025}$       & \underline{\textbf{92.43$_{0.067}$}}       & 67.83$_{0.530}$       & \underline{79.31$_{0.235}$} \\ \midrule
\multicolumn{8}{c}{LoRA} \\
Default & 85.73$_{0.448}$       & 89.55$_{0.907}$       & 98.34$_{0.112}$       & 98.16$_{0.133}$       & 89.95$_{0.359}$       & 64.19$_{1.592}$       & 76.71$_{1.040}$ \\
All Data        & 89.45$_{0.374}$       & 94.32$_{0.686}$       & 98.88$_{0.102}$       & 99.08$_{0.027}$       & 95.12$_{0.075}$       & 72.95$_{0.984}$       & 81.07$_{0.632}$ \\
Best Practices  & 85.87$_{0.477}$       & 91.12$_{0.648}$       & 98.25$_{0.146}$       & 98.25$_{0.087}$       & 90.13$_{0.285}$       & 64.21$_{1.711}$       & 77.25$_{0.912}$ \\
Ensemble        & 86.56$_{0.273}$       & 92.50$_{0.422}$       & 98.39$_{0.054}$       & 98.44$_{0.037}$       & 90.77$_{0.122}$       & 65.49$_{0.807}$       & 78.11$_{0.636}$ \\
Noise$_{Input}$ & 86.04$_{0.515}$       & 91.06$_{1.140}$       & 98.31$_{0.151}$       & 98.30$_{0.086}$       & 90.02$_{0.340}$       & 63.38$_{2.061}$       & 77.00$_{0.762}$ \\
Noise$_{Weights}$       & 85.82$_{0.471}$       & 91.20$_{0.885}$       & 98.28$_{0.103}$       & 98.23$_{0.107}$       & 90.09$_{0.250}$       & 63.65$_{2.281}$       & 77.01$_{0.715}$ \\
SWA     & 86.28$_{0.436}$       & 91.50$_{0.624}$       & 98.37$_{0.125}$       & 98.31$_{0.087}$       & 90.21$_{0.498}$       & 65.24$_{2.331}$       & 77.11$_{0.871}$ \\
Mixout  & 85.83$_{0.551}$       & 91.49$_{0.865}$       & 98.31$_{0.118}$       & 98.20$_{0.131}$       & 90.11$_{0.285}$       & 64.60$_{2.130}$       & 77.20$_{0.855}$ \\
Augment 1       & 86.36$_{0.506}$       & 91.19$_{0.625}$       & 98.49$_{0.143}$       & 98.32$_{0.115}$       & 90.60$_{0.189}$       & 64.38$_{8.130}$       & 78.11$_{0.921}$ \\
Augment 2       & 86.13$_{0.423}$       & 90.45$_{0.641}$       & 98.47$_{0.157}$       & 98.26$_{0.123}$       & 90.46$_{0.224}$       & 64.88$_{1.702}$       & 77.90$_{0.886}$ \\
Our$_{DE}$      & 85.61$_{0.243}$       & 91.10$_{0.402}$       & 98.29$_{0.045}$       & 98.21$_{0.029}$       & 90.10$_{0.105}$       & 64.52$_{0.804}$       & 77.26$_{0.586}$ \\
Our$_{NI}$      & 87.55$_{0.575}$       & 92.73$_{0.759}$       & 98.70$_{0.139}$       & 98.51$_{0.110}$       & 91.98$_{0.279}$       & 66.51$_{1.856}$       & 79.17$_{0.851}$ \\
Our$_{DENI}$    & 87.98$_{0.219}$       & 93.01$_{0.226}$       & \underline{98.84$_{0.029}$}       & \underline{98.58$_{0.027}$}       & 92.44$_{0.072}$       & \underline{66.95$_{0.469}$}       & \underline{79.74$_{0.397}$} \\
Our$_{DENIALS}$ & \underline{\textbf{88.14$_{0.175}$}}       & \underline{\textbf{93.25$_{0.205}$}}       & \textbf{98.96$_{0.038}$}       & \textbf{98.67$_{0.034}$}       & \underline{\textbf{92.70$_{0.070}$}}       & \textbf{67.06$_{0.490}$}       & \textbf{79.95$_{0.882}$} \\ \midrule
\multicolumn{8}{c}{IA3} \\
Default & 84.98$_{0.557}$       & 90.59$_{0.887}$       & 98.07$_{0.156}$       & 97.89$_{0.128}$       & 89.00$_{0.230}$       & 58.58$_{4.528}$       & 73.88$_{2.586}$ \\
All Data        & 88.83$_{0.562}$       & 94.66$_{0.560}$       & 98.93$_{0.094}$       & 98.85$_{0.059}$       & 94.51$_{0.122}$       & 64.83$_{8.397}$       & 73.03$_{13.417}$ \\
Best Practices  & 85.03$_{0.503}$       & 86.20$_{18.439}$      & 93.32$_{20.601}$      & 97.69$_{0.192}$       & 88.51$_{0.374}$       & 55.34$_{8.258}$       & 73.04$_{7.342}$ \\
Ensemble        & 86.39$_{0.296}$       & 91.49$_{0.441}$       & 98.65$_{0.072}$       & 98.29$_{0.063}$       & 90.11$_{0.158}$       & 57.73$_{1.880}$       & 76.63$_{0.542}$ \\
Noise$_{Input}$ & 84.81$_{0.557}$       & 90.35$_{1.104}$       & 98.13$_{0.199}$       & 97.63$_{0.192}$       & 88.46$_{0.395}$       & 56.39$_{5.678}$       & 71.41$_{10.636}$ \\
Noise$_{Weights}$       & 80.88$_{15.803}$      & 90.29$_{1.211}$       & 98.16$_{0.239}$       & 97.61$_{0.179}$       & 87.97$_{0.409}$       & 49.52$_{13.034}$      & 71.59$_{7.086}$ \\
SWA     & 86.13$_{0.474}$       & 89.66$_{0.951}$       & 98.29$_{0.175}$       & 95.47$_{10.860}$      & 88.02$_{0.426}$       & 52.74$_{12.001}$      & 68.29$_{6.409}$ \\
Mixout  & 84.81$_{0.507}$       & 90.26$_{1.052}$       & 98.02$_{0.215}$       & 97.97$_{0.119}$       & 85.22$_{11.347}$      & 33.22$_{7.739}$       & 33.01$_{9.143}$ \\
Augment 1       & 86.30$_{0.660}$       & 91.48$_{0.846}$       & 98.39$_{0.192}$       & 98.21$_{0.146}$       & 86.81$_{12.490}$      & 56.49$_{4.422}$       & 75.56$_{6.946}$ \\
Augment 2       & 86.41$_{0.543}$       & 91.19$_{0.796}$       & 98.64$_{0.176}$       & 98.26$_{0.129}$       & 90.21$_{0.261}$       & 57.95$_{1.282}$       & \textbf{78.70$_{0.790}$} \\
Our$_{DE}$      & 84.93$_{0.275}$       & 89.83$_{0.495}$       & 98.19$_{0.053}$       & 97.73$_{0.058}$       & \underline{88.55$_{0.107}$}       & \underline{58.96$_{0.407}$}       & \underline{75.09$_{0.396}$} \\
Our$_{NI}$      & 87.49$_{0.673}$       & 91.65$_{0.952}$       & 98.64$_{0.163}$       & 98.43$_{0.196}$       & 90.33$_{0.313}$       & 58.43$_{10.157}$      & 74.31$_{11.236}$ \\
Our$_{DENI}$    & 87.70$_{0.225}$       & 91.85$_{0.457}$       & \underline{98.79$_{0.035}$}       & \underline{98.56$_{0.037}$}       & 90.82$_{0.108}$       & 62.40$_{0.517}$       & 77.25$_{0.559}$ \\
Our$_{DENIALS}$ & \underline{\textbf{87.75$_{0.207}$}}       & \underline{\textbf{92.00$_{0.337}$}}       & \textbf{98.88$_{0.042}$}       & \textbf{98.61$_{0.042}$}       & \textbf{91.02$_{0.116}$}       & \textbf{62.81$_{0.554}$}       & 77.41$_{0.546}$ \\ \midrule
\multicolumn{8}{c}{UniPELT} \\
Default & 85.98$_{0.359}$       & 91.84$_{0.870}$       & 98.36$_{0.108}$       & 98.42$_{0.066}$       & 90.34$_{0.304}$       & 62.43$_{7.489}$       & 76.30$_{8.074}$ \\
All Data        & 89.73$_{0.353}$       & 94.31$_{0.526}$       & 99.02$_{0.114}$       & 99.08$_{0.039}$       & 95.07$_{0.093}$       & 73.33$_{0.873}$       & 82.03$_{0.407}$ \\
Best Practices  & 86.07$_{0.446}$       & 92.62$_{0.777}$       & 98.35$_{0.121}$       & 98.38$_{0.102}$       & 90.26$_{0.252}$       & 64.92$_{1.433}$       & 78.45$_{0.754}$ \\
Ensemble        & 87.18$_{0.240}$       & 93.25$_{0.575}$       & 98.59$_{0.076}$       & 98.55$_{0.024}$       & \underline{91.00$_{0.067}$}       & 66.04$_{0.656}$       & 80.26$_{0.372}$ \\
Noise$_{Input}$ & 86.16$_{0.452}$       & 92.18$_{0.809}$       & 98.42$_{0.137}$       & 98.38$_{0.098}$       & 90.25$_{0.206}$       & 59.90$_{10.928}$      & 76.04$_{10.489}$ \\
Noise$_{Weights}$       & 85.96$_{0.623}$       & 92.41$_{0.745}$       & 98.32$_{0.123}$       & 98.25$_{0.114}$       & 90.31$_{0.198}$       & 65.50$_{1.254}$       & 78.07$_{0.758}$ \\
SWA     & 86.64$_{0.402}$       & 92.28$_{0.822}$       & 98.40$_{0.133}$       & 98.42$_{0.088}$       & 88.62$_{8.224}$       & 62.66$_{7.641}$       & 70.06$_{19.817}$ \\
Mixout  & 86.27$_{0.550}$       & 92.23$_{1.027}$       & 98.36$_{0.163}$       & 98.42$_{0.079}$       & 90.22$_{0.257}$       & 64.66$_{1.695}$       & 66.15$_{21.219}$ \\
Augment 1       & 86.02$_{0.432}$       & 91.88$_{1.072}$       & 98.35$_{0.136}$       & 98.55$_{0.073}$       & 90.42$_{0.252}$       & 60.40$_{1.399}$       & 75.97$_{11.790}$ \\
Augment 2       & 85.89$_{0.459}$       & 90.88$_{0.564}$       & 98.33$_{0.153}$       & 98.53$_{0.089}$       & 90.18$_{0.230}$       & 59.23$_{1.251}$       & 78.48$_{0.800}$ \\
Our$_{DE}$      & 86.09$_{0.235}$       & 92.39$_{0.417}$       & 98.36$_{0.056}$       & 98.42$_{0.026}$       & 90.27$_{0.106}$       & 65.42$_{0.512}$       & 79.77$_{0.445}$ \\
Our$_{NI}$      & 87.95$_{0.493}$       & 92.99$_{1.076}$       & 98.76$_{0.144}$       & 98.56$_{0.110}$       & 92.17$_{0.262}$       & 65.97$_{5.566}$       & 72.04$_{19.807}$ \\
Our$_{DENI}$    & \underline{\textbf{88.56$_{0.189}$}}       & \underline{\textbf{93.68$_{0.284}$}}       & 98.91$_{0.043}$       & \underline{\textbf{98.68$_{0.018}$}}       & 92.67$_{0.072}$       & \underline{\textbf{67.56$_{0.275}$}}       & \textbf{80.68$_{0.343}$} \\
Our$_{DENIALS}$ & \underline{\textbf{88.56$_{0.189}$}}       & \textbf{93.68$_{0.298}$}       & \underline{\textbf{98.94$_{0.037}$}}       & 98.67$_{0.030}$       & \textbf{92.70$_{0.084}$}      & 67.46$_{0.396}$       & \underline{80.55$_{0.241}$} \\
\bottomrule
\end{tabular}
\end{sc}
\end{center}
\caption{Comparison of the DENI method with existing mitigation strategies and baselines on full fine-tuning, LoRA, IA3 and UniPELT using RoBERTa model. The comparison is done in terms of overall performance and deviation. The highest performance for each fine-tuning method is in \textbf{bold} and lowest deviation is \underline{underlined} (not considering \textit{All Data} baseline).}
\label{tab:results-roberta}
\end{table*}

\begin{table*}
\begin{center}
\footnotesize
\begin{sc}
\tabcolsep=0.15cm
\begin{tabular}{lccccccc}
\toprule
ALBERT  & AG News       & TREC  & SNIPS & DBPedia       & SST2  & MRPC  & CoLA \\ \midrule
\multicolumn{8}{c}{Full FineTuning} \\
Default & 83.11$_{0.716}$       & 88.45$_{1.794}$       & 97.23$_{0.245}$       & 98.23$_{0.127}$       & 85.49$_{3.983}$       & 60.12$_{2.438}$       & 65.41$_{9.853}$ \\
All Data        & 87.46$_{0.680}$       & 93.98$_{0.912}$       & 98.62$_{0.126}$       & 99.05$_{0.039}$       & 94.58$_{0.189}$       & 67.07$_{7.899}$       & 78.94$_{1.257}$ \\
Best Practices  & 82.95$_{0.679}$       & 89.66$_{1.033}$       & 97.30$_{0.222}$       & 98.21$_{0.143}$       & 86.03$_{1.169}$       & 60.20$_{3.691}$       & 70.22$_{1.473}$ \\
Ensemble        & 83.76$_{0.347}$       & 90.67$_{0.376}$       & 97.83$_{0.122}$       & \underline{98.44$_{0.036}$}       & 88.20$_{0.151}$       & 66.41$_{1.571}$       & 72.13$_{0.664}$ \\
Noise$_{Input}$ & 83.07$_{0.489}$       & 88.33$_{3.974}$       & 97.27$_{0.269}$       & 98.13$_{0.320}$       & 84.64$_{7.497}$       & 60.63$_{4.068}$       & 68.93$_{2.667}$ \\
Noise$_{Weights}$       & 83.09$_{0.683}$       & 89.24$_{1.021}$       & 97.19$_{0.349}$       & 97.82$_{0.312}$       & 85.40$_{2.145}$       & 60.46$_{2.962}$       & 69.54$_{1.370}$ \\
SWA     & 83.77$_{0.766}$       & 89.18$_{1.447}$       & 97.49$_{0.264}$       & 98.07$_{0.367}$       & 86.06$_{2.042}$       & 61.05$_{3.537}$       & 67.06$_{8.933}$ \\
Mixout  & 82.64$_{0.688}$       & 89.43$_{1.237}$       & 97.37$_{0.321}$       & 98.17$_{0.177}$       & 86.09$_{0.901}$       & 59.62$_{4.338}$       & 67.56$_{7.110}$ \\
Augment 1       & 83.09$_{0.522}$       & 89.66$_{1.202}$       & 97.28$_{0.329}$       & 98.23$_{0.205}$       & 86.46$_{1.782}$       & 60.80$_{4.930}$       & 69.96$_{3.846}$ \\
Augment 2       & 82.47$_{0.820}$       & 88.69$_{1.292}$       & 97.32$_{0.283}$       & 98.18$_{0.118}$       & 87.05$_{0.823}$       & 60.55$_{2.952}$       & 70.36$_{3.699}$ \\
Our$_{DE}$      & 83.02$_{0.209}$       & 89.21$_{0.346}$       & 97.34$_{0.079}$       & 98.09$_{0.044}$       & 86.26$_{0.225}$       & 60.26$_{0.919}$       & 70.17$_{0.568}$ \\
Our$_{NI}$      & 85.50$_{0.652}$       & 90.76$_{0.907}$       & 97.95$_{0.667}$       & 98.46$_{0.245}$       & 88.89$_{0.603}$       & 66.70$_{2.416}$       & 72.84$_{1.095}$ \\
Our$_{DENI}$    & \underline{\textbf{85.78$_{0.156}$}}       & \underline{\textbf{91.00$_{0.241}$}}       & \textbf{98.21$_{0.166}$}       & \textbf{98.68$_{0.055}$}       & \textbf{89.49$_{0.176}$}       & \textbf{67.51$_{0.628}$}       & \textbf{73.09$_{0.441}$} \\
Our$_{DENIALS}$ & 85.58$_{0.169}$       & 90.86$_{0.331}$       & \underline{97.95$_{0.067}$}       & 98.39$_{0.052}$       & \underline{89.37$_{0.102}$}       & \underline{67.40$_{0.577}$}       & \underline{72.61$_{0.437}$} \\ \midrule
\multicolumn{8}{c}{LoRA} \\
Default & 79.77$_{10.530}$      & 76.57$_{23.043}$      & 87.12$_{27.864}$      & 92.62$_{21.163}$      & 72.73$_{21.716}$      & 48.45$_{14.781}$      & 58.72$_{14.973}$ \\
All Data        & 87.17$_{0.451}$       & 81.71$_{22.575}$      & 91.14$_{22.510}$      & 98.93$_{0.506}$       & 87.56$_{17.387}$      & 62.28$_{12.957}$      & 59.86$_{18.586}$ \\
Best Practices  & 78.77$_{16.671}$      & 82.90$_{18.145}$      & 96.15$_{3.883}$       & 97.89$_{0.206}$       & 75.02$_{18.841}$      & 61.43$_{5.798}$       & 55.87$_{18.034}$ \\
Ensemble        & 84.36$_{0.452}$       & 90.11$_{0.699}$       & 97.86$_{0.133}$       & 98.44$_{0.052}$       & 86.67$_{2.238}$       & 64.67$_{4.205}$       & 69.33$_{10.165}$ \\
Noise$_{Input}$ & 82.63$_{0.866}$       & 76.36$_{25.050}$      & 87.66$_{28.042}$      & 97.92$_{0.188}$       & 74.27$_{21.892}$      & 55.94$_{14.109}$      & 61.62$_{18.108}$ \\
Noise$_{Weights}$       & 81.90$_{2.025}$       & 83.50$_{9.570}$       & 88.98$_{22.796}$      & 86.63$_{27.099}$      & 74.20$_{21.399}$      & 57.76$_{11.426}$      & 57.38$_{17.570}$ \\
SWA     & 74.06$_{23.224}$      & 69.66$_{32.682}$      & 83.42$_{33.489}$      & 87.86$_{26.928}$      & 54.70$_{23.038}$      & 48.28$_{14.489}$      & 48.01$_{17.929}$ \\
Mixout  & 82.45$_{0.905}$       & 83.40$_{18.324}$      & 97.00$_{0.983}$       & 96.83$_{4.359}$       & 74.78$_{21.506}$      & 56.87$_{12.139}$      & 58.36$_{19.808}$ \\
Augment 1       & 82.33$_{1.315}$       & 76.75$_{24.220}$      & 87.80$_{27.988}$      & 97.95$_{1.020}$       & 72.68$_{22.844}$      & 55.59$_{11.736}$      & 58.15$_{17.500}$ \\
Augment 2       & 82.37$_{1.293}$       & 78.13$_{24.119}$      & 92.70$_{20.126}$      & 98.09$_{0.163}$       & 76.13$_{20.011}$      & 58.94$_{9.478}$       & 64.62$_{13.715}$ \\
Our$_{DE}$      & 80.47$_{0.316}$       & 88.09$_{0.689}$       & 97.83$_{0.123}$       & 98.24$_{0.074}$       & \underline{86.18$_{0.475}$}       & 64.84$_{0.956}$       & \underline{71.67$_{1.595}$} \\
Our$_{NI}$      & 84.78$_{0.901}$       & 88.20$_{5.218}$       & 97.37$_{1.502}$       & 93.41$_{21.096}$      & 84.35$_{11.181}$      & 60.77$_{11.306}$      & 59.19$_{21.022}$ \\
Our$_{DENI}$    & \underline{\textbf{85.21$_{0.111}$}}       & \underline{\textbf{91.82$_{0.560}$}}       & \underline{\textbf{98.45$_{0.079}$}}       & 98.79$_{0.047}$       & 87.92$_{0.774}$       & \textbf{68.87$_{0.943}$}       & 72.95$_{1.791}$ \\
Our$_{DENIALS}$ & 84.73$_{0.149}$       & 91.76$_{0.598}$       & 98.02$_{0.504}$       & \underline{\textbf{98.81$_{0.032}$}}       & \textbf{87.93$_{1.055}$}       & \underline{68.76$_{0.721}$}       & \textbf{73.36$_{1.730}$} \\ \midrule
\multicolumn{8}{c}{IA3} \\
Default & 80.90$_{1.296}$       & 77.85$_{11.902}$      & 96.34$_{0.702}$       & 95.99$_{5.347}$       & 73.63$_{17.822}$      & 51.37$_{10.640}$      & 54.89$_{13.130}$ \\
All Data        & 85.57$_{0.631}$       & 87.23$_{17.164}$      & 97.97$_{1.760}$       & 98.51$_{0.818}$       & 90.37$_{2.450}$       & 65.53$_{3.442}$       & 63.60$_{13.290}$ \\
Best Practices  & 78.84$_{7.546}$       & 84.22$_{5.928}$       & 96.72$_{0.469}$       & 95.04$_{9.257}$       & 77.77$_{12.933}$      & 55.46$_{4.093}$       & 50.14$_{11.646}$ \\
Ensemble        & 84.58$_{0.441}$       & 90.11$_{0.830}$       & 97.73$_{0.118}$       & 98.19$_{0.064}$       & 84.99$_{1.262}$       & 56.57$_{6.732}$       & 54.84$_{11.430}$ \\
Noise$_{Input}$ & 74.27$_{21.131}$      & 76.73$_{24.718}$      & 92.85$_{13.980}$      & 93.18$_{18.298}$      & 75.14$_{14.414}$      & 51.77$_{11.849}$      & 47.77$_{15.243}$ \\
Noise$_{Weights}$       & 81.05$_{2.024}$       & 77.91$_{20.630}$      & 95.05$_{5.857}$       & 95.66$_{4.713}$       & 75.60$_{11.375}$      & 54.77$_{7.076}$       & 48.41$_{9.886}$ \\
SWA     & 81.37$_{6.187}$       & 85.44$_{11.288}$      & 93.33$_{15.927}$      & 96.12$_{6.814}$       & 73.04$_{18.684}$      & 51.83$_{12.638}$      & 54.21$_{15.483}$ \\
Mixout  & 78.54$_{5.491}$       & 56.34$_{28.193}$      & 84.67$_{23.558}$      & 86.39$_{22.527}$      & 46.27$_{17.586}$      & 42.62$_{13.977}$      & 36.26$_{9.978}$ \\
Augment 1       & 82.76$_{1.926}$       & 82.25$_{13.731}$      & 95.11$_{4.686}$       & 97.30$_{3.050}$       & 80.76$_{11.785}$      & 54.16$_{9.843}$       & 64.78$_{7.860}$ \\
Augment 2       & 82.39$_{3.087}$       & 79.28$_{18.981}$      & 93.06$_{19.200}$      & 97.93$_{0.854}$       & 81.69$_{10.545}$      & 58.18$_{4.052}$       & \textbf{67.19$_{6.209}$} \\
Our$_{DE}$      & 82.87$_{0.209}$       & 87.88$_{0.407}$       & 97.66$_{0.075}$       & 97.94$_{0.097}$       & 84.84$_{0.698}$       & 59.99$_{0.408}$       & \underline{61.99$_{1.657}$} \\
Our$_{NI}$      & 83.82$_{1.657}$       & 88.99$_{2.079}$       & 97.64$_{0.225}$       & 98.18$_{0.366}$       & 81.50$_{12.484}$      & 60.40$_{4.143}$       & 57.08$_{15.157}$ \\
Our$_{DENI}$    & \underline{84.96$_{0.143}$}       & \underline{\textbf{91.02$_{0.320}$}}       & \underline{97.94$_{0.066}$}       & \underline{98.35$_{0.051}$}       & 86.88$_{0.644}$       & \underline{\textbf{62.55$_{0.468}$}}       & 63.35$_{2.045}$ \\
Our$_{DENIALS}$ & \textbf{85.24$_{0.163}$}       & 90.99$_{0.716}$       & \textbf{98.08$_{0.123}$}       & \textbf{98.53$_{0.275}$}       & \underline{\textbf{87.27$_{0.441}$}}       & 62.51$_{0.751}$       & 64.12$_{2.093}$ \\ \midrule
\multicolumn{8}{c}{UniPELT} \\
Default & 74.93$_{2.312}$       & 51.52$_{6.672}$       & 83.82$_{7.394}$       & 83.05$_{16.371}$      & 80.73$_{8.096}$       & 58.91$_{1.770}$       & 60.37$_{6.526}$ \\
All Data        & 84.21$_{3.139}$       & 78.14$_{4.317}$       & 97.82$_{0.441}$       & 99.00$_{0.034}$       & 93.10$_{2.212}$       & 65.22$_{1.804}$       & 74.23$_{7.717}$ \\
Best Practices  & 79.66$_{0.960}$       & 69.89$_{8.061}$       & 93.48$_{2.481}$       & 95.52$_{1.320}$       & 84.04$_{1.959}$       & 57.31$_{2.411}$       & 63.67$_{7.417}$ \\
Ensemble        & 83.15$_{0.387}$       & 84.49$_{1.287}$       & 96.92$_{0.308}$       & 97.61$_{0.161}$       & 87.17$_{0.487}$       & 60.63$_{0.955}$       & 68.46$_{1.455}$ \\
Noise$_{Input}$ & 76.92$_{8.502}$       & 66.41$_{7.107}$       & 92.94$_{2.395}$       & 89.56$_{20.237}$      & 81.47$_{11.838}$      & 57.85$_{2.175}$       & 65.19$_{4.367}$ \\
Noise$_{Weights}$       & 80.16$_{0.903}$       & 73.46$_{7.650}$       & 94.69$_{1.304}$       & 92.64$_{14.449}$      & 82.37$_{8.680}$       & 59.22$_{2.550}$       & 61.56$_{11.011}$ \\
SWA     & 79.99$_{1.582}$       & 74.04$_{6.722}$       & 94.51$_{1.184}$       & 96.05$_{1.194}$       & 81.97$_{9.083}$       & 58.29$_{2.459}$       & 60.22$_{14.255}$ \\
Mixout  & 79.63$_{1.743}$       & 72.86$_{5.823}$       & 94.55$_{1.471}$       & 94.38$_{4.587}$       & 79.05$_{12.674}$      & 57.85$_{1.957}$       & 63.49$_{6.818}$ \\
Augment 1       & 77.67$_{13.450}$      & 82.67$_{2.436}$       & 95.55$_{0.665}$       & 97.26$_{0.547}$       & 82.55$_{9.967}$       & 57.88$_{1.454}$       & 68.52$_{3.480}$ \\
Augment 2       & 80.47$_{1.248}$       & 81.60$_{2.290}$       & 95.36$_{0.597}$       & 97.08$_{0.884}$       & 83.69$_{6.801}$       & 57.47$_{1.482}$       & 68.55$_{1.417}$ \\
Our$_{DE}$      & 81.15$_{0.217}$       & 80.47$_{1.109}$       & 96.25$_{0.325}$       & 97.23$_{0.096}$       & 85.81$_{0.349}$       & 59.88$_{0.489}$       & 68.23$_{0.794}$ \\
Our$_{NI}$      & 81.01$_{3.406}$       & 76.86$_{5.754}$       & 95.47$_{1.276}$       & 97.40$_{0.670}$       & 85.38$_{3.221}$       & 59.73$_{1.891}$       & 64.91$_{6.295}$ \\
Our$_{DENI}$    & \underline{\textbf{83.91$_{0.160}$}}       & 85.48$_{0.914}$       & \textbf{97.46$_{0.221}$}       & 97.89$_{0.095}$       & \underline{\textbf{87.89$_{0.335}$}}       & 62.32$_{0.309}$       & 72.07$_{0.552}$ \\
Our$_{DENIALS}$ & 83.78$_{0.182}$       & \underline{\textbf{86.20$_{0.529}$}}       & \underline{96.74$_{0.090}$}       & \underline{\textbf{98.96$_{0.047}$}}       & 87.64$_{0.564}$       & \underline{\textbf{62.79$_{0.232}$}}       & \underline{\textbf{72.60$_{0.231}$}} \\
\bottomrule
\end{tabular}
\end{sc}
\end{center}
\caption{Comparison of the DENI method with existing mitigation strategies and baselines on full fine-tuning, LoRA, IA3 and UniPELT using ALBERT model. The comparison is done in terms of overall performance and deviation. The highest performance for each fine-tuning method is in \textbf{bold} and lowest deviation is \underline{underlined} (not considering \textit{All Data} baseline).}
\label{tab:results-albert}
\end{table*}

\end{document}